\documentclass[sigconf]{acmart}
\AtBeginDocument{%
  }

\setcopyright{acmlicensed}
\copyrightyear{2018}
\acmYear{2018}
\acmDOI{XXXXXXX.XXXXXXX}

\acmConference[ACM KDD 2025]{31st SIGKDD Conference on Knowledge Discovery and Data Mining - Research Track (August 2024 Deadline)}{Aug 03--07,
  2025}{Toronto, ON, Canada}
\acmISBN{978-1-4503-XXXX-X/18/06}

\usepackage{caption}
\usepackage{subcaption}
\usepackage[para,online,flushleft]{threeparttable}
\usepackage{multirow}
\usepackage[nice]{nicefrac}
\newtheorem{lemma}{Lemma}

\DeclareMathOperator{\vol}{vol}
\DeclareMathOperator{\mlp}{MLP}

\begin{document}

\title{Attribute-Enhanced Similarity Ranking for Sparse Link Prediction}

\definecolor{darkgreen}{RGB}{20, 99, 44}

\author{João Mattos}
\authornote{Corresponding Author.}
\email{jrm28@rice.edu}
\orcid{0000-0002-6877-16082}
\affiliation{%
  \institution{Rice University}
  \city{Houston}
  \state{Texas}
  \country{USA}
}

\author{Zexi Huang}
\authornote{Work done prior to joining TikTok Inc.}
\email{zexi_huang@cs.ucsb.edu}
\affiliation{%
  \institution{UC Santa Barbara}
  \city{Santa Barbara}
  \state{California}
  \country{USA}}

\author{Mert Kosan}
\authornote{Work done prior to joining Visa Inc.}
\email{mertkosan@ucsb.edu}
\affiliation{%
 \institution{UC Santa Barbara}
 \city{Santa Barbara}
 \state{California}
 \country{USA}}

\author{Ambuj Singh}
\email{ambuj@cs.ucsb.edu}
\affiliation{%
  \institution{UC Santa Barbara}
  \city{Santa Barbara}
  \state{California}
  \country{USA}}

\author{Arlei Silva}
\email{arlei@rice.edu}
\affiliation{%
  \institution{Rice University}
  \city{Houston}
  \state{Texas}
  \country{USA}}

\renewcommand{\shortauthors}{Mattos et al.}

\begin{abstract}
  Link prediction is a fundamental problem in graph data. In its most realistic setting, the problem consists of predicting missing or future links between random pairs of nodes from the set of disconnected pairs. Graph Neural Networks (GNNs) have become the predominant framework for link prediction. GNN-based methods treat link prediction as a binary classification problem and handle the extreme class imbalance---real graphs are very sparse---by sampling (uniformly at random) a balanced number of disconnected pairs not only for training but also for evaluation. However, we show that the reported performance of GNNs for link prediction in the balanced setting does not translate to the more realistic imbalanced setting and that simpler topology-based approaches are often better at handling sparsity. These findings motivate Gelato, a similarity-based link-prediction method that applies (1) graph learning based on node attributes to enhance a topological heuristic, (2) a ranking loss for addressing class imbalance, and (3) a negative sampling scheme that efficiently selects hard training pairs via graph partitioning. Experiments show that Gelato outperforms existing GNN-based alternatives.
\end{abstract}



\maketitle

\section{Introduction}
\label{sec::introduction}


Machine learning on graphs supports various structured-data applications including social network analysis \citep{tang2008arnetminer, li2017deepcas, qiu2018deepinf}, recommender systems \citep{jamali2009trustwalker,monti2017geometric,wang2019kgat}, natural language processing \citep{sun2018open, sahu2019inter, yao2019graph}, and physics modeling \citep{sanchez2018graph,ivanovic2019trajectron,da2020combining}. Among the graph-related tasks, one could argue that link prediction, which consists of predicting missing or future links \citep{lu2011link, martinez2016survey}, is the most fundamental one. This is because link prediction not only has many concrete applications \citep{qi2006evaluation, liben2007link} 
but can also be considered an (implicit or explicit) step of the graph-based machine learning pipeline \citep{martin2016structural, bahulkar2018community, wilder2019end}---as the observed graph is usually noisy and/or incomplete. 

Graph Neural Networks (GNNs) \citep{kipf2016semi, hamilton2017inductive, velivckovic2018graph} have emerged as the predominant paradigm for machine learning on graphs. Similar to their great success in node classification \citep{klicpera2018predict, wu2019simplifying, zheng2020robust} and graph classification \citep{ying2018hierarchical, zhang2018end, morris2019weisfeiler}, GNNs have been shown to achieve state-of-the-art link prediction performance \citep{zhang2018link,liu2020feature,pan2021neural,yun2021neo,chamberlain2022graph, wang2023neural}. Compared to classical approaches that rely on expert-designed heuristics to extract topological information (e.g., Common Neighbors \citep{newman2001clustering}, Adamic-Adar \citep{adamic2003friends}, Preferential Attachment \citep{barabasi2002evolution}), GNNs can naturally incorporate attributes and are believed to be able to learn new effective heuristics directly from data via supervised learning. \looseness=-1

However, we argue that \textit{the evaluation of GNN-based link prediction methods paints an overly optimistic view of their model performance}. Most real graphs are sparse and have a modular structure \citep{barabsi2016network,newman2018networks}. In \textsc{Cora} and \textsc{Citeseer} (citation networks), less than 0.2\% of the node pairs are links/positive (see \autoref{tab::dataset}) and modules arise around research topics. Yet GNN-based link prediction methods are evaluated on an artificially balanced test set that includes every positive pair but only a small sample of the negative ones chosen uniformly at random \citep{hu2020open}. Due to modularity, the majority of negative pairs sampled are expected to be relatively far from each other (i.e. across different modules) compared to positive pairs. As a consequence, performance metrics reported for this balanced setting, which we call \textit{biased testing}, differ widely from the ones observed for the more challenging \textit{unbiased testing}, where the test set includes every disconnected pair of nodes. In particular, we have found that unsupervised topological heuristics are more competitive in the \emph{unbiased setting}, often outperforming recent GNN-based link prediction methods. This finding has motivated us to rethink the design of link prediction methods for sparse graphs.

A key hypothesis of our work is that effective unbiased link prediction in sparse graphs requires a similarity metric that can distinguish positive pairs from hard negative ones. More specifically, link prediction should be seen as a ``needle in the haystack'' type of problem, where extreme class imbalance makes even the most similar pairs still more likely to be negative. Existing GNN-based approaches fail in this sparse regime due to (1) their use of a binary classification loss that is highly sensitive to class imbalance; (2) their \emph{biased training} that mimics \emph{biased testing}; (3) their inability to learn effective topological heuristics directly from data.

The goal of this paper is to address the key limitations of GNNs for link prediction mentioned above. We present \emph{Gelato}, a novel similarity-based framework for link prediction that combines a topological heuristic and graph learning to leverage both topological and attribute information. Gelato applies a ranking-based N-pair loss and partitioning-based negative sampling to select hard training node pairs. Extensive experiments demonstrate that our model significantly outperforms state-of-the-art GNNs in both accuracy and scalability. \autoref{fig::overview} provides an overview of our approach.




To summarize, our contributions are: (1) We scrutinize the evaluation of supervised link prediction methods and identify their limitations in handling class imbalance; (2) we propose a simple, effective, and efficient framework to combine topological and attribute information for link prediction in an innovative fashion;
(3) we introduce an N-pair link prediction loss that we show to be more effective at addressing class imbalance; and (4) we propose an efficient partitioning-based negative sampling scheme that improves link prediction generalization in the sparse setting.

\looseness=-1
\section{Limitations in supervised link prediction evaluation}
\label{sec::limitations}
Supervised link prediction is often formulated as binary classification, where the positive (or negative) class includes node pairs connected (or not connected) by a link. A key difference between link prediction and other classification problems is that the two classes in link prediction are \emph{extremely} imbalanced as most graphs of interest are sparse---e.g. the datasets from \autoref{tab::dataset} are significantly more imbalanced than those in \cite{tang2008svms}. However, the class imbalance is not properly addressed in the evaluation of existing approaches. 


Existing link prediction methods \citep{kipf2016variational, zhang2018link, chami2019hyperbolic, zhang2021lorentzian, cai2021line, yan2021link, zhu2021neural, chen2022bscnets, pan2021neural} are evaluated on a test set containing all positive test pairs and only an equal number of random negative pairs. Similarly, the Open Graph Benchmark (OGB) ranks predicted links against a very small sample of random negative pairs.  We term these approaches \emph{biased testing} as they highly overestimate the ratio of positive pairs in the graph. This issue is exacerbated in most real graphs, where community structure \citep{newman2006modularity} causes random negative pairs to be particularly easy to identify \cite{li2024evaluating}---they likely involve members of different communities. Evaluation metrics based on biased testing provide an overly optimistic assessment of the performance in \emph{unbiased testing}, where every negative pair is included in the test set. In fact, in real applications where positive test edges are not known a priori, it is impossible to construct those biased test sets to begin with. 


Regarding evaluation metrics, Area Under the Receiver Operating Characteristic Curve (AUC) and Average Precision (AP) are the two most popular evaluation metrics for supervised link prediction \citep{kipf2016variational, zhang2018link, chami2019hyperbolic, zhang2021lorentzian, cai2021line, yan2021link, zhu2021neural, chen2022bscnets, pan2021neural}. We first argue that, as in other imbalanced classification problems \citep{davis2006relationship, saito2015precision}, AUC is not an effective evaluation metric for link prediction as it is biased towards the majority class (non-edges). On the other hand, AP and other rank-based metrics such as Hits@$k$---used in OGB \citep{hu2020open}---are effective for imbalanced classification \emph{but only if evaluated on an unbiased test}.


\emph{Example}: Consider an instance of Stochastic Block Model (SBM) \citep{karrer2011stochastic} with $10$ blocks of size 1k, intra-block density $0.9$, and inter-block density $0.1$. The number of inter-block negative pairs is $10\times 1\text{k}\times(10-1)\times 1\text{k}\times (1-0.1)/2=40.5\text{M}$, while the number of intra-block negative pairs, which have high topological similarities like the ground-truth positive pairs and are much harder to contrast against, is $10\times 1\text{k}\times 1\text{k} \times(1-0.9)/2=0.5\text{M}$. Biased testing would select less than $0.5\text{M}/(0.5\text{M}+40.5\text{M})<2$\% of the test negative pairs among the (hard) intra-block ones.  In this scenario, even a random classifier 
is expected to obtain 50\% precision. However, the expected precision drops to less than 22\% (9M positive pairs vs. 41M negative pairs) under \textit{unbiased testing}. 

We will formalize the argument used in the example above by performing link prediction on a generic instance of the SBM with intra-block density $p$, inter-block density $q$, where $p > q$, and $k$ blocks of size $n$. In particular, we will consider an instance of SBM corresponding to the expected node pattern given the parameters, where a node is connected to $(n-1)p$ other nodes within its block and $(nk-n)q$ nodes outside its block. In this setting, the optimal link prediction algorithm can only distinguish potential links within or across blocks---as pairs within each set are connected with probability $p$ and $q$, respectively.

\begin{lemma}
The ratio $\alpha$ between inter-cluster and intra-cluster negative node pairs in the SBM is such that:
\begin{equation*}
\alpha \geq (k-1)\frac{1-q}{1-p}
\end{equation*}
\end{lemma}

The above lemma follows directly from the definition of the SBM and shows that the set of negative pairs is dominated by (easy) inter-cluster pairs as $p$ increases compared to $q$. 

\begin{theorem}
In the unbiased setting, the optimal accuracy link prediction method based on binary classification for the SBM predicts no links if $p < 0.5$.
\label{thm::unbiased}
\end{theorem}

The proof is given in the Appendix \ref{ap:proof_unbiased}. Intuitively, even if the classifier has access to the SBM block structure, most within-block pairs are disconnected and thus the accuracy is maximized if no links are predicted. On the other hand, if $p > q$, an effective link prediction method should be able to leverage the SBM block structure to predict within block links. This motivates our formulation of link prediction as a ``needle in the haystack'' type of problem, where even the top candidate links (i.e., within-block pairs) are still more likely to be negative due to the sparsity of the graph. We show datasets considered fit this scenario, as shown in Appendix \ref{ap::estimated_sbm}.

\begin{lemma}
    In the biased setting, there exist non-trivial link prediction methods with optimal accuracy based on binary classification for the SBM with $p < 0.5$. 
    \label{lemma::biased}
\end{lemma}

The proof is given in the Appendix \ref{lemma::biased_proof}. The idea is that in the biased setting, a link prediction method that predicts within-block pairs as links can outperform the trivial classifier described in Theorem \ref{thm::unbiased}. This illustrates how biased testing, which is applied by recent work on supervised link prediction, can be misleading for sparse graphs. More specifically, a model trained under the biased setting might perform poorly if evaluated in the, more realistic, unbiased setting due to possibly unforeseen distribution shifts across the settings. This is a key motivation for our work.

The above discussion motivates a more representative evaluation setting for supervised link prediction. We argue for the use of rank-based evaluation metrics---AP, Precision@$k$ \citep{lu2011link}, and Hits@$k$ \citep{bordes2013translating}---with \emph{unbiased testing}, where positive edges are ranked against hard negative node pairs. 
These metrics have been widely applied in related problems, such as unsupervised link prediction \citep{lu2011link, ou2016asymmetric, zhang2018arbitrary, random-walk-embedding}, knowledge graph completion \citep{bordes2013translating, yang2015embedding, sun2018rotate}, and information retrieval \citep{schutze2008introduction}, where class imbalance is also significant. 
In our experiments, we will illustrate how these evaluation metrics combined with \textit{unbiased testing} provide a drastically different and more informative performance evaluation compared to existing approaches. 

\section{Method}

\begin{figure*}
    \centering
    \includegraphics[width=1\textwidth]{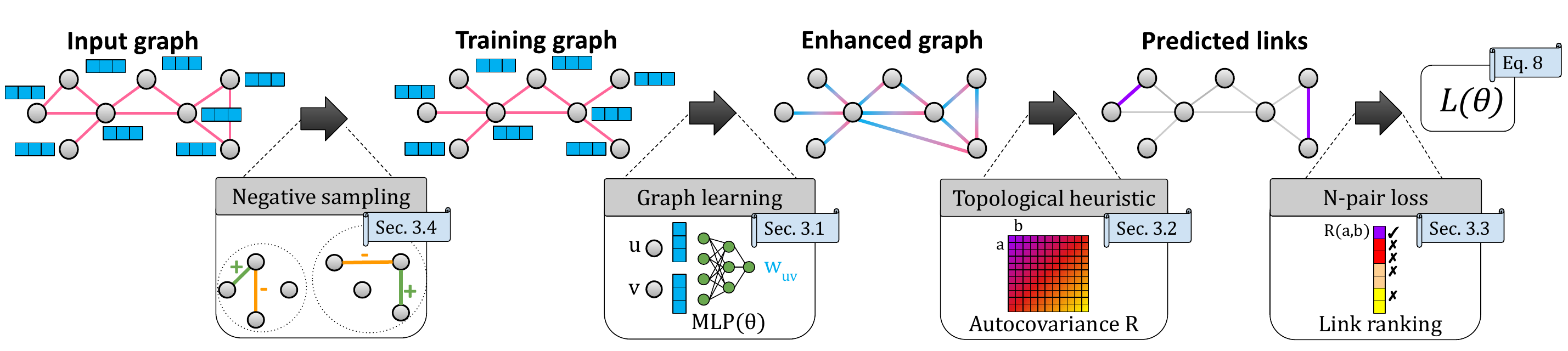}
    \caption{Gelato applies graph learning to incorporate attribute information into the topology. The learned graph is given to a topological heuristic that predicts edges between node pairs with high Autocovariance similarity. The parameters of the MLP are optimized end-to-end using the N-pair loss over node pairs selected via a partitioning-based negative sampling scheme. Experiments show that Gelato outperforms state-of-the-art GNN-based link prediction methods. \looseness=-1
    }
    \label{fig::overview}
\end{figure*}

\label{sec::method}

The limitations of supervised link prediction methods, including GNNs, to handle \emph{unbiased testing} in sparse graphs motivate the design of a novel link prediction approach. First, preliminary results (see Table \ref{tab::performance_ap}) have shown that topological heuristics are not impacted by class imbalance. That is because these heuristics are sensitive to small differences in structural similarity between positive and hard negative pairs while not relying on any learning---and thus not being affected by biased training. However, local structure proximity heuristics, such as Common Neighbors, are known to be less efficient in highly sparse scenarios observed in many real-world applications \cite{mao2023revisiting}---Table \ref{tab::dataset} shows the sparsity of our datasets. Further, unlike GNNs, topological heuristics are unable to leverage attribute information. Our approach addresses these limitations by integrating supervision into a powerful topological heuristic to leverage attribute data via graph learning.

\textbf{Notation and problem.} Consider an attributed graph $G = (V, E, X)$, where $V$ is the set of $n$ nodes, $E$ is the set of $m$ edges (links), and $X = (x_1, ..., x_n)^T \in \mathbb{R}^{n\times r}$ collects $r$-dimensional node attributes. The topological (structural) information of the graph is represented by its adjacency matrix $A \in \mathbb{R}^{n\times n}$, with $A_{uv} > 0$ if an edge of weight $A_{uv}$ connects nodes $u$ and $v$ and $A_{uv} = 0$, otherwise. The (weighted) degree of node $u$ is given as $d_u = \sum_{v}A_{uv}$ and the corresponding degree vector (matrix) is denoted as $d \in \mathbb{R}^n$ ($D \in \mathbb{R}^{n\times n})$. The volume of the graph is $\vol(G)=\sum_u d_u$. Our goal is to infer missing links in $G$ based on its topological and attribute information, $A$ and $X$. \looseness=-1

\textbf{Model overview.} \autoref{fig::overview} provides an overview of our model. It starts by selecting training node pairs using a novel partitioning-based negative sampling scheme. Next, a topology-centric graph learning phase incorporates node attribute information directly into the graph structure via a Multi-layer Perceptron (MLP).  
We then apply a topological heuristic, Autocovariance (AC), to the attribute-enhanced graph to obtain a pairwise score matrix. Node pairs with the highest scores are predicted as links. The scores for training pairs are collected to compute an N-pair loss. 
Finally, the loss is used to train the MLP parameters in an end-to-end manner. We name our model Gelato (\underline{G}raph \underline{e}nhancement for \underline{l}ink prediction with \underline{a}u\underline{to}covariance). Gelato represents a different paradigm in supervised link prediction combining a graph encoding of attributes with a topological heuristic instead of relying on node embeddings. While the building blocks of Gelato have been proposed by previous work, our paper is the first to apply these building blocks to address challenges in supervised link prediction for sparse graphs.


\subsection{Graph learning}
\label{subsec::graph_learning}
The goal of graph learning is to generate an enhanced graph that incorporates node attribute information into the topology. This can be considered as the ``dual'' operation of message-passing in GNNs, which incorporates topological information into attributes (embeddings). We propose graph learning as a more suitable scheme to combine attributes and topology for link prediction since it does not rely on the GNN to learn a topological heuristic, which we have verified empirically to be a challenge. \looseness=-1


Specifically, our first step of graph learning is to augment the original edges with a set of node pairs based on their (untrained) attribute similarity (i.e., adding an $\epsilon$-neighborhood graph): 
\begin{equation}
    \widetilde{E} = E + \{(u, v) \mid s(x_u, x_v) > \epsilon_\eta\}
\end{equation}
where $s(\cdot)$ can be any similarity function (we use cosine in our experiments) and $\epsilon_\eta$ is a threshold that determines the number of added pairs as a ratio $\eta$ of the original number of edges $m$.

A simple MLP then maps the pairwise node attributes into a trained edge weight for every edge in $\widetilde{E}$:
\begin{equation}
    w_{uv} = \mlp([x_u; x_v]; \theta)
\end{equation}
where $[x_u;x_v]$ denotes the concatenation of $x_u$ and $x_v$ and $\theta$ contains the trainable parameters. For undirected graphs, we instead use the following permutation invariant operator \citep{chen2014unsupervised}:
\begin{equation}
    w_{uv} = \mlp([x_u+x_v; |x_u - x_v|]; \theta)
\end{equation}

The final weights of the enhanced graph are a combination of the topological, untrained, and trained weights:
\begin{equation}
    \widetilde{A}_{uv} = \alpha A_{uv} + (1-\alpha)(\beta w_{uv} + (1-\beta) s(x_u, x_v))
\end{equation}
where $\alpha$ and $\beta$ are hyperparameters. The enhanced adjacency matrix $\widetilde{A}$ is then fed into a topological heuristic for link prediction introduced in the next section. The MLP is not trained directly to predict the links but instead trained end-to-end to enhance the input graph given to the topological heuristic.
Further, the MLP can be easily replaced by a more powerful model such as a GNN (see Appendix \ref{ap::gnn}), but the goal of this paper is to demonstrate the general effectiveness of our framework and we will show that even a simple MLP leads to significant improvement over the base heuristic. 

\subsection{Topological heuristic}
\label{sec:topological-heuristic}
Assuming that the learned adjacency matrix $\widetilde{A}$ incorporates structural and attribute information, Gelato applies a topological  
heuristic to $\widetilde{A}$. 
Specifically, we generalize Autocovariance, which has been shown to be effective for non-attributed graphs \citep{random-walk-embedding}, to the attributed setting. 
Autocovariance is a random-walk-based similarity metric. Intuitively, it measures the difference between the co-visiting probabilities for a pair of nodes in a truncated walk and in an infinitely long walk. 
Given the enhanced graph $\widetilde{G}$, the Autocovariance similarity matrix $R \in \mathbb{R}^{n\times n}$ is given as
\begin{equation}
    \label{eqn:autocov}
    R = \frac{\widetilde{D}}{\text{vol}(\widetilde{G})}(\widetilde{D}^{-1}\widetilde{A})^t - \frac{\Tilde{d}\Tilde{d}^T}{\text{vol}^2(\widetilde{G})}
\end{equation}
where $t \in \mathbb{N}_0$ is the scaling parameter of the truncated walk. Each entry $R_{uv}$ represents a similarity score for node pair $(u, v)$, and top similarity pairs are predicted as links. Note that $R_{uv}$ only depends on the $t$-hop enclosing subgraph of $(u, v)$ and can be easily differentiated with respect to the edge weights in the subgraph. Gelato could be applied with any differentiable topological heuristics or even a combination of them. In our experiments (Section \ref{subsec::lp_results}), we will show that Autocovariance alone enables state-of-the-art link prediction without requiring any learning. Moreover, Appendix \ref{ap::learn_autocov} discusses whether GNNs can learn Autocovariance from data.

\noindent\textbf{Autocovariance versus other heuristics.} Following \cite{mao2023revisiting}, we show that local structural heuristics commonly employed by GNNs, such as Common Neighbors, exhibit reduced efficacy in sparse networks with less informative neighborhood structures. This observation motivates our selection of Autocovariance as our topological heuristic, given its ability to capture global structural patterns through random walks. Further, the parameter $t$ in Autocovariance offers adaptability to varying network sparsity levels\cite{mao2023revisiting}, ranging from denser (lower $t$ values) to sparser (higher $t$ values) networks.


\noindent\textbf{Autocovariance distinguishes negative pairs.} Autocovariance can be seen as a general case of the Modularity metric $Q$ \cite{delvenne2010stability}:
\begin{equation}
\label{eq:modularity}
    Q=\cfrac{1}{4m}\sum_{ij}(A_{ij} - \cfrac{d_id_j}{2m})s_is_j,
\end{equation}
in which $m=\text{vol}(G)/2$, $d_i$ and $d_j$ are the degrees of nodes $i$ and $j$, and $s_is_j$ is a product that indicates whether both nodes are in the same partition. More specifically, for $t=1$, Autocovariance expresses the graph partitioning resulting in the optimal Modularity value, which captures the relationship between the expected number of edges between two partitions compared to the probability of any random edge in the graph. This key property directly applies to our scenario, enabling Gelato to distinguish between hard (same partitions) and easy (different partitions) negative pairs and motivating us to adopt Autocovariance as our graph heuristic. Further, as $t$ increases, Autocovariance expresses growing coarser partitions until approximating spectral clustering (for $t \rightarrow \infty$), being flexible regarding partition sizes according to different domains.

\noindent\textbf{Scaling up Gelato with batching and sparse operations.} Naively implementing Gelato using dense tensors is infeasible, due to the quadratic VRAM requirement ($R \in \mathbb{R}^{|V| \times |V|}$). To address this limitation, we propose storing $\widetilde{A}$ as a sparse matrix. Then, instead of directly computing $(\widetilde{D}^{-1}\widetilde{A})^t$ from \autoref{eqn:autocov} (resulting on a dense $|V| \times |V|$ matrix), we compute

\begin{equation}
\begin{aligned}
    \hspace{25pt} P_{l + 1} = P_l(\widetilde{D}^{-1}\widetilde{A}),\hspace{25pt} l \in \{1, 2, ..., t\}
\end{aligned}
\end{equation}

\begin{figure}[t!]
    \centering
    \includegraphics[width=0.5\textwidth]{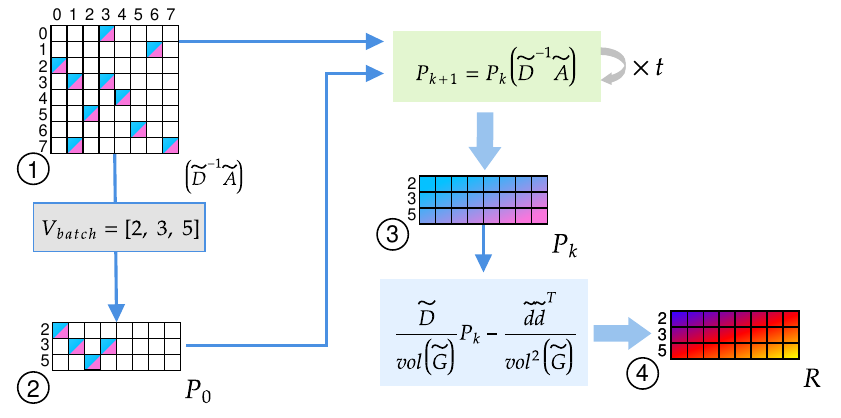}
    \caption{Scaling up Gelato using batching and sparse tensors. We represent sparse tensors (1 and 2) as matrices with blank entries and dense tensors (3 and 4) as color-filled matrices. We extract from the enhanced transition matrix (1) a slice $P_0$ (2) given a batch of node indices $V_{batch}$. Instead of a  matrix exponentiation, we compute $P_0$ $(\widetilde{D}^{-1}\widetilde{A})$ repeatedly for $t$ times to obtain $P_k$ (3), a dense tensor. Finally, we use $P_k$ to obtain the autocovariance $R$ (4) for nodes in the batch. This is implemented efficiently using dense-sparse tensor multiplication. }
    
    \label{fig:sparse_gelato}
\end{figure}

\begin{equation}
\begin{aligned}
    \hspace{25pt} R = \frac{\widetilde{D}}{\text{vol}(\widetilde{G})}P_t - \frac{\Tilde{d}\Tilde{d}^T}{\text{vol}^2(\widetilde{G})}
\end{aligned}
\end{equation}
where $P_0=(\widetilde{D}^{-1}\widetilde{A})_{ij}$, for all $i \in V_{batch}$, where $V_{batch}$ consists of the nodes in the current batch. This operation substitution allows us to compute a sequence of $t$ multiplications between a dense $P_k \in \mathbb{R}^{|batch| \times |V|}$ matrix and a sparse matrix $(\widetilde{D}^{-1}\widetilde{A}) \in \mathbb{R}^{|V| \times |V|}$ instead of a dense matrix power operation, $(\widetilde{D}^{-1}\widetilde{A})^t$. The overall VRAM usage is reduced from $O(|V|^2)$ to $O(|batch| \cdot |V|)$.


\subsection{N-pair loss}
\label{subsec::loss}
Supervised link prediction methods rely on the cross entropy loss (CE) to optimize model parameters. However, CE is known to be sensitive to class imbalance \citep{byrd2019effect}. Instead, Gelato leverages the N-pair loss \citep{sohn2016improved} that is inspired by the metric learning and learning-to-rank literature \citep{mcfee2010metric,cakir2019deep, revaud2019learning, wang2019ranked} to train the parameters of our graph learning model from highly imbalanced \emph{unbiased training} data. 


The N-pair loss (NP) contrasts each positive training edge $(u,v)$ against a set of negative pairs $N(u,v)$. It is computed as follows:
\begin{equation}
    L(\theta) = -\sum_{(u, v) \in E} \log \frac{\exp(R_{uv})}{\exp(R_{uv}) + \sum_{(p, q) \in N(u,v)} \exp(R_{pq})}
\end{equation}

Intuitively, $L(\theta)$ is minimized when each positive edge $(u, v)$ has a much higher similarity than its contrasted negative pairs: $R_{uv} \gg R_{pq}, \forall (p, q) \in N(u,v)$. Compared to CE, NP is more sensitive to negative pairs that have comparable similarities to those of positive pairs---they are more likely to be false positives. While NP achieves good performance in our experiments, alternative losses from the learning-to-rank literature \citep{freund2003efficient,xia2008listwise,bruch2021alternative} could also be applied. 

\subsection{Negative sampling}
\label{sec::neg_sampling}



Supervised methods for link prediction sample a small number of negative pairs uniformly at random but most of these pairs are expected to be easy (see Section \ref{sec::limitations}). To minimize distribution shifts between training and test, negative samples $N(u, v)$ should ideally be generated using \emph{unbiased training} (see additional example in  Appendix \ref{ap::example}). This means that $N(u, v)$ is a random subset of all disconnected pairs in the training graph, and $|N(u,v)|$ is proportional to the ratio of negative pairs. In this way, we enforce $N(u, v)$ to include hard negative pairs. However, due to graph sparsity (see \autoref{tab::dataset}), this approach does not scale to large graphs as the total number of negative pairs would be $O(|V|^2 - |E|)$. 

\begin{lemma}
\label{lemma:autocov}
Let a Stochastic Block Model with intra-block density $p$, inter-block density $q$, and $p > q$. Then the expected Autocovariance of intra-block pairs ($R_{intra}$) is greater than the expected Autocovariance of inter-block pairs $R_{inter}$, i.e. $\mathbb{E}[R_{intra}] > \mathbb{E}[R_{inter}]$.\end{lemma}

\begin{lemma}
\label{lemma:increasing_k}
Let a Stochastic Block Model with intra-block density $p$, inter-block density $q$, and $p > q$. Then, $\mathbb{E}[R_{intra}]$ monotonically increases as the number of partitions increases.\end{lemma}

Considering Lemma \ref{lemma:autocov} (see proof in the Appendix \ref{ap::lemma}), we argue that it is unlikely for an inter-block pair to be ranked within the top Autocovariance pairs, implying that removing these pairs from training would not affect the results. To efficiently generate a small number of hard negative pairs, we propose a novel negative sampling scheme for link prediction based on graph partitioning \citep{fortunato2010community,chiang2019cluster}. The idea is to select negative samples inside partitions (or communities) as they are expected to have similarity values comparable to positive pairs. We adopt METIS \citep{karypis1998fast} as our graph partitioning method due to its scalability and its flexibility to generate partitions of a size given as a parameter (see Appendix \ref{ap:clustering}). METIS' partitions are expected to be densely connected inside and sparsely connected across (partitions). We apply METIS to obtain $k$ partitions in which $\forall i\in\{1, 2,...,k\} : G_i = (V_i, E_i, X_i), V_i \subset V, E_i \subset E, X_i \subset X$, such that $V = \bigcup_{i=1}^k V_i$ and  $|V_i| \approx |V|/k$. Then, we apply \emph{unbiased training} only \textit{within each partition}, reducing the number of sampled negative pairs to $|E^{-}| = \sum_i^k |V_i|^2 - |E_i|$.  Following Lemma \ref{lemma:increasing_k} (see proof on Appendix \ref{ap::lemma_increasing_k}), the choice of the value of $k$ should consider a trade-off between training speed and link prediction performance (see Appendix \ref{sec::similarities_non_normalized}). Further, the algorithm proposed by \cite{newman2006modularity} could be adopted to find the optimal value of $k$ that maximizes the Modularity gain while obtaining the minimal training time. In the remainder of the paper, we refer to this approach as \textit{partitioned training}. We claim that this procedure filters (easy) pairs consisting of nodes that would be too far away in the network topology from training while maintaining the more informative (hard) pairs that are closer and topologically similar, according to METIS. We include in the Appendix \ref{sec::similarities_non_normalized} (See \autoref{fig::ap::partitioned_vs_unbiased}) a performance comparison between Gelato trained using \emph{unbiased training} against \emph{partitioned training}.

\section{Experiments}
\label{sec::experiments}

In this section, we provide empirical evidence for our claims regarding supervised link prediction and demonstrate the accuracy and efficiency of Gelato. We present ablation studies in Subsection \ref{subsec::ablation} and training time comparisons in Appendix \ref{ap::time_comparison}. Our implementation is available at Anonymous GitHub\footnote{\url{https://anonymous.4open.science/r/Gelato/}}. 

\subsection{Experiment settings}
\label{subsec::exp_setting}
\textbf{Datasets.} Our method is evaluated on four attributed graphs commonly used for link prediction \citep{chami2019hyperbolic, zhang2021lorentzian, yan2021link, zhu2021neural, chen2022bscnets, pan2021neural,hu2020open}. \autoref{tab::dataset} shows dataset statistics.

\begin{table}[htbp]
 \centering
 \small
 \caption{A summary of dataset statistics. }
   \begin{tabular}{ccccccc}
   \toprule
          & \#Nodes     & \#Edges     & \#Attrs     & Avg. degree & Density\\
   \midrule
   \textsc{Cora}  & 2,708  & 5,278  & 1,433  & 3.90 & 0.14\% \\
   \textsc{CiteSeer} & 3,327  & 4,552  & 3,703  & 2.74 & 0.08\% \\
  \textsc{PubMed} & 19,717 & 44,324 & 500   & 4.50 & 0.02\% \\
   \textsc{ogbl-ddi} & 4,267 & 1,334,889 & 0 & 500,5 & 7.33\% \\
   \textsc{ogbl-collab} & 235,868 & 1,285,465 & 128 &  8.2 & 0.0046\% \\ 
   \bottomrule
   \end{tabular}%
 \label{tab::dataset}%
\end{table}%

\noindent\textbf{Baselines.} For GNN-based link prediction, we include four state-of-the-art methods published in the past two years: Neo-GNN \citep{yun2021neo}, BUDDY \citep{chamberlain2022graph}, and NCN / NCNC \citep{wang2023neural}, as well as the pioneering work---SEAL \citep{zhang2018link}. For topological link prediction heuristics, we consider Common Neighbors (CN) \citep{newman2001clustering}, Adamic Adar (AA) \citep{adamic2003friends}, and Autocovariance (AC) \citep{random-walk-embedding}---the base heuristic in our model. 

\noindent\textbf{Hyperparameters.} For Gelato, we tune the proportion of added edges $\eta$ from \{0.0, 0.25, 0.5, 0.75, 1.0\}, the topological weight $\alpha$ from \{0.0, 0.25, 0.5, 0.75\}, and the trained weight $\beta$ from \{0.25, 0.5, 0.75, 1.0\}. We present a sensitivity analysis of all hyperparameters in Appendix \ref{ap::sensitivity_analysis}. All other settings are fixed across datasets: MLP with one hidden layer of 128 neurons, AC scaling parameter $t=3$, Adam optimizer \citep{kingma2014adam} 
with a learning rate of 0.001, a dropout rate of 0.5, and \emph{unbiased training} without downsampling. To maintain fairness in our results, we also tuned the baselines and exposed our procedures in detail in Appendix \ref{ap::setting}. For all models, including Gelato, the tuning process is done in all datasets, except for \textsc{ogbl-collab}.

\noindent\textbf{Data splits for unbiased training and unbiased testing.} 
Following \cite{kipf2016variational, zhang2018link, chami2019hyperbolic, zhang2021lorentzian, chen2022bscnets, pan2021neural}, we adopt 85\%/5\%/10\% ratios for training, validation, and testing. Specifically, for \emph{unbiased training} and \emph{unbiased testing}, we first randomly divide the (positive) edges $E$ of the original graph into $E_{train}^+$, $E_{valid}^+$, and $E_{test}^+$ for training, validation, and testing based on the selected ratios. Then, we set the negative pairs in these three sets as (1) $E_{train}^{-} = E^{-} + E_{valid}^+ + E_{test}^+$, (2) $E_{valid}^{-} = E^{-} + E_{test}^+$, and (3) $E_{test}^{-} = E^{-}$, where $E^{-}$ is the set of all negative pairs (excluding self-loops) in the original graph. Notice that the validation and testing \emph{positive} edges are included in the \emph{negative} training set, and the testing \emph{positive} edges are included in the \emph{negative} validation set. This setting simulates the real-world scenario where the test edges (and the validation edges) are unobserved during validation (training). For \emph{negative sampling}, we repeat the dividing procedure above for each generated partition $G_i$. 
The final sets are unions of individual sets for each partition: $E_{train}^{+/-} = \bigcup_{i=1}^k E_{train_i}^{+/-}$, $E_{valid}^{+/-} = \bigcup_{i=1}^k E_{valid_i}^{+/-}$, and $E_{test}^{+/-} = \bigcup_{i=1}^k E_{test_i}^{+/-}$. We notice that these splits do not leak training data to the test, as both positive and negative test pairs are disconnected during training. 


\noindent\textbf{Evaluation metrics.} 
We adopt $hits@k$ ---the ratio of positive edges individually ranked above $k$th place against all negative pairs---as our evaluation metric since it represents a good notion of class distinction under heavily imbalanced scenarios in information retrieval, compatible with the intuition of link prediction as a similarity-based ranking task.

\subsection{Partitioned Sampling and Link prediction as a similarity task}
\label{subsec::partitioned_sampling_results}

This section provides empirical evidence for some of the claims made regarding limitations in the evaluation of supervised link prediction methods (see Section \ref{sec::limitations}). It also demonstrates the effectiveness of Gelato to distinguish true links from hard negative node pairs in sparse graphs.

\begin{figure*}
    \centering
    \includegraphics[width=1\textwidth]{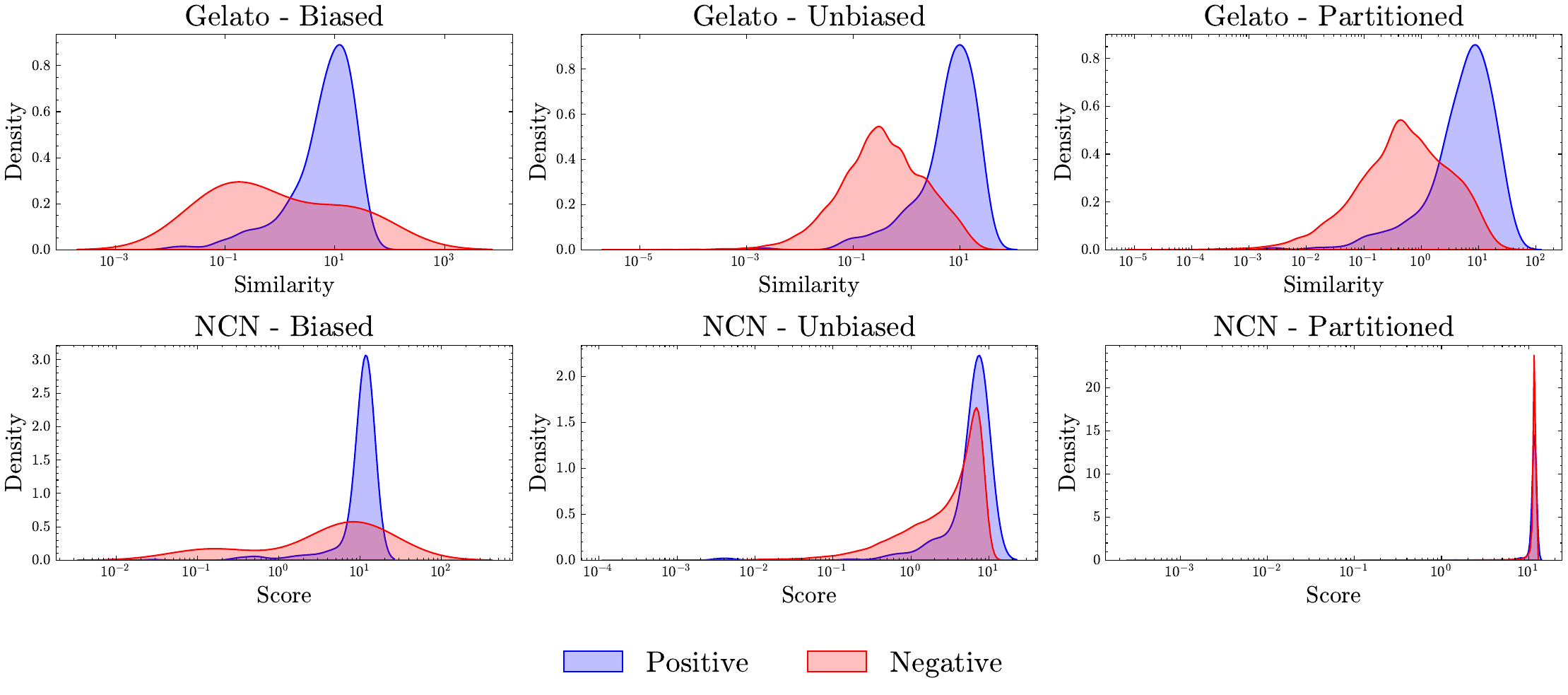}
    \caption{We analyze classification-based and similarity-based link prediction approaches through a comparison between the probability density functions of predicted similarities/scores by Gelato and NCN (state-of-the-art GNN), on the test set in three different regimes (biased, unbiased, and partitioned). Negative pairs are represented in \textcolor{red}{red}, and positive pairs are represented in \textcolor{blue}{blue}. By treating link prediction as a similarity-based problem, Gelato presents better separation (smaller overlap) between the similarity curves in the harder scenarios, distinguishing between positive and negative pairs across all testing regimes. NCN presents a drastic increase in overlap as negative pairs become harder, struggling to separate positive and negative pairs. }
    \label{fig::three_regimes}
\end{figure*}

\noindent\textbf{Negative sampling for harder pairs.} Based on the hardness of negative pairs, the easiest scenario is the \emph{biased testing}, followed by \emph{unbiased testing} and \emph{partitioned testing}---i.e. only negative pairs from inside partitions are sampled. This can be verified by \autoref{fig::three_regimes}, which compares the predicted scores of NCN against the similarities computed by Gelato on the test set of \textsc{CiteSeer}. \emph{Biased testing}, the easiest and most unrealistic scenario, shows a good separation between positive and negative pairs both in NCN and Gelato. For \emph{unbiased testing}, which is more realistic, Gelato is better at distinguishing positive and negative pairs. Finally, \emph{partitioned testing} presents a particular challenge but Gelato still ranks most positive pairs above negative ones. Other GNN-based link prediction approaches have shown similar behaviors to NCN.


\noindent\textbf{Similarity-based link prediction.} \autoref{fig::three_regimes} shows densities normalized by the size of positive and negative sets, respectively. However, in real-world sparse graphs, the number of negative pairs is much larger than that of positive ones. To better understand the ranking of positive pairs over negative pairs, we also show the same plot with non-normalized densities by the total number of all pairs in \autoref{fig::ap::mesh1} in the Appendix \ref{ap::non_normalized_plots}. 
The results show that for \emph{unbiased} and \emph{partitioned testing}, ranking positive pairs over hard negative pairs is especially challenging due to their overwhelming number, 
i.e. positive pairs are ``needles in a haystack''. 
This provides evidence that classifiers, such as GNNs for link prediction, are not suitable for finding decision boundaries in these extremely imbalanced settings, which motivates the design of Gelato as a similarity ranking model trained using an N-pair loss.


\subsection{Link prediction performance}
\label{subsec::lp_results}

\begin{figure*}

    \centering
    \includegraphics[width=1\textwidth]{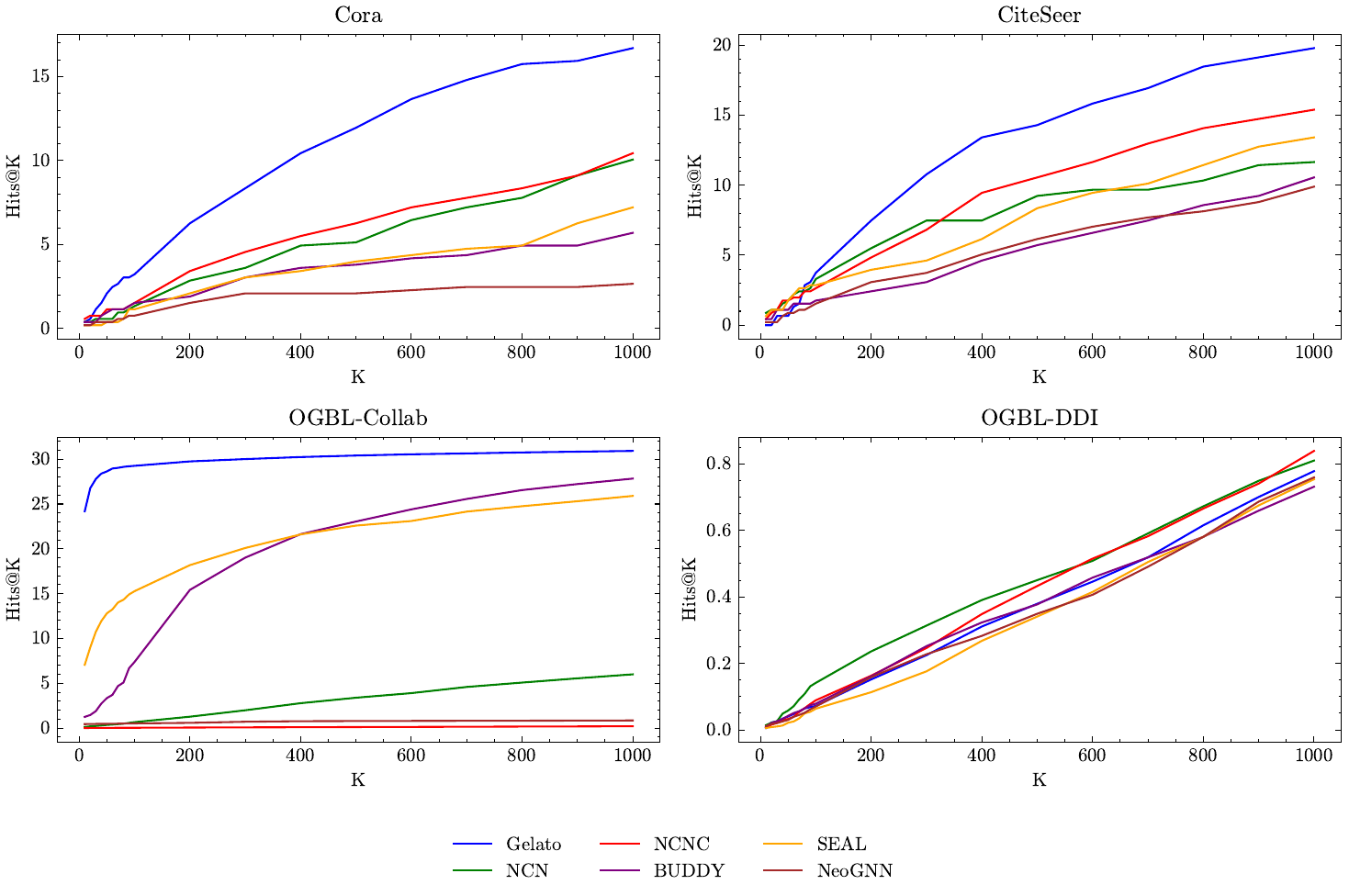}
    \caption{Link prediction comparison in terms of $hits@k$ varying $k$ using Cora, CiteSeer, OGBL-DDI and OGBL-Collab. All datasets were split using \textit{unbiased} sampling, except OGBL-Collab, which was split using \textit{partitioned} sampling. Gelato outperforms the baselines across different values of $k$ and remains competitive on OGBL-DDI, a dataset in which all methods struggle.}
    \label{fig::hits@k}
\end{figure*}

\autoref{tab::performance_hits} summarizes the link prediction performance in terms of the mean and standard deviation of $hits@1000$ for all methods. We show the same results for varying values of $k$ in Figure \ref{fig::hits@k}.  We also include the results of $MRR$ (Mean Reciprocal Rank), $AP$ (Average Precision) (see Tables \ref{tab::performance_ap} and \ref{tab::performance_mrr}) and $prec@k$ results for varying values of $k$ (see Figure \ref{fig::ap::prec@k}) in Appendix \ref{ap::prec@k}. 

\begin{table*}[tb!]
    \setlength\tabcolsep{4.65pt}
    \small
  \centering
  \caption{Link prediction performance comparison (mean ± std hits@1000) for all datasets considered. Gelato consistently outperforms GNN-based methods, topological heuristics, and two-stage approaches combining attributes/topology. For \textsc{Cora}, \textsc{CiteSeer}, \textsc{ogbl-ddi} and \textsc{PubMed} results we used \emph{unbiased} training, while for \textsc{ogbl-collab} \emph{partitioned} sampling is used, for scalability reasons. The top three models are colored by \textcolor{blue}{\textbf{First}}, \textcolor{red}{\textbf{Second}} and \textcolor{brown}{\textbf{Third}}.}
  \label{tab::performance_hits}%
  \begin{threeparttable}
  
    \begin{tabular}{ccccccc}
    \toprule
          &       & \textsc{Cora}  & \textsc{CiteSeer} & \textsc{PubMed} & \textsc{ogbl-ddi} & \textsc{ogbl-collab} \\
    \midrule
    \multirow{5}[2]{*}{GNN} 

        & SEAL & 0.0\tnote{*} & 7.25\tnote{*} & *** & 0.75\tnote{*} &  \textcolor{brown}{25.9}\tnote{*}\\
        & Neo-GNN & \textcolor{brown}{\textbf{6.96 ± 4.24}} & 5.42 ± 0.13 & \textcolor{brown}{\textbf{1.63 ± 0.32}} & 0.76\tnote{*} & 0.85\tnote{*} \\ 
        & BUDDY & 4.81 ± 0.72 & 5.86 ± 0.34 & OOM & 0.74 ± 0.01 & \textcolor{red}{\textbf{27.66 ± 0.24}} \\
        & NCN & 4.11 ± 1.22 & 7.84 ± 1.13 & 0.06 ± 0.1 & \textcolor{red}{\textbf{0.82 ± 0.02}} & 7.16 ± 1.42 \\
        & NCNC & 6.58 ± 0.58 & \textcolor{brown}{\textbf{8.72 ± 2.08}} & 1.04 ± 0.09 & \textcolor{blue}{\textbf{0.89 ± 0.09}} & 0.44 ± 0.37 \\
    \midrule
    \multicolumn{1}{c}{\multirow{2}[2]{*}{\parbox{1.8cm}{\centering Topological Heuristics}}} 
        & CN & 4.17 ± 0.00 & 4.4 ± 0.00 & 0.36 ± 0.00 & \textcolor{brown}{\textbf{0.8 ± 0.00}} & 2.4 ± 0.00 \\
        & AA & 6.64 ± 0.00 & 4.4 ± 0.00 & 1.13 ± 0.00 & 0.79 ± 0.00 & 4.88 ± 0.00 \\
        & AC & \textcolor{red}{\textbf{11.20 ± 0.00}} & \textcolor{red}{\textbf{14.29 ± 0.00}} & \textcolor{red}{\textbf{3.81 ± 0.00}} &  0.78 ± 0.00 & 12.89 ± 0.00\\
    \midrule
    \multicolumn{2}{c}{Gelato} & \textcolor{blue}{\textbf{16.62 ± 0.31}} & \textcolor{blue}{\textbf{19.78 ± 0.23}} & \textcolor{blue}{\textbf{4.18 ± 0.19}} &  0.78 ± 0.00      & \textcolor{blue}{\textbf{30.92}}\tnote{*} \\
    \bottomrule
    \end{tabular}%
	\begin{tablenotes}[para]
	\item[*] Run only once as each run takes $>$24 hrs; \hspace{5pt} *** Each run takes $>$1000 hrs; \hspace{5pt} OOM: Out Of Memory.
    \end{tablenotes}
    
  \end{threeparttable}
\end{table*}


First, we want to highlight the drastically different performance of GNN-based methods compared to those found in the original papers \citep{zhang2018link, yun2021neo,chamberlain2022graph,wang2023neural}. Some of them underperform even the simplest topological heuristics such as Common Neighbors under \emph{unbiased testing}. Moreover, Autocovariance, which is the base topological heuristic applied by Gelato and does not account for node attributes, outperforms all the GNN-based baselines for the majority of the datasets. These results support our arguments from Section \ref{sec::limitations} that evaluation metrics based on \emph{biased testing} can produce misleading results compared to \emph{unbiased testing}.

The overall best-performing GNN model is NCNC, which generalizes a pairwise topological heuristic (Common Neighbors) using message-passing. NCNC only outperforms Gelato on \textsc{OGBL-ddi}, which is consistent with previous results \cite{mao2023revisiting} showing that local structural heuristics are effective for very dense networks (see Table \ref{tab::dataset}). Moreover, \textsc{OGBL-ddi} is the only dataset considered that does not contain natural node features, which explains why our approach achieves the same performance as AC. Gelato also remains superior for different values of $hits@K$, especially for \textsc{Cora}, \textsc{CiteSeer} and \textsc{OGBL-collab}, and being remains competitive for \textsc{OGBL-ddi} being competitive as shown in Figure \ref{fig::hits@k}. This characteristic is especially relevant in real-world scenarios where robustness is desired, mainly in more conservative regimes with lower values of $k$. Overall, Gelato outperforms the best GNN-based method by \textbf{138}\%, \textbf{125}\%, \textbf{156}\%, and \textbf{11}\% for \textsc{Cora}, \textsc{Citeseer}, \textsc{Pubmed}, and \textsc{OGBL-collab}, respectively. Further, Gelato outperforms its base topological heuristic (Autocovariance) by \textbf{48}\%, \textbf{39}\%, \textbf{10}\%, and \textbf{139}\% for \textsc{Cora}, \textsc{Citeseer}, \textsc{Pubmed}, and \textsc{OGBL-collab}, respectively. Additional results are provided in Appendices \ref{ap::prec@k} and \ref{ap::biased_training}.

\subsection{Ablation study}
\label{subsec::ablation}
Here, we collect the results with the same hyperparameter setting as Gelato and present a comprehensive ablation study in \autoref{tab::ablation}. Specifically, \emph{Gelato$-$MLP} (\emph{AC}) represents Gelato without the MLP (Autocovariance) component, i.e., only using Autocovariance (MLP) for link prediction. 
\emph{Gelato$-$NP} (\emph{UT}) replaces the proposed N-pair loss (\emph{unbiased training}) with the cross entropy loss (\emph{biased training}) applied by the baselines. Finally, \emph{Gelato$-$NP+UT} replaces both the loss and the training setting. 


\begin{table}[htbp]
    \setlength\tabcolsep{7pt}
  \centering
  \caption{Results of the ablation study based on hits@1000 scores. Each component of Gelato plays an important role in enabling state-of-the-art link prediction performance. }
  \begin{threeparttable}
    \begin{tabular}{lccccc}
    \toprule
          & \textsc{Cora}  & \textsc{CiteSeer} & \textsc{PubMed}
          \\
    \midrule
        \emph{Gelato$-$MLP} & 16.13 ± 0.00 & 19.78 ± 0.00 & 3.81 ± 0.0
        \\
        \emph{Gelato$-$AC} & 2.66 ± 2.57 & 12.6 ± 0.71 & 0.0 ± 0.0
        \\
        \emph{Gelato$-$NP+UT} & 16.32 ± 0.19 & 19.41 ± 0.34 & 4.05 ± 0.12\\
        \emph{Gelato$-$NP} & 16.51 ± 0.19 & 17.88 ± 0.46 & 1.74 ± 0.14\\
        \emph{Gelato} & \textbf{16.62 ± 0.31} & \textbf{19.89 ± 0.23} & \textbf{4.18 ± 0.19}\\
    \bottomrule
    \end{tabular}%
  \end{threeparttable}
  \label{tab::ablation}%
\end{table}

We observe that removing either MLP or Autocovariance leads to inferior performance, as the corresponding attribute or topology information would be missing. Further, to address the class imbalance problem of link prediction, both the N-pair loss and \emph{unbiased training} are crucial for the effective training of Gelato.

We also present results for Gelato using different ranking-based loss functions. In particular, we choose between Precision@k, pairwise hinge, pairwise exponential, and pairwise logistic losses as candidates for replacing the N-pair loss based on \cite{chen2009ranking}. The results are shown in \autoref{tab::losses}, demonstrating that there is no clear winner considering the $hits@1000$ metric in the two datasets used  (\textsc{Cora} and \textsc{CiteSeer}).

\begin{table}[htbp]
    \setlength\tabcolsep{7pt}
  \centering
  \caption{Comparison between N-pair loss (Gelato) against the Precision@K (PK), pairwise hinge (PH), pairwise exponential (PE), and pairwise logistic (PL) losses considering the $hits@1000$ metric. }
  \begin{threeparttable}
    \begin{tabular}{lccccc}
    \toprule
          & \textsc{Cora}  & \textsc{CiteSeer}
          \\
    \midrule
        \textit{Gelato-PK} & 16.32 ± 0.19 & 19.19 ± 0.99 \\
        \textit{Gelato-PH} & \textbf{18.09 ± 0.48} & 16.56 ± 0.13 \\
        \textit{Gelato-PE} & 16.82 ± 0.48 & 15.9 ± 0.34 \\
        \textit{Gelato-PL} & 18.03 ± 0.38 & 17.14 ± 0.66 \\
        \textit{Gelato} & 16.62 ± 0.31 & \textbf{19.89 ± 0.24} \\
    \bottomrule
    \end{tabular}%
  \end{threeparttable}
  \label{tab::losses}%
\end{table}

\section{Related work}
\label{sec::relatedwork}

\textbf{Topological heuristics for link prediction.} The early link prediction literature focuses on topology-based heuristics. This includes approaches based on local (e.g., Common Neighbors \citep{newman2001clustering}, Adamic Adar \citep{adamic2003friends}, and Resource Allocation \citep{zhou2009predicting}) and higher-order (e.g., Katz \citep{katz1953new}, PageRank \citep{page1999pagerank}, and SimRank \citep{jeh2002simrank}) information. More recently, random-walk based graph embedding methods, which learn vector representations for nodes \citep{perozzi2014deepwalk, grover2016node2vec, random-walk-embedding}, have achieved promising results in graph machine learning tasks. Popular embedding approaches, such as DeepWalk \citep{perozzi2014deepwalk} and node2vec \citep{grover2016node2vec}, have been shown to implicitly approximate the Pointwise Mutual Information similarity \citep{qiu2018network}, which can also be used as a link prediction heuristic. This has motivated the investigation of other similarity metrics 
such as Autocovariance \citep{delvenne2010stability,random-walk-embedding, pole}. However, these heuristics are unsupervised and cannot take advantage of data beyond the topology. \looseness=-1

\noindent\textbf{Graph Neural Networks for link prediction.} GNN-based link prediction addresses the limitations of topological heuristics by training a neural network to combine topological and attribute information and potentially learn new heuristics. These works often assume that links are correlated with homophily in node attributes \cite{di2024link,zhou2022link}, as also is the case for this paper. GAE \citep{kipf2016variational} combines a graph convolution network \citep{kipf2016semi} and an inner product decoder based on node embeddings. SEAL \citep{zhang2018link} models link prediction as a binary subgraph classification problem (edge/non-edge), and follow-up work (e.g., SHHF \citep{liu2020feature}, WalkPool \citep{pan2021neural}) investigates different pooling strategies.
Other recent approaches for GNN-based link prediction include learning representations in hyperbolic space (e.g., HGCN \citep{chami2019hyperbolic}, LGCN \citep{zhang2021lorentzian}), generalizing topological heuristics (e.g., Neo-GNN \citep{yun2021neo}, NBFNet \citep{zhu2021neural}), and incorporating additional topological features (e.g., TLC-GNN \citep{yan2021link}, BScNets \citep{chen2022bscnets}). ELPH and BUDDY \citep{chamberlain2022graph} apply hashing to efficiently approximate subgraph-based link prediction models, such as SEAL, using a message-passing neural network (MPNN) with distance-based structural features. NCNC \citep{wang2023neural} combines the Common Neighbors heuristic with an MPNN achieving state-of-the-art results. Motivated by the growing popularity of GNNs for link prediction, this work investigates key questions regarding their training, evaluation, and ability to learn effective topological heuristics directly from data. We propose Gelato, which is simpler, more accurate, and faster than most state-of-the-art GNN-based link prediction methods. 


\noindent\textbf{Graph learning.} Gelato learns a graph that combines topological and attribute information. Our goal differs from generative models \citep{you2018graphrnn, li2018learning, grover2019graphite}, which learn to sample from a distribution over graphs. 
Graph learning also enables the application of GNNs when the graph is unavailable, noisy, or incomplete \citep{zhao2022graph}. LDS \citep{franceschi2019learning} and GAug \citep{zhao2021data} jointly learn a probability distribution over edges and GNN parameters. 
IDGL \citep{chen2020iterative} and EGLN \citep{yang2021enhanced} alternate between optimizing the graph and embeddings for node/graph classification and collaborative filtering. 
\cite{singh2021edge} proposes two-stage link prediction by augmenting the graph as a preprocessing step. In comparison, Gelato effectively 
learns a graph in an end-to-end manner by minimizing the loss of a topological heuristic. \looseness=-1

\section{Conclusion}
\label{sec::conclusion}



This work exposes key limitations in evaluating supervised link prediction methods due to the widespread use of \emph{biased testing}. These limitations led to a consensus in the graph machine learning community that (1) GNNs are superior for link prediction, casting topological heuristics obsolete; and (2) link prediction is now an easy task due to deep learning advances. We challenge both assumptions, demonstrating that link prediction in sparse graphs remains a hard problem when evaluated properly. GNNs struggle with link prediction in sparse graphs due to extreme class imbalance, motivating Gelato, our novel link prediction framework.

Gelato is a similarity-based method that combines graph learning and autocovariance to leverage attribute and topological information. Gelato employs an N-pair loss instead of cross-entropy to address the class imbalance and introduces a partitioning-based negative sampling scheme for efficient hard negative pair sampling. Through extensive experiments, we demonstrate superior accuracy and scalability of Gelato when compared to state-of-the-art GNN-based solutions across various datasets.

\bibliographystyle{ACM-Reference-Format}
\bibliography{main}


\begin{thebibliography}{105}


\ifx \showCODEN    \undefined \def \showCODEN     #1{\unskip}     \fi
\ifx \showDOI      \undefined \def \showDOI       #1{#1}\fi
\ifx \showISBNx    \undefined \def \showISBNx     #1{\unskip}     \fi
\ifx \showISBNxiii \undefined \def \showISBNxiii  #1{\unskip}     \fi
\ifx \showISSN     \undefined \def \showISSN      #1{\unskip}     \fi
\ifx \showLCCN     \undefined \def \showLCCN      #1{\unskip}     \fi
\ifx \shownote     \undefined \def \shownote      #1{#1}          \fi
\ifx \showarticletitle \undefined \def \showarticletitle #1{#1}   \fi
\ifx \showURL      \undefined \def \showURL       {\relax}        \fi
\providecommand\bibfield[2]{#2}
\providecommand\bibinfo[2]{#2}
\providecommand\natexlab[1]{#1}
\providecommand\showeprint[2][]{arXiv:#2}

\bibitem[Adamic and Adar(2003)]%
        {adamic2003friends}
\bibfield{author}{\bibinfo{person}{Lada~A Adamic} {and} \bibinfo{person}{Eytan Adar}.} \bibinfo{year}{2003}\natexlab{}.
\newblock \showarticletitle{Friends and neighbors on the web}.
\newblock \bibinfo{journal}{\emph{Social networks}} \bibinfo{volume}{25}, \bibinfo{number}{3} (\bibinfo{year}{2003}), \bibinfo{pages}{211--230}.
\newblock


\bibitem[Bahulkar et~al\mbox{.}(2018)]%
        {bahulkar2018community}
\bibfield{author}{\bibinfo{person}{Ashwin Bahulkar}, \bibinfo{person}{Boleslaw~K Szymanski}, \bibinfo{person}{N~Orkun Baycik}, {and} \bibinfo{person}{Thomas~C Sharkey}.} \bibinfo{year}{2018}\natexlab{}.
\newblock \showarticletitle{Community detection with edge augmentation in criminal networks}. In \bibinfo{booktitle}{\emph{ASONAM}}.
\newblock


\bibitem[Barab{\'a}si(2016)]%
        {barabsi2016network}
\bibfield{author}{\bibinfo{person}{Albert-L{\'a}szl{\'o} Barab{\'a}si}.} \bibinfo{year}{2016}\natexlab{}.
\newblock \bibinfo{booktitle}{\emph{Network Science}}.
\newblock \bibinfo{publisher}{Cambridge University Press}.
\newblock


\bibitem[Barab{\'a}si et~al\mbox{.}(2002)]%
        {barabasi2002evolution}
\bibfield{author}{\bibinfo{person}{Albert-Laszlo Barab{\'a}si}, \bibinfo{person}{Hawoong Jeong}, \bibinfo{person}{Zoltan N{\'e}da}, \bibinfo{person}{Erzsebet Ravasz}, \bibinfo{person}{Andras Schubert}, {and} \bibinfo{person}{Tamas Vicsek}.} \bibinfo{year}{2002}\natexlab{}.
\newblock \showarticletitle{Evolution of the social network of scientific collaborations}.
\newblock \bibinfo{journal}{\emph{Physica A: Statistical mechanics and its applications}} \bibinfo{volume}{311}, \bibinfo{number}{3-4} (\bibinfo{year}{2002}), \bibinfo{pages}{590--614}.
\newblock


\bibitem[Bordes et~al\mbox{.}(2013)]%
        {bordes2013translating}
\bibfield{author}{\bibinfo{person}{Antoine Bordes}, \bibinfo{person}{Nicolas Usunier}, \bibinfo{person}{Alberto Garc{\'\i}a-Dur{\'a}n}, \bibinfo{person}{Jason Weston}, {and} \bibinfo{person}{Oksana Yakhnenko}.} \bibinfo{year}{2013}\natexlab{}.
\newblock \showarticletitle{Translating Embeddings for Modeling Multi-relational Data}. In \bibinfo{booktitle}{\emph{NeurIPS}}.
\newblock


\bibitem[Bruch(2021)]%
        {bruch2021alternative}
\bibfield{author}{\bibinfo{person}{Sebastian Bruch}.} \bibinfo{year}{2021}\natexlab{}.
\newblock \showarticletitle{An alternative cross entropy loss for learning-to-rank}. In \bibinfo{booktitle}{\emph{WebConf}}.
\newblock


\bibitem[Byrd and Lipton(2019)]%
        {byrd2019effect}
\bibfield{author}{\bibinfo{person}{Jonathon Byrd} {and} \bibinfo{person}{Zachary Lipton}.} \bibinfo{year}{2019}\natexlab{}.
\newblock \showarticletitle{What is the effect of importance weighting in deep learning?}. In \bibinfo{booktitle}{\emph{International conference on machine learning}}. PMLR, \bibinfo{pages}{872--881}.
\newblock


\bibitem[Cai et~al\mbox{.}(2021)]%
        {cai2021line}
\bibfield{author}{\bibinfo{person}{Lei Cai}, \bibinfo{person}{Jundong Li}, \bibinfo{person}{Jie Wang}, {and} \bibinfo{person}{Shuiwang Ji}.} \bibinfo{year}{2021}\natexlab{}.
\newblock \showarticletitle{Line graph neural networks for link prediction}.
\newblock \bibinfo{journal}{\emph{IEEE TPAMI}} (\bibinfo{year}{2021}).
\newblock


\bibitem[Cakir et~al\mbox{.}(2019)]%
        {cakir2019deep}
\bibfield{author}{\bibinfo{person}{Fatih Cakir}, \bibinfo{person}{Kun He}, \bibinfo{person}{Xide Xia}, \bibinfo{person}{Brian Kulis}, {and} \bibinfo{person}{Stan Sclaroff}.} \bibinfo{year}{2019}\natexlab{}.
\newblock \showarticletitle{Deep metric learning to rank}. In \bibinfo{booktitle}{\emph{CVPR}}.
\newblock


\bibitem[Chamberlain et~al\mbox{.}(2023)]%
        {chamberlain2022graph}
\bibfield{author}{\bibinfo{person}{Benjamin~Paul Chamberlain}, \bibinfo{person}{Sergey Shirobokov}, \bibinfo{person}{Emanuele Rossi}, \bibinfo{person}{Fabrizio Frasca}, \bibinfo{person}{Thomas Markovich}, \bibinfo{person}{Nils Hammerla}, \bibinfo{person}{Michael~M Bronstein}, {and} \bibinfo{person}{Max Hansmire}.} \bibinfo{year}{2023}\natexlab{}.
\newblock \showarticletitle{Graph Neural Networks for Link Prediction with Subgraph Sketching}. In \bibinfo{booktitle}{\emph{ICLR}}.
\newblock


\bibitem[Chami et~al\mbox{.}(2019)]%
        {chami2019hyperbolic}
\bibfield{author}{\bibinfo{person}{Ines Chami}, \bibinfo{person}{Zhitao Ying}, \bibinfo{person}{Christopher R{\'e}}, {and} \bibinfo{person}{Jure Leskovec}.} \bibinfo{year}{2019}\natexlab{}.
\newblock \showarticletitle{Hyperbolic graph convolutional neural networks}. In \bibinfo{booktitle}{\emph{NeurIPS}}.
\newblock


\bibitem[Chen et~al\mbox{.}(2009)]%
        {chen2009ranking}
\bibfield{author}{\bibinfo{person}{Wei Chen}, \bibinfo{person}{Tie-Yan Liu}, \bibinfo{person}{Yanyan Lan}, \bibinfo{person}{Zhi-Ming Ma}, {and} \bibinfo{person}{Hang Li}.} \bibinfo{year}{2009}\natexlab{}.
\newblock \showarticletitle{Ranking measures and loss functions in learning to rank}.
\newblock \bibinfo{journal}{\emph{Advances in Neural Information Processing Systems}}  \bibinfo{volume}{22} (\bibinfo{year}{2009}).
\newblock


\bibitem[Chen et~al\mbox{.}(2014)]%
        {chen2014unsupervised}
\bibfield{author}{\bibinfo{person}{Xu Chen}, \bibinfo{person}{Xiuyuan Cheng}, {and} \bibinfo{person}{St{\'e}phane Mallat}.} \bibinfo{year}{2014}\natexlab{}.
\newblock \showarticletitle{Unsupervised deep haar scattering on graphs}. In \bibinfo{booktitle}{\emph{NeurIPS}}.
\newblock


\bibitem[Chen et~al\mbox{.}(2022)]%
        {chen2022bscnets}
\bibfield{author}{\bibinfo{person}{Yuzhou Chen}, \bibinfo{person}{Yulia~R Gel}, {and} \bibinfo{person}{H~Vincent Poor}.} \bibinfo{year}{2022}\natexlab{}.
\newblock \showarticletitle{BScNets: Block Simplicial Complex Neural Networks}. In \bibinfo{booktitle}{\emph{AAAI}}.
\newblock


\bibitem[Chen et~al\mbox{.}(2020b)]%
        {chen2020iterative}
\bibfield{author}{\bibinfo{person}{Yu Chen}, \bibinfo{person}{Lingfei Wu}, {and} \bibinfo{person}{Mohammed Zaki}.} \bibinfo{year}{2020}\natexlab{b}.
\newblock \showarticletitle{Iterative deep graph learning for graph neural networks: Better and robust node embeddings}. In \bibinfo{booktitle}{\emph{NeurIPS}}.
\newblock


\bibitem[Chen et~al\mbox{.}(2020a)]%
        {chen2020can}
\bibfield{author}{\bibinfo{person}{Zhengdao Chen}, \bibinfo{person}{Lei Chen}, \bibinfo{person}{Soledad Villar}, {and} \bibinfo{person}{Joan Bruna}.} \bibinfo{year}{2020}\natexlab{a}.
\newblock \showarticletitle{Can graph neural networks count substructures?}
\newblock \bibinfo{journal}{\emph{Advances in neural information processing systems}}  \bibinfo{volume}{33} (\bibinfo{year}{2020}), \bibinfo{pages}{10383--10395}.
\newblock


\bibitem[Chiang et~al\mbox{.}(2019)]%
        {chiang2019cluster}
\bibfield{author}{\bibinfo{person}{Wei-Lin Chiang}, \bibinfo{person}{Xuanqing Liu}, \bibinfo{person}{Si Si}, \bibinfo{person}{Yang Li}, \bibinfo{person}{Samy Bengio}, {and} \bibinfo{person}{Cho-Jui Hsieh}.} \bibinfo{year}{2019}\natexlab{}.
\newblock \showarticletitle{Cluster-gcn: An efficient algorithm for training deep and large graph convolutional networks}. In \bibinfo{booktitle}{\emph{Proceedings of the 25th ACM SIGKDD international conference on knowledge discovery \& data mining}}. \bibinfo{pages}{257--266}.
\newblock


\bibitem[da~Silva et~al\mbox{.}(2020)]%
        {da2020combining}
\bibfield{author}{\bibinfo{person}{Arlei~Lopes da Silva}, \bibinfo{person}{Furkan Kocayusufoglu}, \bibinfo{person}{Saber Jafarpour}, \bibinfo{person}{Francesco Bullo}, \bibinfo{person}{Ananthram Swami}, {and} \bibinfo{person}{Ambuj Singh}.} \bibinfo{year}{2020}\natexlab{}.
\newblock \showarticletitle{Combining Physics and Machine Learning for Network Flow Estimation}. In \bibinfo{booktitle}{\emph{ICLR}}.
\newblock


\bibitem[Davis and Goadrich(2006)]%
        {davis2006relationship}
\bibfield{author}{\bibinfo{person}{Jesse Davis} {and} \bibinfo{person}{Mark Goadrich}.} \bibinfo{year}{2006}\natexlab{}.
\newblock \showarticletitle{The relationship between Precision-Recall and ROC curves}. In \bibinfo{booktitle}{\emph{ICML}}.
\newblock


\bibitem[Delvenne et~al\mbox{.}(2010)]%
        {delvenne2010stability}
\bibfield{author}{\bibinfo{person}{J-C Delvenne}, \bibinfo{person}{Sophia~N Yaliraki}, {and} \bibinfo{person}{Mauricio Barahona}.} \bibinfo{year}{2010}\natexlab{}.
\newblock \showarticletitle{Stability of graph communities across time scales}.
\newblock \bibinfo{journal}{\emph{PNAS}} \bibinfo{volume}{107}, \bibinfo{number}{29} (\bibinfo{year}{2010}), \bibinfo{pages}{12755--12760}.
\newblock


\bibitem[Di~Francesco et~al\mbox{.}(2024)]%
        {di2024link}
\bibfield{author}{\bibinfo{person}{Andrea~Giuseppe Di~Francesco}, \bibinfo{person}{Francesco Caso}, \bibinfo{person}{Maria~Sofia Bucarelli}, {and} \bibinfo{person}{Fabrizio Silvestri}.} \bibinfo{year}{2024}\natexlab{}.
\newblock \showarticletitle{Link Prediction under Heterophily: A Physics-Inspired Graph Neural Network Approach}.
\newblock \bibinfo{journal}{\emph{arXiv preprint arXiv:2402.14802}} (\bibinfo{year}{2024}).
\newblock


\bibitem[Fortunato(2010)]%
        {fortunato2010community}
\bibfield{author}{\bibinfo{person}{Santo Fortunato}.} \bibinfo{year}{2010}\natexlab{}.
\newblock \showarticletitle{Community detection in graphs}.
\newblock \bibinfo{journal}{\emph{Physics reports}} \bibinfo{volume}{486}, \bibinfo{number}{3-5} (\bibinfo{year}{2010}), \bibinfo{pages}{75--174}.
\newblock


\bibitem[Franceschi et~al\mbox{.}(2019)]%
        {franceschi2019learning}
\bibfield{author}{\bibinfo{person}{Luca Franceschi}, \bibinfo{person}{Mathias Niepert}, \bibinfo{person}{Massimiliano Pontil}, {and} \bibinfo{person}{Xiao He}.} \bibinfo{year}{2019}\natexlab{}.
\newblock \showarticletitle{Learning discrete structures for graph neural networks}. In \bibinfo{booktitle}{\emph{ICML}}.
\newblock


\bibitem[Freund et~al\mbox{.}(2003)]%
        {freund2003efficient}
\bibfield{author}{\bibinfo{person}{Yoav Freund}, \bibinfo{person}{Raj Iyer}, \bibinfo{person}{Robert~E Schapire}, {and} \bibinfo{person}{Yoram Singer}.} \bibinfo{year}{2003}\natexlab{}.
\newblock \showarticletitle{An efficient boosting algorithm for combining preferences}.
\newblock \bibinfo{journal}{\emph{JMLR}} \bibinfo{volume}{4}, \bibinfo{number}{Nov} (\bibinfo{year}{2003}), \bibinfo{pages}{933--969}.
\newblock


\bibitem[Grover and Leskovec(2016)]%
        {grover2016node2vec}
\bibfield{author}{\bibinfo{person}{Aditya Grover} {and} \bibinfo{person}{Jure Leskovec}.} \bibinfo{year}{2016}\natexlab{}.
\newblock \showarticletitle{node2vec: Scalable feature learning for networks}. In \bibinfo{booktitle}{\emph{SIGKDD}}.
\newblock


\bibitem[Grover et~al\mbox{.}(2019)]%
        {grover2019graphite}
\bibfield{author}{\bibinfo{person}{Aditya Grover}, \bibinfo{person}{Aaron Zweig}, {and} \bibinfo{person}{Stefano Ermon}.} \bibinfo{year}{2019}\natexlab{}.
\newblock \showarticletitle{Graphite: Iterative generative modeling of graphs}. In \bibinfo{booktitle}{\emph{ICML}}.
\newblock


\bibitem[Hamilton et~al\mbox{.}(2017)]%
        {hamilton2017inductive}
\bibfield{author}{\bibinfo{person}{Will Hamilton}, \bibinfo{person}{Zhitao Ying}, {and} \bibinfo{person}{Jure Leskovec}.} \bibinfo{year}{2017}\natexlab{}.
\newblock \showarticletitle{Inductive representation learning on large graphs}. In \bibinfo{booktitle}{\emph{NeurIPS}}.
\newblock


\bibitem[Hu et~al\mbox{.}(2020)]%
        {hu2020open}
\bibfield{author}{\bibinfo{person}{Weihua Hu}, \bibinfo{person}{Matthias Fey}, \bibinfo{person}{Marinka Zitnik}, \bibinfo{person}{Yuxiao Dong}, \bibinfo{person}{Hongyu Ren}, \bibinfo{person}{Bowen Liu}, \bibinfo{person}{Michele Catasta}, {and} \bibinfo{person}{Jure Leskovec}.} \bibinfo{year}{2020}\natexlab{}.
\newblock \showarticletitle{Open graph benchmark: Datasets for machine learning on graphs}. In \bibinfo{booktitle}{\emph{NeurIPS}}.
\newblock


\bibitem[Hu et~al\mbox{.}(2022)]%
        {hu2022two}
\bibfield{author}{\bibinfo{person}{Yang Hu}, \bibinfo{person}{Xiyuan Wang}, \bibinfo{person}{Zhouchen Lin}, \bibinfo{person}{Pan Li}, {and} \bibinfo{person}{Muhan Zhang}.} \bibinfo{year}{2022}\natexlab{}.
\newblock \showarticletitle{Two-Dimensional Weisfeiler-Lehman Graph Neural Networks for Link Prediction}.
\newblock \bibinfo{journal}{\emph{arXiv preprint arXiv:2206.09567}} (\bibinfo{year}{2022}).
\newblock


\bibitem[Huang et~al\mbox{.}(2021)]%
        {random-walk-embedding}
\bibfield{author}{\bibinfo{person}{Zexi Huang}, \bibinfo{person}{Arlei Silva}, {and} \bibinfo{person}{Ambuj Singh}.} \bibinfo{year}{2021}\natexlab{}.
\newblock \showarticletitle{A Broader Picture of Random-walk Based Graph Embedding}. In \bibinfo{booktitle}{\emph{SIGKDD}}.
\newblock


\bibitem[Huang et~al\mbox{.}(2022)]%
        {pole}
\bibfield{author}{\bibinfo{person}{Zexi Huang}, \bibinfo{person}{Arlei Silva}, {and} \bibinfo{person}{Ambuj Singh}.} \bibinfo{year}{2022}\natexlab{}.
\newblock \showarticletitle{POLE: Polarized Embedding for Signed Networks}. In \bibinfo{booktitle}{\emph{WSDM}}.
\newblock


\bibitem[Ivanovic and Pavone(2019)]%
        {ivanovic2019trajectron}
\bibfield{author}{\bibinfo{person}{Boris Ivanovic} {and} \bibinfo{person}{Marco Pavone}.} \bibinfo{year}{2019}\natexlab{}.
\newblock \showarticletitle{The trajectron: Probabilistic multi-agent trajectory modeling with dynamic spatiotemporal graphs}. In \bibinfo{booktitle}{\emph{ICCV}}.
\newblock


\bibitem[Jamali and Ester(2009)]%
        {jamali2009trustwalker}
\bibfield{author}{\bibinfo{person}{Mohsen Jamali} {and} \bibinfo{person}{Martin Ester}.} \bibinfo{year}{2009}\natexlab{}.
\newblock \showarticletitle{Trustwalker: a random walk model for combining trust-based and item-based recommendation}. In \bibinfo{booktitle}{\emph{SIGKDD}}.
\newblock


\bibitem[Jeh and Widom(2002)]%
        {jeh2002simrank}
\bibfield{author}{\bibinfo{person}{Glen Jeh} {and} \bibinfo{person}{Jennifer Widom}.} \bibinfo{year}{2002}\natexlab{}.
\newblock \showarticletitle{Simrank: a measure of structural-context similarity}. In \bibinfo{booktitle}{\emph{SIGKDD}}.
\newblock


\bibitem[Karrer and Newman(2011)]%
        {karrer2011stochastic}
\bibfield{author}{\bibinfo{person}{Brian Karrer} {and} \bibinfo{person}{Mark~EJ Newman}.} \bibinfo{year}{2011}\natexlab{}.
\newblock \showarticletitle{Stochastic blockmodels and community structure in networks}.
\newblock \bibinfo{journal}{\emph{PRE}} \bibinfo{volume}{83}, \bibinfo{number}{1} (\bibinfo{year}{2011}), \bibinfo{pages}{016107}.
\newblock


\bibitem[Karypis and Kumar(1998)]%
        {karypis1998fast}
\bibfield{author}{\bibinfo{person}{George Karypis} {and} \bibinfo{person}{Vipin Kumar}.} \bibinfo{year}{1998}\natexlab{}.
\newblock \showarticletitle{A fast and high quality multilevel scheme for partitioning irregular graphs}.
\newblock \bibinfo{journal}{\emph{SIAM Journal on scientific Computing}} \bibinfo{volume}{20}, \bibinfo{number}{1} (\bibinfo{year}{1998}), \bibinfo{pages}{359--392}.
\newblock


\bibitem[Katz(1953)]%
        {katz1953new}
\bibfield{author}{\bibinfo{person}{Leo Katz}.} \bibinfo{year}{1953}\natexlab{}.
\newblock \showarticletitle{A new status index derived from sociometric analysis}.
\newblock \bibinfo{journal}{\emph{Psychometrika}} \bibinfo{volume}{18}, \bibinfo{number}{1} (\bibinfo{year}{1953}), \bibinfo{pages}{39--43}.
\newblock


\bibitem[Kingma and Ba(2015)]%
        {kingma2014adam}
\bibfield{author}{\bibinfo{person}{Diederik~P Kingma} {and} \bibinfo{person}{Jimmy Ba}.} \bibinfo{year}{2015}\natexlab{}.
\newblock \showarticletitle{Adam: A method for stochastic optimization}. In \bibinfo{booktitle}{\emph{ICLR}}.
\newblock


\bibitem[Kipf and Welling(2016)]%
        {kipf2016variational}
\bibfield{author}{\bibinfo{person}{Thomas~N Kipf} {and} \bibinfo{person}{Max Welling}.} \bibinfo{year}{2016}\natexlab{}.
\newblock \showarticletitle{Variational graph auto-encoders}.
\newblock \bibinfo{journal}{\emph{arXiv preprint arXiv:1611.07308}} (\bibinfo{year}{2016}).
\newblock


\bibitem[Kipf and Welling(2017)]%
        {kipf2016semi}
\bibfield{author}{\bibinfo{person}{Thomas~N Kipf} {and} \bibinfo{person}{Max Welling}.} \bibinfo{year}{2017}\natexlab{}.
\newblock \showarticletitle{Semi-supervised classification with graph convolutional networks}. In \bibinfo{booktitle}{\emph{ICLR}}.
\newblock


\bibitem[Klicpera et~al\mbox{.}(2018)]%
        {klicpera2018predict}
\bibfield{author}{\bibinfo{person}{Johannes Klicpera}, \bibinfo{person}{Aleksandar Bojchevski}, {and} \bibinfo{person}{Stephan G{\"u}nnemann}.} \bibinfo{year}{2018}\natexlab{}.
\newblock \showarticletitle{Predict then Propagate: Graph Neural Networks meet Personalized PageRank}. In \bibinfo{booktitle}{\emph{ICLR}}.
\newblock


\bibitem[Li et~al\mbox{.}(2017)]%
        {li2017deepcas}
\bibfield{author}{\bibinfo{person}{Cheng Li}, \bibinfo{person}{Jiaqi Ma}, \bibinfo{person}{Xiaoxiao Guo}, {and} \bibinfo{person}{Qiaozhu Mei}.} \bibinfo{year}{2017}\natexlab{}.
\newblock \showarticletitle{Deepcas: An end-to-end predictor of information cascades}. In \bibinfo{booktitle}{\emph{WebConf}}.
\newblock


\bibitem[Li et~al\mbox{.}(2024)]%
        {li2024evaluating}
\bibfield{author}{\bibinfo{person}{Juanhui Li}, \bibinfo{person}{Harry Shomer}, \bibinfo{person}{Haitao Mao}, \bibinfo{person}{Shenglai Zeng}, \bibinfo{person}{Yao Ma}, \bibinfo{person}{Neil Shah}, \bibinfo{person}{Jiliang Tang}, {and} \bibinfo{person}{Dawei Yin}.} \bibinfo{year}{2024}\natexlab{}.
\newblock \showarticletitle{Evaluating graph neural networks for link prediction: Current pitfalls and new benchmarking}.
\newblock \bibinfo{journal}{\emph{Advances in Neural Information Processing Systems}}  \bibinfo{volume}{36} (\bibinfo{year}{2024}).
\newblock


\bibitem[Li et~al\mbox{.}(2020)]%
        {li2020distance}
\bibfield{author}{\bibinfo{person}{Pan Li}, \bibinfo{person}{Yanbang Wang}, \bibinfo{person}{Hongwei Wang}, {and} \bibinfo{person}{Jure Leskovec}.} \bibinfo{year}{2020}\natexlab{}.
\newblock \showarticletitle{Distance encoding: Design provably more powerful neural networks for graph representation learning}.
\newblock \bibinfo{journal}{\emph{Advances in Neural Information Processing Systems}}  \bibinfo{volume}{33} (\bibinfo{year}{2020}), \bibinfo{pages}{4465--4478}.
\newblock


\bibitem[Li et~al\mbox{.}(2018)]%
        {li2018learning}
\bibfield{author}{\bibinfo{person}{Yujia Li}, \bibinfo{person}{Oriol Vinyals}, \bibinfo{person}{Chris Dyer}, \bibinfo{person}{Razvan Pascanu}, {and} \bibinfo{person}{Peter Battaglia}.} \bibinfo{year}{2018}\natexlab{}.
\newblock \showarticletitle{Learning deep generative models of graphs}. In \bibinfo{booktitle}{\emph{ICML}}.
\newblock


\bibitem[Liben-Nowell and Kleinberg(2007)]%
        {liben2007link}
\bibfield{author}{\bibinfo{person}{David Liben-Nowell} {and} \bibinfo{person}{Jon Kleinberg}.} \bibinfo{year}{2007}\natexlab{}.
\newblock \showarticletitle{The link-prediction problem for social networks}.
\newblock \bibinfo{journal}{\emph{Journal of the American society for information science and technology}} \bibinfo{volume}{58}, \bibinfo{number}{7} (\bibinfo{year}{2007}), \bibinfo{pages}{1019--1031}.
\newblock


\bibitem[Liu et~al\mbox{.}(2020)]%
        {liu2020feature}
\bibfield{author}{\bibinfo{person}{Zheyi Liu}, \bibinfo{person}{Darong Lai}, \bibinfo{person}{Chuanyou Li}, {and} \bibinfo{person}{Meng Wang}.} \bibinfo{year}{2020}\natexlab{}.
\newblock \showarticletitle{Feature Fusion Based Subgraph Classification for Link Prediction}. In \bibinfo{booktitle}{\emph{CIKM}}.
\newblock


\bibitem[L{\"u} and Zhou(2011)]%
        {lu2011link}
\bibfield{author}{\bibinfo{person}{Linyuan L{\"u}} {and} \bibinfo{person}{Tao Zhou}.} \bibinfo{year}{2011}\natexlab{}.
\newblock \showarticletitle{Link prediction in complex networks: A survey}.
\newblock \bibinfo{journal}{\emph{Physica A: statistical mechanics and its applications}} \bibinfo{volume}{390}, \bibinfo{number}{6} (\bibinfo{year}{2011}), \bibinfo{pages}{1150--1170}.
\newblock


\bibitem[Mao et~al\mbox{.}(2023)]%
        {mao2023revisiting}
\bibfield{author}{\bibinfo{person}{Haitao Mao}, \bibinfo{person}{Juanhui Li}, \bibinfo{person}{Harry Shomer}, \bibinfo{person}{Bingheng Li}, \bibinfo{person}{Wenqi Fan}, \bibinfo{person}{Yao Ma}, \bibinfo{person}{Tong Zhao}, \bibinfo{person}{Neil Shah}, {and} \bibinfo{person}{Jiliang Tang}.} \bibinfo{year}{2023}\natexlab{}.
\newblock \showarticletitle{Revisiting link prediction: A data perspective}.
\newblock \bibinfo{journal}{\emph{arXiv preprint arXiv:2310.00793}} (\bibinfo{year}{2023}).
\newblock


\bibitem[Martin et~al\mbox{.}(2016)]%
        {martin2016structural}
\bibfield{author}{\bibinfo{person}{Travis Martin}, \bibinfo{person}{Brian Ball}, {and} \bibinfo{person}{Mark~EJ Newman}.} \bibinfo{year}{2016}\natexlab{}.
\newblock \showarticletitle{Structural inference for uncertain networks}.
\newblock \bibinfo{journal}{\emph{Physical Review E}} \bibinfo{volume}{93}, \bibinfo{number}{1} (\bibinfo{year}{2016}), \bibinfo{pages}{012306}.
\newblock


\bibitem[Mart{\'\i}nez et~al\mbox{.}(2016)]%
        {martinez2016survey}
\bibfield{author}{\bibinfo{person}{V{\'\i}ctor Mart{\'\i}nez}, \bibinfo{person}{Fernando Berzal}, {and} \bibinfo{person}{Juan-Carlos Cubero}.} \bibinfo{year}{2016}\natexlab{}.
\newblock \showarticletitle{A survey of link prediction in complex networks}.
\newblock \bibinfo{journal}{\emph{ACM computing surveys (CSUR)}} \bibinfo{volume}{49}, \bibinfo{number}{4} (\bibinfo{year}{2016}), \bibinfo{pages}{1--33}.
\newblock


\bibitem[McFee and Lanckriet(2010)]%
        {mcfee2010metric}
\bibfield{author}{\bibinfo{person}{Brian McFee} {and} \bibinfo{person}{Gert Lanckriet}.} \bibinfo{year}{2010}\natexlab{}.
\newblock \showarticletitle{Metric learning to rank}. In \bibinfo{booktitle}{\emph{ICML}}.
\newblock


\bibitem[Monti et~al\mbox{.}(2017)]%
        {monti2017geometric}
\bibfield{author}{\bibinfo{person}{Federico Monti}, \bibinfo{person}{Michael Bronstein}, {and} \bibinfo{person}{Xavier Bresson}.} \bibinfo{year}{2017}\natexlab{}.
\newblock \showarticletitle{Geometric matrix completion with recurrent multi-graph neural networks}. In \bibinfo{booktitle}{\emph{NeurIPS}}.
\newblock


\bibitem[Morris et~al\mbox{.}(2019)]%
        {morris2019weisfeiler}
\bibfield{author}{\bibinfo{person}{Christopher Morris}, \bibinfo{person}{Martin Ritzert}, \bibinfo{person}{Matthias Fey}, \bibinfo{person}{William~L Hamilton}, \bibinfo{person}{Jan~Eric Lenssen}, \bibinfo{person}{Gaurav Rattan}, {and} \bibinfo{person}{Martin Grohe}.} \bibinfo{year}{2019}\natexlab{}.
\newblock \showarticletitle{Weisfeiler and leman go neural: Higher-order graph neural networks}. In \bibinfo{booktitle}{\emph{AAAI}}.
\newblock


\bibitem[Newman(2018)]%
        {newman2018networks}
\bibfield{author}{\bibinfo{person}{Mark Newman}.} \bibinfo{year}{2018}\natexlab{}.
\newblock \bibinfo{booktitle}{\emph{Networks}}.
\newblock \bibinfo{publisher}{Oxford university press}.
\newblock


\bibitem[Newman(2001)]%
        {newman2001clustering}
\bibfield{author}{\bibinfo{person}{Mark~EJ Newman}.} \bibinfo{year}{2001}\natexlab{}.
\newblock \showarticletitle{Clustering and preferential attachment in growing networks}.
\newblock \bibinfo{journal}{\emph{Physical review E}} \bibinfo{volume}{64}, \bibinfo{number}{2} (\bibinfo{year}{2001}), \bibinfo{pages}{025102}.
\newblock


\bibitem[Newman(2006)]%
        {newman2006modularity}
\bibfield{author}{\bibinfo{person}{Mark~EJ Newman}.} \bibinfo{year}{2006}\natexlab{}.
\newblock \showarticletitle{Modularity and community structure in networks}.
\newblock \bibinfo{journal}{\emph{PNAS}} \bibinfo{volume}{103}, \bibinfo{number}{23} (\bibinfo{year}{2006}), \bibinfo{pages}{8577--8582}.
\newblock


\bibitem[Ou et~al\mbox{.}(2016)]%
        {ou2016asymmetric}
\bibfield{author}{\bibinfo{person}{Mingdong Ou}, \bibinfo{person}{Peng Cui}, \bibinfo{person}{Jian Pei}, \bibinfo{person}{Ziwei Zhang}, {and} \bibinfo{person}{Wenwu Zhu}.} \bibinfo{year}{2016}\natexlab{}.
\newblock \showarticletitle{Asymmetric transitivity preserving graph embedding}. In \bibinfo{booktitle}{\emph{SIGKDD}}.
\newblock


\bibitem[Page et~al\mbox{.}(1999)]%
        {page1999pagerank}
\bibfield{author}{\bibinfo{person}{Lawrence Page}, \bibinfo{person}{Sergey Brin}, \bibinfo{person}{Rajeev Motwani}, {and} \bibinfo{person}{Terry Winograd}.} \bibinfo{year}{1999}\natexlab{}.
\newblock \bibinfo{booktitle}{\emph{The PageRank citation ranking: Bringing order to the web.}}
\newblock \bibinfo{type}{{T}echnical {R}eport}. \bibinfo{institution}{Stanford InfoLab}.
\newblock


\bibitem[Pan et~al\mbox{.}(2022)]%
        {pan2021neural}
\bibfield{author}{\bibinfo{person}{Liming Pan}, \bibinfo{person}{Cheng Shi}, {and} \bibinfo{person}{Ivan Dokmani{\'c}}.} \bibinfo{year}{2022}\natexlab{}.
\newblock \showarticletitle{Neural Link Prediction with Walk Pooling}. In \bibinfo{booktitle}{\emph{ICLR}}.
\newblock


\bibitem[Paszke et~al\mbox{.}(2019)]%
        {pytorch}
\bibfield{author}{\bibinfo{person}{Adam Paszke}, \bibinfo{person}{Sam Gross}, \bibinfo{person}{Francisco Massa}, \bibinfo{person}{Adam Lerer}, \bibinfo{person}{James Bradbury}, \bibinfo{person}{Gregory Chanan}, \bibinfo{person}{Trevor Killeen}, \bibinfo{person}{Zeming Lin}, \bibinfo{person}{Natalia Gimelshein}, \bibinfo{person}{Luca Antiga}, {et~al\mbox{.}}} \bibinfo{year}{2019}\natexlab{}.
\newblock \showarticletitle{Pytorch: An imperative style, high-performance deep learning library}. In \bibinfo{booktitle}{\emph{NeurIPS}}.
\newblock


\bibitem[Perozzi et~al\mbox{.}(2014)]%
        {perozzi2014deepwalk}
\bibfield{author}{\bibinfo{person}{Bryan Perozzi}, \bibinfo{person}{Rami Al-Rfou}, {and} \bibinfo{person}{Steven Skiena}.} \bibinfo{year}{2014}\natexlab{}.
\newblock \showarticletitle{Deepwalk: Online learning of social representations}. In \bibinfo{booktitle}{\emph{SIGKDD}}.
\newblock


\bibitem[Qi et~al\mbox{.}(2006)]%
        {qi2006evaluation}
\bibfield{author}{\bibinfo{person}{Yanjun Qi}, \bibinfo{person}{Ziv Bar-Joseph}, {and} \bibinfo{person}{Judith Klein-Seetharaman}.} \bibinfo{year}{2006}\natexlab{}.
\newblock \showarticletitle{Evaluation of different biological data and computational classification methods for use in protein interaction prediction}.
\newblock \bibinfo{journal}{\emph{Proteins: Structure, Function, and Bioinformatics}} \bibinfo{volume}{63}, \bibinfo{number}{3} (\bibinfo{year}{2006}), \bibinfo{pages}{490--500}.
\newblock


\bibitem[Qiu et~al\mbox{.}(2018a)]%
        {qiu2018network}
\bibfield{author}{\bibinfo{person}{Jiezhong Qiu}, \bibinfo{person}{Yuxiao Dong}, \bibinfo{person}{Hao Ma}, \bibinfo{person}{Jian Li}, \bibinfo{person}{Kuansan Wang}, {and} \bibinfo{person}{Jie Tang}.} \bibinfo{year}{2018}\natexlab{a}.
\newblock \showarticletitle{Network embedding as matrix factorization: Unifying deepwalk, line, pte, and node2vec}. In \bibinfo{booktitle}{\emph{WSDM}}.
\newblock


\bibitem[Qiu et~al\mbox{.}(2018b)]%
        {qiu2018deepinf}
\bibfield{author}{\bibinfo{person}{Jiezhong Qiu}, \bibinfo{person}{Jian Tang}, \bibinfo{person}{Hao Ma}, \bibinfo{person}{Yuxiao Dong}, \bibinfo{person}{Kuansan Wang}, {and} \bibinfo{person}{Jie Tang}.} \bibinfo{year}{2018}\natexlab{b}.
\newblock \showarticletitle{Deepinf: Social influence prediction with deep learning}. In \bibinfo{booktitle}{\emph{SIGKDD}}.
\newblock


\bibitem[Revaud et~al\mbox{.}(2019)]%
        {revaud2019learning}
\bibfield{author}{\bibinfo{person}{Jerome Revaud}, \bibinfo{person}{Jon Almaz{\'a}n}, \bibinfo{person}{Rafael~S Rezende}, {and} \bibinfo{person}{Cesar Roberto~de Souza}.} \bibinfo{year}{2019}\natexlab{}.
\newblock \showarticletitle{Learning with average precision: Training image retrieval with a listwise loss}. In \bibinfo{booktitle}{\emph{ICCV}}.
\newblock


\bibitem[Sahu et~al\mbox{.}(2019)]%
        {sahu2019inter}
\bibfield{author}{\bibinfo{person}{Sunil~Kumar Sahu}, \bibinfo{person}{Fenia Christopoulou}, \bibinfo{person}{Makoto Miwa}, {and} \bibinfo{person}{Sophia Ananiadou}.} \bibinfo{year}{2019}\natexlab{}.
\newblock \showarticletitle{Inter-sentence Relation Extraction with Document-level Graph Convolutional Neural Network}. In \bibinfo{booktitle}{\emph{ACL}}.
\newblock


\bibitem[Saito and Rehmsmeier(2015)]%
        {saito2015precision}
\bibfield{author}{\bibinfo{person}{Takaya Saito} {and} \bibinfo{person}{Marc Rehmsmeier}.} \bibinfo{year}{2015}\natexlab{}.
\newblock \showarticletitle{The precision-recall plot is more informative than the ROC plot when evaluating binary classifiers on imbalanced datasets}.
\newblock \bibinfo{journal}{\emph{PloS one}} \bibinfo{volume}{10}, \bibinfo{number}{3} (\bibinfo{year}{2015}), \bibinfo{pages}{e0118432}.
\newblock


\bibitem[Sanchez-Gonzalez et~al\mbox{.}(2018)]%
        {sanchez2018graph}
\bibfield{author}{\bibinfo{person}{Alvaro Sanchez-Gonzalez}, \bibinfo{person}{Nicolas Heess}, \bibinfo{person}{Jost~Tobias Springenberg}, \bibinfo{person}{Josh Merel}, \bibinfo{person}{Martin Riedmiller}, \bibinfo{person}{Raia Hadsell}, {and} \bibinfo{person}{Peter Battaglia}.} \bibinfo{year}{2018}\natexlab{}.
\newblock \showarticletitle{Graph networks as learnable physics engines for inference and control}. In \bibinfo{booktitle}{\emph{ICML}}.
\newblock


\bibitem[Sch{\"u}tze et~al\mbox{.}(2008)]%
        {schutze2008introduction}
\bibfield{author}{\bibinfo{person}{Hinrich Sch{\"u}tze}, \bibinfo{person}{Christopher~D Manning}, {and} \bibinfo{person}{Prabhakar Raghavan}.} \bibinfo{year}{2008}\natexlab{}.
\newblock \bibinfo{booktitle}{\emph{Introduction to information retrieval}}. Vol.~\bibinfo{volume}{39}.
\newblock \bibinfo{publisher}{Cambridge University Press Cambridge}.
\newblock


\bibitem[Singh et~al\mbox{.}(2021)]%
        {singh2021edge}
\bibfield{author}{\bibinfo{person}{Abhay Singh}, \bibinfo{person}{Qian Huang}, \bibinfo{person}{Sijia~Linda Huang}, \bibinfo{person}{Omkar Bhalerao}, \bibinfo{person}{Horace He}, \bibinfo{person}{Ser-Nam Lim}, {and} \bibinfo{person}{Austin~R Benson}.} \bibinfo{year}{2021}\natexlab{}.
\newblock \showarticletitle{Edge proposal sets for link prediction}.
\newblock \bibinfo{journal}{\emph{arXiv preprint arXiv:2106.15810}} (\bibinfo{year}{2021}).
\newblock


\bibitem[Sohn(2016)]%
        {sohn2016improved}
\bibfield{author}{\bibinfo{person}{Kihyuk Sohn}.} \bibinfo{year}{2016}\natexlab{}.
\newblock \showarticletitle{Improved deep metric learning with multi-class n-pair loss objective}. In \bibinfo{booktitle}{\emph{NeurIPS}}.
\newblock


\bibitem[Srinivasan and Ribeiro(2020)]%
        {srinivasanequivalence}
\bibfield{author}{\bibinfo{person}{Balasubramaniam Srinivasan} {and} \bibinfo{person}{Bruno Ribeiro}.} \bibinfo{year}{2020}\natexlab{}.
\newblock \showarticletitle{On the Equivalence between Positional Node Embeddings and Structural Graph Representations}. In \bibinfo{booktitle}{\emph{International Conference on Learning Representations}}.
\newblock


\bibitem[Sun et~al\mbox{.}(2018b)]%
        {sun2018open}
\bibfield{author}{\bibinfo{person}{Haitian Sun}, \bibinfo{person}{Bhuwan Dhingra}, \bibinfo{person}{Manzil Zaheer}, \bibinfo{person}{Kathryn Mazaitis}, \bibinfo{person}{Ruslan Salakhutdinov}, {and} \bibinfo{person}{William Cohen}.} \bibinfo{year}{2018}\natexlab{b}.
\newblock \showarticletitle{Open Domain Question Answering Using Early Fusion of Knowledge Bases and Text}. In \bibinfo{booktitle}{\emph{EMNLP}}.
\newblock


\bibitem[Sun et~al\mbox{.}(2018a)]%
        {sun2018rotate}
\bibfield{author}{\bibinfo{person}{Zhiqing Sun}, \bibinfo{person}{Zhi-Hong Deng}, \bibinfo{person}{Jian-Yun Nie}, {and} \bibinfo{person}{Jian Tang}.} \bibinfo{year}{2018}\natexlab{a}.
\newblock \showarticletitle{RotatE: Knowledge Graph Embedding by Relational Rotation in Complex Space}. In \bibinfo{booktitle}{\emph{ICLR}}.
\newblock


\bibitem[Tang et~al\mbox{.}(2008b)]%
        {tang2008arnetminer}
\bibfield{author}{\bibinfo{person}{Jie Tang}, \bibinfo{person}{Jing Zhang}, \bibinfo{person}{Limin Yao}, \bibinfo{person}{Juanzi Li}, \bibinfo{person}{Li Zhang}, {and} \bibinfo{person}{Zhong Su}.} \bibinfo{year}{2008}\natexlab{b}.
\newblock \showarticletitle{Arnetminer: extraction and mining of academic social networks}. In \bibinfo{booktitle}{\emph{SIGKDD}}.
\newblock


\bibitem[Tang et~al\mbox{.}(2008a)]%
        {tang2008svms}
\bibfield{author}{\bibinfo{person}{Yuchun Tang}, \bibinfo{person}{Yan-Qing Zhang}, \bibinfo{person}{Nitesh~V Chawla}, {and} \bibinfo{person}{Sven Krasser}.} \bibinfo{year}{2008}\natexlab{a}.
\newblock \showarticletitle{SVMs modeling for highly imbalanced classification}.
\newblock \bibinfo{journal}{\emph{IEEE Transactions on Systems, Man, and Cybernetics, Part B (Cybernetics)}} \bibinfo{volume}{39}, \bibinfo{number}{1} (\bibinfo{year}{2008}), \bibinfo{pages}{281--288}.
\newblock


\bibitem[Veli{\v{c}}kovi{\'c} et~al\mbox{.}(2018)]%
        {velivckovic2018graph}
\bibfield{author}{\bibinfo{person}{Petar Veli{\v{c}}kovi{\'c}}, \bibinfo{person}{Guillem Cucurull}, \bibinfo{person}{Arantxa Casanova}, \bibinfo{person}{Adriana Romero}, \bibinfo{person}{Pietro Li{\`o}}, {and} \bibinfo{person}{Yoshua Bengio}.} \bibinfo{year}{2018}\natexlab{}.
\newblock \showarticletitle{Graph Attention Networks}. In \bibinfo{booktitle}{\emph{ICLR}}.
\newblock


\bibitem[Wang et~al\mbox{.}(2019a)]%
        {wang2019kgat}
\bibfield{author}{\bibinfo{person}{Xiang Wang}, \bibinfo{person}{Xiangnan He}, \bibinfo{person}{Yixin Cao}, \bibinfo{person}{Meng Liu}, {and} \bibinfo{person}{Tat-Seng Chua}.} \bibinfo{year}{2019}\natexlab{a}.
\newblock \showarticletitle{Kgat: Knowledge graph attention network for recommendation}. In \bibinfo{booktitle}{\emph{SIGKDD}}.
\newblock


\bibitem[Wang et~al\mbox{.}(2019b)]%
        {wang2019ranked}
\bibfield{author}{\bibinfo{person}{Xinshao Wang}, \bibinfo{person}{Yang Hua}, \bibinfo{person}{Elyor Kodirov}, \bibinfo{person}{Guosheng Hu}, \bibinfo{person}{Romain Garnier}, {and} \bibinfo{person}{Neil~M Robertson}.} \bibinfo{year}{2019}\natexlab{b}.
\newblock \showarticletitle{Ranked list loss for deep metric learning}. In \bibinfo{booktitle}{\emph{ICCV}}.
\newblock


\bibitem[Wang et~al\mbox{.}(2023)]%
        {wang2023neural}
\bibfield{author}{\bibinfo{person}{Xiyuan Wang}, \bibinfo{person}{Haotong Yang}, {and} \bibinfo{person}{Muhan Zhang}.} \bibinfo{year}{2023}\natexlab{}.
\newblock \showarticletitle{Neural Common Neighbor with Completion for Link Prediction}.
\newblock \bibinfo{journal}{\emph{arXiv preprint arXiv:2302.00890}} (\bibinfo{year}{2023}).
\newblock


\bibitem[Wilder et~al\mbox{.}(2019)]%
        {wilder2019end}
\bibfield{author}{\bibinfo{person}{Bryan Wilder}, \bibinfo{person}{Eric Ewing}, \bibinfo{person}{Bistra Dilkina}, {and} \bibinfo{person}{Milind Tambe}.} \bibinfo{year}{2019}\natexlab{}.
\newblock \showarticletitle{End to end learning and optimization on graphs}.
\newblock \bibinfo{journal}{\emph{NeurIPS}} (\bibinfo{year}{2019}).
\newblock


\bibitem[Wu et~al\mbox{.}(2019)]%
        {wu2019simplifying}
\bibfield{author}{\bibinfo{person}{Felix Wu}, \bibinfo{person}{Amauri Souza}, \bibinfo{person}{Tianyi Zhang}, \bibinfo{person}{Christopher Fifty}, \bibinfo{person}{Tao Yu}, {and} \bibinfo{person}{Kilian Weinberger}.} \bibinfo{year}{2019}\natexlab{}.
\newblock \showarticletitle{Simplifying graph convolutional networks}. In \bibinfo{booktitle}{\emph{ICML}}.
\newblock


\bibitem[Xia et~al\mbox{.}(2008)]%
        {xia2008listwise}
\bibfield{author}{\bibinfo{person}{Fen Xia}, \bibinfo{person}{Tie-Yan Liu}, \bibinfo{person}{Jue Wang}, \bibinfo{person}{Wensheng Zhang}, {and} \bibinfo{person}{Hang Li}.} \bibinfo{year}{2008}\natexlab{}.
\newblock \showarticletitle{Listwise approach to learning to rank: theory and algorithm}. In \bibinfo{booktitle}{\emph{ICML}}.
\newblock


\bibitem[Xu et~al\mbox{.}(2018)]%
        {xu2018powerful}
\bibfield{author}{\bibinfo{person}{Keyulu Xu}, \bibinfo{person}{Weihua Hu}, \bibinfo{person}{Jure Leskovec}, {and} \bibinfo{person}{Stefanie Jegelka}.} \bibinfo{year}{2018}\natexlab{}.
\newblock \showarticletitle{How Powerful are Graph Neural Networks?}. In \bibinfo{booktitle}{\emph{ICLR}}.
\newblock


\bibitem[Yan et~al\mbox{.}(2021)]%
        {yan2021link}
\bibfield{author}{\bibinfo{person}{Zuoyu Yan}, \bibinfo{person}{Tengfei Ma}, \bibinfo{person}{Liangcai Gao}, \bibinfo{person}{Zhi Tang}, {and} \bibinfo{person}{Chao Chen}.} \bibinfo{year}{2021}\natexlab{}.
\newblock \showarticletitle{Link prediction with persistent homology: An interactive view}. In \bibinfo{booktitle}{\emph{ICML}}.
\newblock


\bibitem[Yang et~al\mbox{.}(2015)]%
        {yang2015embedding}
\bibfield{author}{\bibinfo{person}{Bishan Yang}, \bibinfo{person}{Scott Wen-tau Yih}, \bibinfo{person}{Xiaodong He}, \bibinfo{person}{Jianfeng Gao}, {and} \bibinfo{person}{Li Deng}.} \bibinfo{year}{2015}\natexlab{}.
\newblock \showarticletitle{Embedding Entities and Relations for Learning and Inference in Knowledge Bases}. In \bibinfo{booktitle}{\emph{ICLR}}.
\newblock


\bibitem[Yang et~al\mbox{.}(2021)]%
        {yang2021enhanced}
\bibfield{author}{\bibinfo{person}{Yonghui Yang}, \bibinfo{person}{Le Wu}, \bibinfo{person}{Richang Hong}, \bibinfo{person}{Kun Zhang}, {and} \bibinfo{person}{Meng Wang}.} \bibinfo{year}{2021}\natexlab{}.
\newblock \showarticletitle{Enhanced graph learning for collaborative filtering via mutual information maximization}. In \bibinfo{booktitle}{\emph{SIGIR}}.
\newblock


\bibitem[Yao et~al\mbox{.}(2019)]%
        {yao2019graph}
\bibfield{author}{\bibinfo{person}{Liang Yao}, \bibinfo{person}{Chengsheng Mao}, {and} \bibinfo{person}{Yuan Luo}.} \bibinfo{year}{2019}\natexlab{}.
\newblock \showarticletitle{Graph convolutional networks for text classification}. In \bibinfo{booktitle}{\emph{AAAI}}.
\newblock


\bibitem[Ying et~al\mbox{.}(2018)]%
        {ying2018hierarchical}
\bibfield{author}{\bibinfo{person}{Zhitao Ying}, \bibinfo{person}{Jiaxuan You}, \bibinfo{person}{Christopher Morris}, \bibinfo{person}{Xiang Ren}, \bibinfo{person}{Will Hamilton}, {and} \bibinfo{person}{Jure Leskovec}.} \bibinfo{year}{2018}\natexlab{}.
\newblock \showarticletitle{Hierarchical graph representation learning with differentiable pooling}. In \bibinfo{booktitle}{\emph{NeurIPS}}.
\newblock


\bibitem[You et~al\mbox{.}(2021)]%
        {you2021identity}
\bibfield{author}{\bibinfo{person}{Jiaxuan You}, \bibinfo{person}{Jonathan~M Gomes-Selman}, \bibinfo{person}{Rex Ying}, {and} \bibinfo{person}{Jure Leskovec}.} \bibinfo{year}{2021}\natexlab{}.
\newblock \showarticletitle{Identity-aware graph neural networks}. In \bibinfo{booktitle}{\emph{Proceedings of the AAAI conference on artificial intelligence}}. \bibinfo{pages}{10737--10745}.
\newblock


\bibitem[You et~al\mbox{.}(2018)]%
        {you2018graphrnn}
\bibfield{author}{\bibinfo{person}{Jiaxuan You}, \bibinfo{person}{Rex Ying}, \bibinfo{person}{Xiang Ren}, \bibinfo{person}{William Hamilton}, {and} \bibinfo{person}{Jure Leskovec}.} \bibinfo{year}{2018}\natexlab{}.
\newblock \showarticletitle{Graphrnn: Generating realistic graphs with deep auto-regressive models}. In \bibinfo{booktitle}{\emph{ICML}}.
\newblock


\bibitem[Yun et~al\mbox{.}(2021)]%
        {yun2021neo}
\bibfield{author}{\bibinfo{person}{Seongjun Yun}, \bibinfo{person}{Seoyoon Kim}, \bibinfo{person}{Junhyun Lee}, \bibinfo{person}{Jaewoo Kang}, {and} \bibinfo{person}{Hyunwoo~J Kim}.} \bibinfo{year}{2021}\natexlab{}.
\newblock \showarticletitle{Neo-GNNs: Neighborhood Overlap-aware Graph Neural Networks for Link Prediction}. In \bibinfo{booktitle}{\emph{NeurIPS}}.
\newblock


\bibitem[Zhang and Chen(2017)]%
        {zhang2017weisfeiler}
\bibfield{author}{\bibinfo{person}{Muhan Zhang} {and} \bibinfo{person}{Yixin Chen}.} \bibinfo{year}{2017}\natexlab{}.
\newblock \showarticletitle{Weisfeiler-lehman neural machine for link prediction}. In \bibinfo{booktitle}{\emph{Proceedings of the 23rd ACM SIGKDD international conference on knowledge discovery and data mining}}. \bibinfo{pages}{575--583}.
\newblock


\bibitem[Zhang and Chen(2018)]%
        {zhang2018link}
\bibfield{author}{\bibinfo{person}{Muhan Zhang} {and} \bibinfo{person}{Yixin Chen}.} \bibinfo{year}{2018}\natexlab{}.
\newblock \showarticletitle{Link prediction based on graph neural networks}. In \bibinfo{booktitle}{\emph{NeurIPS}}.
\newblock


\bibitem[Zhang et~al\mbox{.}(2018a)]%
        {zhang2018end}
\bibfield{author}{\bibinfo{person}{Muhan Zhang}, \bibinfo{person}{Zhicheng Cui}, \bibinfo{person}{Marion Neumann}, {and} \bibinfo{person}{Yixin Chen}.} \bibinfo{year}{2018}\natexlab{a}.
\newblock \showarticletitle{An end-to-end deep learning architecture for graph classification}. In \bibinfo{booktitle}{\emph{AAAI}}.
\newblock


\bibitem[Zhang et~al\mbox{.}(2021a)]%
        {zhang2021labeling}
\bibfield{author}{\bibinfo{person}{Muhan Zhang}, \bibinfo{person}{Pan Li}, \bibinfo{person}{Yinglong Xia}, \bibinfo{person}{Kai Wang}, {and} \bibinfo{person}{Long Jin}.} \bibinfo{year}{2021}\natexlab{a}.
\newblock \showarticletitle{Labeling Trick: A Theory of Using Graph Neural Networks for Multi-Node Representation Learning}. In \bibinfo{booktitle}{\emph{NeurIPS}}.
\newblock


\bibitem[Zhang et~al\mbox{.}(2021b)]%
        {zhang2021lorentzian}
\bibfield{author}{\bibinfo{person}{Yiding Zhang}, \bibinfo{person}{Xiao Wang}, \bibinfo{person}{Chuan Shi}, \bibinfo{person}{Nian Liu}, {and} \bibinfo{person}{Guojie Song}.} \bibinfo{year}{2021}\natexlab{b}.
\newblock \showarticletitle{Lorentzian graph convolutional networks}. In \bibinfo{booktitle}{\emph{WebConf}}.
\newblock


\bibitem[Zhang et~al\mbox{.}(2018b)]%
        {zhang2018arbitrary}
\bibfield{author}{\bibinfo{person}{Ziwei Zhang}, \bibinfo{person}{Peng Cui}, \bibinfo{person}{Xiao Wang}, \bibinfo{person}{Jian Pei}, \bibinfo{person}{Xuanrong Yao}, {and} \bibinfo{person}{Wenwu Zhu}.} \bibinfo{year}{2018}\natexlab{b}.
\newblock \showarticletitle{Arbitrary-order proximity preserved network embedding}. In \bibinfo{booktitle}{\emph{SIGKDD}}.
\newblock


\bibitem[Zhao et~al\mbox{.}(2022)]%
        {zhao2022graph}
\bibfield{author}{\bibinfo{person}{Tong Zhao}, \bibinfo{person}{Gang Liu}, \bibinfo{person}{Stephan G{\"u}nnemann}, {and} \bibinfo{person}{Meng Jiang}.} \bibinfo{year}{2022}\natexlab{}.
\newblock \showarticletitle{Graph data augmentation for graph machine learning: A survey}.
\newblock \bibinfo{journal}{\emph{arXiv preprint arXiv:2202.08871}} (\bibinfo{year}{2022}).
\newblock


\bibitem[Zhao et~al\mbox{.}(2021)]%
        {zhao2021data}
\bibfield{author}{\bibinfo{person}{Tong Zhao}, \bibinfo{person}{Yozen Liu}, \bibinfo{person}{Leonardo Neves}, \bibinfo{person}{Oliver Woodford}, \bibinfo{person}{Meng Jiang}, {and} \bibinfo{person}{Neil Shah}.} \bibinfo{year}{2021}\natexlab{}.
\newblock \showarticletitle{Data augmentation for graph neural networks}. In \bibinfo{booktitle}{\emph{AAAI}}.
\newblock


\bibitem[Zheng et~al\mbox{.}(2020)]%
        {zheng2020robust}
\bibfield{author}{\bibinfo{person}{Cheng Zheng}, \bibinfo{person}{Bo Zong}, \bibinfo{person}{Wei Cheng}, \bibinfo{person}{Dongjin Song}, \bibinfo{person}{Jingchao Ni}, \bibinfo{person}{Wenchao Yu}, \bibinfo{person}{Haifeng Chen}, {and} \bibinfo{person}{Wei Wang}.} \bibinfo{year}{2020}\natexlab{}.
\newblock \showarticletitle{Robust graph representation learning via neural sparsification}. In \bibinfo{booktitle}{\emph{ICML}}.
\newblock


\bibitem[Zhou et~al\mbox{.}(2022)]%
        {zhou2022link}
\bibfield{author}{\bibinfo{person}{Shijie Zhou}, \bibinfo{person}{Zhimeng Guo}, \bibinfo{person}{Charu Aggarwal}, \bibinfo{person}{Xiang Zhang}, {and} \bibinfo{person}{Suhang Wang}.} \bibinfo{year}{2022}\natexlab{}.
\newblock \showarticletitle{Link prediction on heterophilic graphs via disentangled representation learning}.
\newblock \bibinfo{journal}{\emph{arXiv preprint arXiv:2208.01820}} (\bibinfo{year}{2022}).
\newblock


\bibitem[Zhou et~al\mbox{.}(2009)]%
        {zhou2009predicting}
\bibfield{author}{\bibinfo{person}{Tao Zhou}, \bibinfo{person}{Linyuan L{\"u}}, {and} \bibinfo{person}{Yi-Cheng Zhang}.} \bibinfo{year}{2009}\natexlab{}.
\newblock \showarticletitle{Predicting missing links via local information}.
\newblock \bibinfo{journal}{\emph{The European Physical Journal B}} \bibinfo{volume}{71}, \bibinfo{number}{4} (\bibinfo{year}{2009}), \bibinfo{pages}{623--630}.
\newblock


\bibitem[Zhu et~al\mbox{.}(2021)]%
        {zhu2021neural}
\bibfield{author}{\bibinfo{person}{Zhaocheng Zhu}, \bibinfo{person}{Zuobai Zhang}, \bibinfo{person}{Louis-Pascal Xhonneux}, {and} \bibinfo{person}{Jian Tang}.} \bibinfo{year}{2021}\natexlab{}.
\newblock \showarticletitle{Neural bellman-ford networks: A general graph neural network framework for link prediction}. In \bibinfo{booktitle}{\emph{NeurIPS}}.
\newblock


\end{thebibliography}

\clearpage
\appendix
\appendix



\section{Analysis of link prediction evaluation metrics with different test settings}
\label{ap::example}

\emph{Example}: Consider a graph with $10K$ nodes, $100K$ edges, and $99.9M$ disconnected (or negative) pairs. A (bad) model that ranks 1M false positives higher than the true edges achieves $0.99$ AUC and $0.95$ in AP under \emph{biased testing} with equal negative samples.

\begin{figure*}[htbp]
  \centering
  \subfloat[ROC]{\label{subfig::roc}\includegraphics[width=0.333\textwidth]{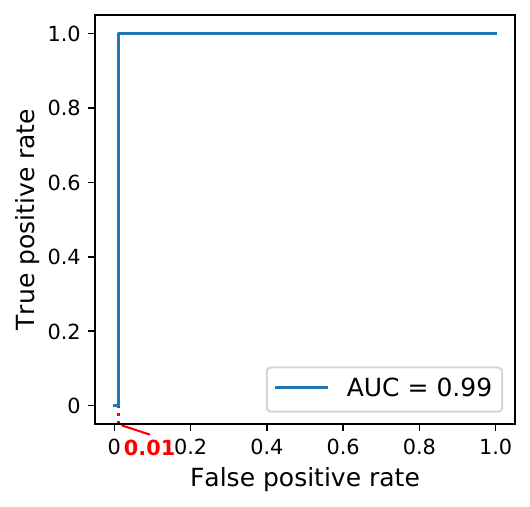}}
  \subfloat[PR under \textit{biased testing}]{\label{subfig::pr}\includegraphics[width=0.333\textwidth]{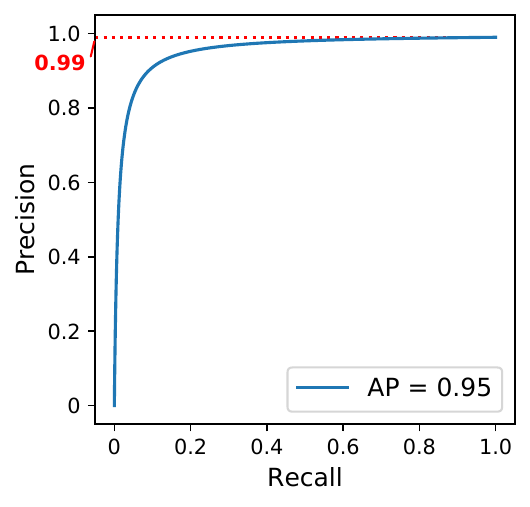}}
  \subfloat[PR under \textit{unbiased testing}]{\label{subfig::pr_unbiased}\includegraphics[width=0.333\textwidth]{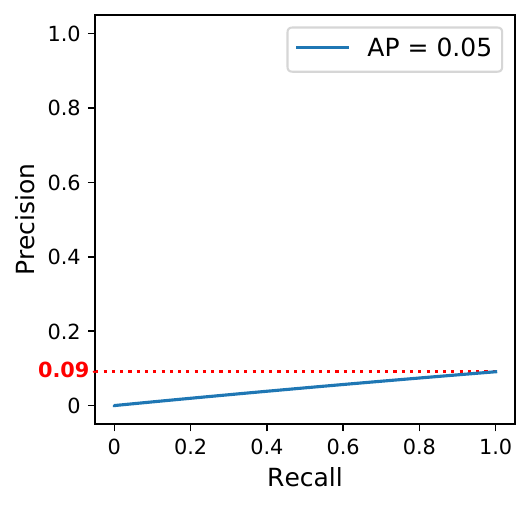}}
  \caption{Receiver operating characteristic and precision-recall curves for the bad link prediction model that ranks 1M false positives higher than the 100k true edges. 
  The model achieves 0.99 in AUC and 0.95 AP under \emph{biased testing}, while the more informative performance evaluation metric, Average Precision (AP) under \emph{unbiased testing}, is only 0.05. }
  \label{fig::roc_pr}
\end{figure*}

Figures \ref{subfig::roc} and \ref{subfig::pr} show the receiver operating characteristic (ROC) and precision-recall (PR) curves for the model under \emph{biased testing} with equal number of negative samples. Due to the downsampling, only 100k (out of 99.9M) negative pairs are included in the test set, among which only 
${100\text{k}}/{99.9\text{M}} \times 1\text{M} \approx 1\text{k}$ pairs are ranked higher than the positive edges. In the ROC curve, this means that once the false positive rate reaches ${1\text{k}}/{100\text{k}}=0.01$, the true positive rate would reach 1.0, leading to an AUC score of 0.99. Similarly, in the PR curve, when the recall reaches 1.0, the precision is ${100\text{k}}/({1\text{k}+100\text{k}})\approx 0.99$, leading to an overall AP score of \raise.17ex\hbox{$\scriptstyle\sim$}0.95. 

By comparison, as shown in \autoref{subfig::pr_unbiased}, when the recall reaches 1.0, the precision under \emph{unbiased testing} is only
${100\text{k}}/({1\text{M}+100\text{k}})\approx 0.09$, leading to an AP score of \raise.17ex\hbox{$\scriptstyle\sim$}0.05. 
This demonstrates that evaluation metrics based on \emph{biased testing} provide an overly optimistic measurement of link prediction model performance compared to the more realistic \emph{unbiased testing} setting.

\section{Proof of Theorem \ref{thm::unbiased}}
\label{ap:proof_unbiased}

There are only three classifiers that we need to consider in this setting, assuming that the classifier can recover the block structure:
\begin{enumerate}
\item It predicts every disconnected pair as a link;
\item It predicts every disconnected pair as a non-link;
\item It predicts within-block pairs as links and across-block pairs as non-links.
\end{enumerate}

The classifier 1 cannot be optimal for sparse graphs---i.e., density lower than $.5$---and thus we will focus on classifiers 2 and 3. We will compute the expected number of True Positives (TP), False Positives (FP), False Negatives (FN), and True Negatives (TN) per node for each of them:

\paragraph{Classifier 2:} 

\begin{align*}
    TP &= 0\\
    FN &= 0\\
    FP &= (n-1)p+(nk-n)q\\
    TN &= (n-1)(1-p)+(nk-n)(1-q)
\end{align*}

\paragraph{Classifier 3:} 

\begin{align*}
    TP &= (n-1)p\\
    FN &= (nk-n)q\\
    FP &= (n-1)p\\
    TN &= (nk-n)(1-q)
\end{align*}

The accuracy of the classifiers is computed as $(TP+TN)/(TP+TN+FP+FN)$. It follows that the difference between accuracy of the classifier 2 and 3 is as follows:
\begin{equation*}
    \frac{(n-1)(1-p)+(nk-n)(1-q)}{nk-1}-\frac{(n-1)p+(nk-n)(1-q)}{nk-1}
\end{equation*}

And thus, classifier 2 outperforms classifier 3 for $p<0.5$.

\section{Proof of Lemma \ref{lemma::biased}}
\label{lemma::biased_proof}

We will consider the same classifiers 2 and 3 from the proof of Theorem \ref{thm::unbiased}. Moreover, we will assume that the number of sampled negative pairs is the same as the number of positive pairs (i.e., balanced sampling). 

By definition, the accuracy of classifier 2 is 0.5, as all predictions for negative pairs will be correct and all those for positive pairs will be incorrect. Thus, we only have to show that there exists an SBM instance for which classifier 3 achieves better accuracy than 2.

The accuracy of classifier 3 is computed as $a_1+a_2/2$, where:
\begin{align*}
    a_1 &= \frac{(n-1)p}{(n-1)p+(nk-n)q}\\
    a_2 &= \frac{(nk-n)(1-q)}{(nk-n)(1-q)+(n-1)(1-p)}
\end{align*}

It follows that, as $q \to 0$, classifier 3 can achieve an accuracy higher than $0.5$. 

\section{Proof of Lemma \ref{lemma:autocov}}
\label{ap::lemma}

    
Let us initially consider Autocovariance with $t=1$ computed in the Stochastic Block Model described in Lemma \ref{lemma:autocov}. We will adopt the entry-wise notation of the original Autocovariance definition presented in Section \ref{sec:topological-heuristic}, using lower-case letters to represent individual entries in matrices and vectors, and for the sake of consistency with the Modularity definition, we adopt $\text{vol}(G)=2m$. We first obtain the shortened form of Autocovariance for $t=1$:

\begin{align}
R_{ij}
&=\cfrac{1}{2m}(a_{ij} - \cfrac{d_id_j}{2m}).
\end{align}

We can obtain the expected expression value for the case where $(i, j)$ is an intra-cluster pair ($\mathbb{E}[R_{intra}]$):

\begin{align}
    \mathbb{E}[R_{intra}]&=\cfrac{1}{2m}((1-\cfrac{d_id_j}{2m})p + (0-\cfrac{d_id_j}{2m})(1-p))
    \\
    &=\cfrac{1}{2m}(p-\cfrac{d_id_j}{2m}).
\end{align}

Likewise, we follow the same procedure for the case where $(i, j)$ is an inter-cluster pair ($\mathbb{E}[R_{inter}]$): 

\begin{align}
    \mathbb{E}[R_{inter}]&=\cfrac{1}{2m}((1-\cfrac{d_id_j}{2m})(1-p) + (0-\cfrac{d_id_j}{2m})p)
    \\
    &=\cfrac{1}{2m}(1-p-\cfrac{d_id_j}{2m})
    \\
    &=\cfrac{1}{2m}(q-\cfrac{d_id_j}{2m}).
\end{align}

Due to the reversible property of Markov chains, this holds for larger values of $t$.

Since $p > q \implies \mathbb{E}[R_{intra}] > \mathbb{E}[R_{inter}]$.


\section{Proof of Lemma \ref{lemma:increasing_k}}
\label{ap::lemma_increasing_k}

From Appendix \ref{ap::lemma}, we have $\mathbb{E}[R_{intra}]=\cfrac{1}{2m}(p-\cfrac{d_id_j}{2m})$ is solely dependent on the value of $p$, since all the other terms are constants. We will denominate $V_{ik}$ and $E_{ik}^+$ the number of nodes and positive pairs in the $i$-th partition of our graph partitioned in $k$ partitions. 

Considering the estimate $p=\nicefrac{|E_{ik}^+|}{|V_{ik}|^2}$, for simplicity, the number of positive pairs we can lose by increasing $k$ to $k+1$ is \textit{at most} $|E_{ik+1}^+| \geq |E_{ik}^+| - (|V_{ik}|^2 - |V_{ik+1}|^2)$, if we consider the extreme scenario in which every pair lost was positive. With this estimate, we can compare with the actual $p$ estimate:

\begin{align}
    \cfrac{|E_{ik+1}^+|}{|V_{ik+1}|^2}&\geq\cfrac{|E_{ik}^+| - (|V_{ik}|^2 - |V_{ik+1}|^2)}{|V_{ik+1}|^2}
    \\
    |V_{ik}|^2 - |V_{ik+1}|^2 &\geq |E_{ik}^+| - |E_{ik+1}^+|
\end{align}

It follows that, since the number of pairs drops faster than the number of positive edges for a given partition, $\mathbb{E}[R_{intra}]$ increases when $k$ increases.

\section{Can GNNs learn autocovariance?}
\label{ap::learn_autocov}

\begin{figure}[htbp]
\centering
\includegraphics[height=.4\linewidth]{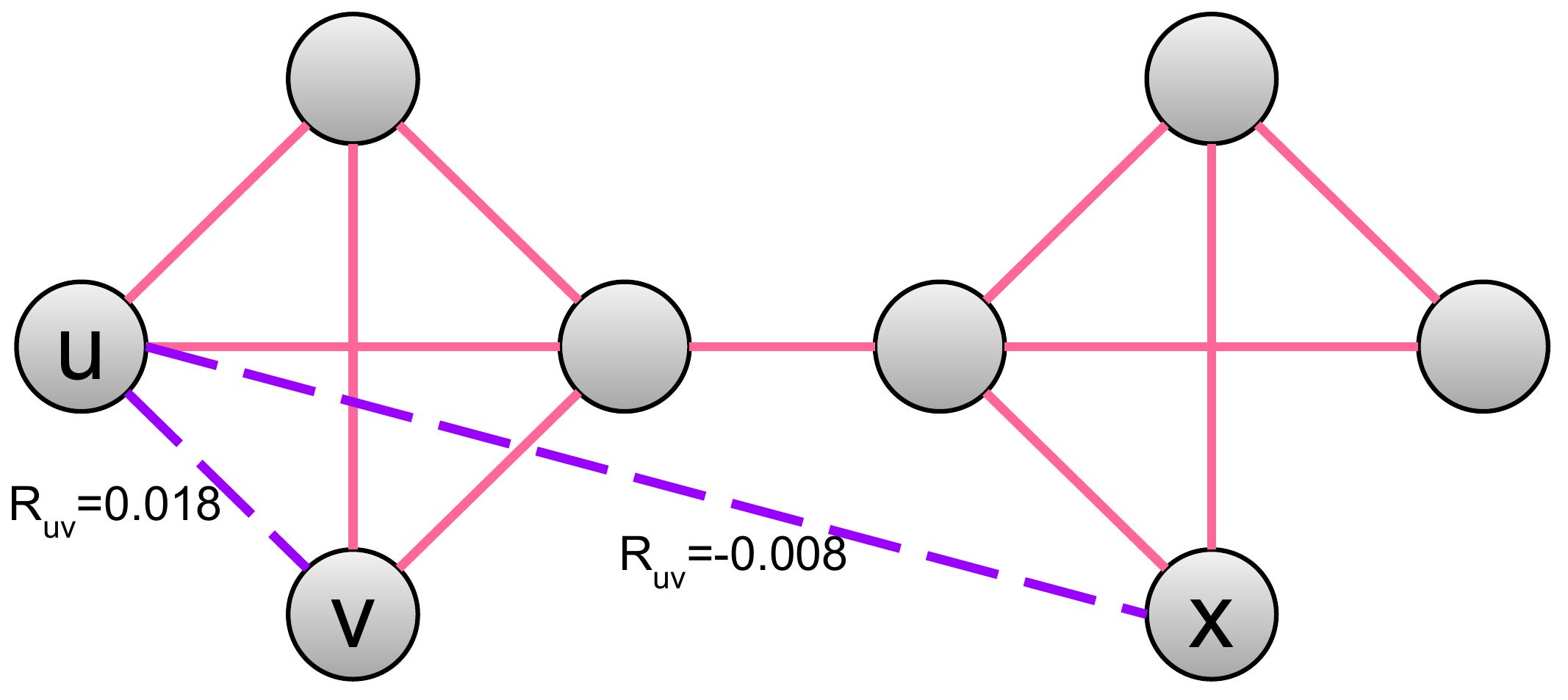}
\caption{MPNNs for link prediction cannot distinguish pairs $(u,v)$ and $(u,x)$ but they have different Autocovariance values, $R_{uv}=0.018$ and $R_{ux}=0.008$, for $t=2$. \label{fig::mpnn_autocov}}
\end{figure}

\textbf{Message-passing Neural Networks:} Classical message-passing neural networks (MPNNs) are known to be as powerful as the 1-WL isomorphism test. Recent papers have shown how this limitation affects link prediction performance \citep{chamberlain2022graph,you2021identity,zhang2017weisfeiler,srinivasanequivalence}. Node pairs $(u,v)$ and $(u,x)$ are indistinguishable by MPNNs if $v$ and $x$ have the same receptive field (or k-hop neighborhood). Figure \ref{fig::mpnn_autocov} shows that MPNNs are also unable distinguish node pairs with different values of Autocovariance within a graph $G$. Recently, more powerful GNNs for link prediction have also been proposed \citep{hu2022two}. These GNNs are as powerful as the 2-WL and 2-FWL isomorphism tests, which are two versions of the 2-dimensional WL test and are more discriminative than 1-WL for link prediction. While 2-WL powerful GNNs are still not able to distinguish pairs $(u,v)$ and $(u,x)$ in Figure \ref{fig::mpnn_autocov}, 2-FWL powerful GNNs can. This is due to the ability of 2-FWL powerful GNNs to count open and closed triads involving pairs of nodes---$(u,v)$ is part of two triangles while $(u,x)$ is part of none. However, we notice that counting triads is not sufficient to compute probabilities of paths longer than 2 hops connecting a pair of nodes. Moreover, training a GNN based on a 2-dimensional WL test takes $O(n^3)$ time, which prevents their application to large graphs.

\textbf{Subgraph Neural Networks:} Subgraph neural networks (SGNNs) differ from MPNNs as they learn representations based on node enclosing subgraphs \citep{li2020distance,zhang2017weisfeiler,zhang2021labeling,hu2022two, chen2020can}. These subgraphs are augmented with structural features that have been proven to increase their expressive power. However, SGNNs are also known to be computationally intractable \citep{chamberlain2022graph}. Previous work has shown that SGNNs can count the number of paths of fixed length between pairs of nodes when the aggregation operator is SUM \cite{you2021identity}. Autocovariance is a function of path counts, node degrees, and the graph volume (constant). Therefore, it is straightforward to design a SGNN that can predict Autocovariance. However, we note that our empirical results show that SEAL and BUDDY are often outperformed by Gelato. This can be explained by the specific design of these GNNs (e.g. aggregation operator) and the sampling complexity of accurately learning Autocovariance directly from data.

\section{Estimated Stochastic Block Model Parameters}
\label{ap::estimated_sbm}

We estimate the intra-block ($p$) and inter-block ($q$) parameters of each dataset considered in our experiments using either the node labels as ground-truth partitions (for Cora, CiteSeer, and PubMed) or METIS partitions (for OGBL-DDI and OGBL-Collab) following the values exposed in Table \ref{tab::ap::clustering_times}. The intra-block parameter is obtained through the ratio between the number of edges of the biggest partition and all the edges in the graph. We argue that Autocovariance-based design leverages the topology of datasets heavily organized as communities (Cora, CiteSeer, and PubMed) or even highly sparse (OGBL-Collab) to obtain state-of-the-art performance. We notice, however, that these benefits diminish in extremely dense networks, such as OGBL-DDI, a challenging scenario for all methods.

\begin{table}
\begin{tabular}{llllll}
\hline
 & Cora   & CiteSeer & PubMed & OGBL-DDI & OGBL-Collab \\ \hline
$p$ & 0.0217 & 0.0070   & 0.0006 & 0.2937   & 0.0005    \\
$q$ & 0.0004 & 0.0005   & 0.00007 & 0.1161   & 0.00004      \\ \hline
\end{tabular}
\caption{Estimated Stochastic Block Model parameters for each dataset considered in our work:  intra-block density parameter ($p$) and inter-block density parameter ($q$). The Autocovariance mechanism enables leveraging the community organization and/or extreme sparsity to achieve state-of-the-art link prediction results. We note that, unlike the SBM model, real graphs have blocks of different sizes. \label{table::sbm}}
\end{table}

\section{Detailed experiment settings}
\label{ap::setting}


\noindent\textbf{Positive masking.} For \emph{unbiased training}, a trick similar to \emph{negative injection} \citep{zhang2018link} in \emph{biased training} is needed to guarantee model generalizability. Specifically, we divide the training positive edges into batches and during the training with each batch $E_b$, we feed in only the residual edges $E-E_b$ as the structural information to the model. This setting simulates the testing phase, where the model is expected to predict edges without using their own connectivity information. We term this trick \textit{positive masking}.


\noindent\textbf{Other implementation details.} We add self-loops to the enhanced adjacency matrix to ensure that each node has a valid transition probability distribution that is used in computing Autocovariance. The self-loops are added to all isolated nodes in the training graph for all datasets. Following the postprocessing of the Autocovariance matrix for embedding in \cite{random-walk-embedding}, we standardize Gelato similarity scores before computing the loss. 
We optimize our model with gradient descent via \texttt{autograd} in \texttt{pytorch} \citep{pytorch}. We find that the gradients are sometimes invalid when training our model (especially with the cross-entropy loss), and we address this by skipping the parameter updates for batches leading to invalid gradients. Finally, we use $prec@100\%$ on the (unbiased) validation set as the criteria for selecting the best model from all training epochs. The maximum number of epochs for \textsc{Cora}/\textsc{CiteSeer} and \textsc{OGBL-DDI}/\textsc{OGBL-Collab} is set to be 100 and 250, respectively. For \emph{partitioned testing}, we apply METIS \citep{karypis1998fast} as our graph partitioning algorithm, due to its scalability and a balanced number of nodes per partition.

\noindent\textbf{Experiment environment.} We run our experiments in an \emph{a2-highgpu-1g} node of the Google Cloud Compute Engine. It has one NVIDIA A100 GPU with 40GB HBM2 GPU memory and 12 Intel Xeon Scalable Processor (Cascade Lake) 2nd Generation vCPUs with 85GB memory. 

\noindent\textbf{Reference of baselines.} We list link prediction baselines and their reference repositories we use in our experiments in \autoref{tab::baseline}. Note that we had to implement the batched training and testing for several baselines as their original implementations do not scale to \emph{unbiased training} and \emph{unbiased testing} without downsampling. 


\begin{table}[htbp]
  \small
  \centering
  \setlength{\tabcolsep}{3pt}
  \caption{Reference of baseline code repositories. }
    \begin{tabular}{cc}
    \toprule
     Baseline  & Repository \\
    \midrule
    SEAL \citep{zhang2018link}  & \url{https://github.com/facebookresearch/SEAL_OGB} 
    \\
    Neo-GNN  \citep{yun2021neo}  & \url{https://github.com/seongjunyun/Neo-GNNs} 
    \\

    BUDDY  \citep{chamberlain2022graph}  & \url{https://github.com/melifluos/subgraph-sketching}  

    \\
    NCN / NCNC  \citep{wang2023neural}  & \url{https://github.com/zexihuang/random-walk-embedding}  

    \\
    \bottomrule
    \end{tabular}%
  \label{tab::baseline}%
\end{table}


\section{Gelato - Unbiased vs Partitioned}
\label{sec::similarities_non_normalized}

\begin{figure}

    \includegraphics[width=0.4\textwidth]{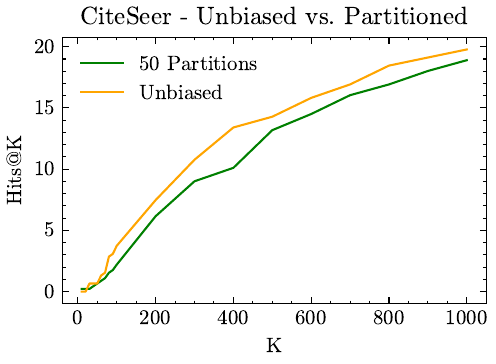}
    \caption{Comparison between Gelato trained using \emph{unbiased} sampling against \emph{partitioned} sampling on CiteSeer for different values of $K$. We verify that even in extreme partitioning scenarios ($k=50$, $\approx 66$ nodes per partition), there is only a small performance gap between both models, but the \textit{partitioned} sampling approach trains almost 6x times faster than the \textit{unbiased} sampling approach. The speedup increases with the number of partitions.}
    \label{fig::ap::partitioned_vs_unbiased}
\end{figure}

Figure \ref{fig::ap::partitioned_vs_unbiased} demonstrates we obtain splits that are both realistic and scalable using \textit{partitioned sampling} through varying values of K in hits@K metric evaluated on CiteSeer. It is possible to verify that there is almost no performance gap between \textit{partitioned} and \textit{unbiased} training, even in a very extreme partitioning scenario. Unbiased training takes $O(V^2 - E)$ for sparse graphs due to a large number of negative samples, while proposed negative sampling significantly reduces the training time to $O(\sum_i^p |V_i|^2 - |E_i|)$, where $(V_i,E_i)$ are the sets of nodes and edges within partition $i$. We experimented with different values of $p$ and obtained negligible impact on performance, demonstrating that the choice of the parameter $p$ is not critical to the success of the approach.

\section{Additional Link Prediction Model Comparison}
\label{ap::prec@k}

\begin{figure*}

    \centering
    \includegraphics[width=1\textwidth]{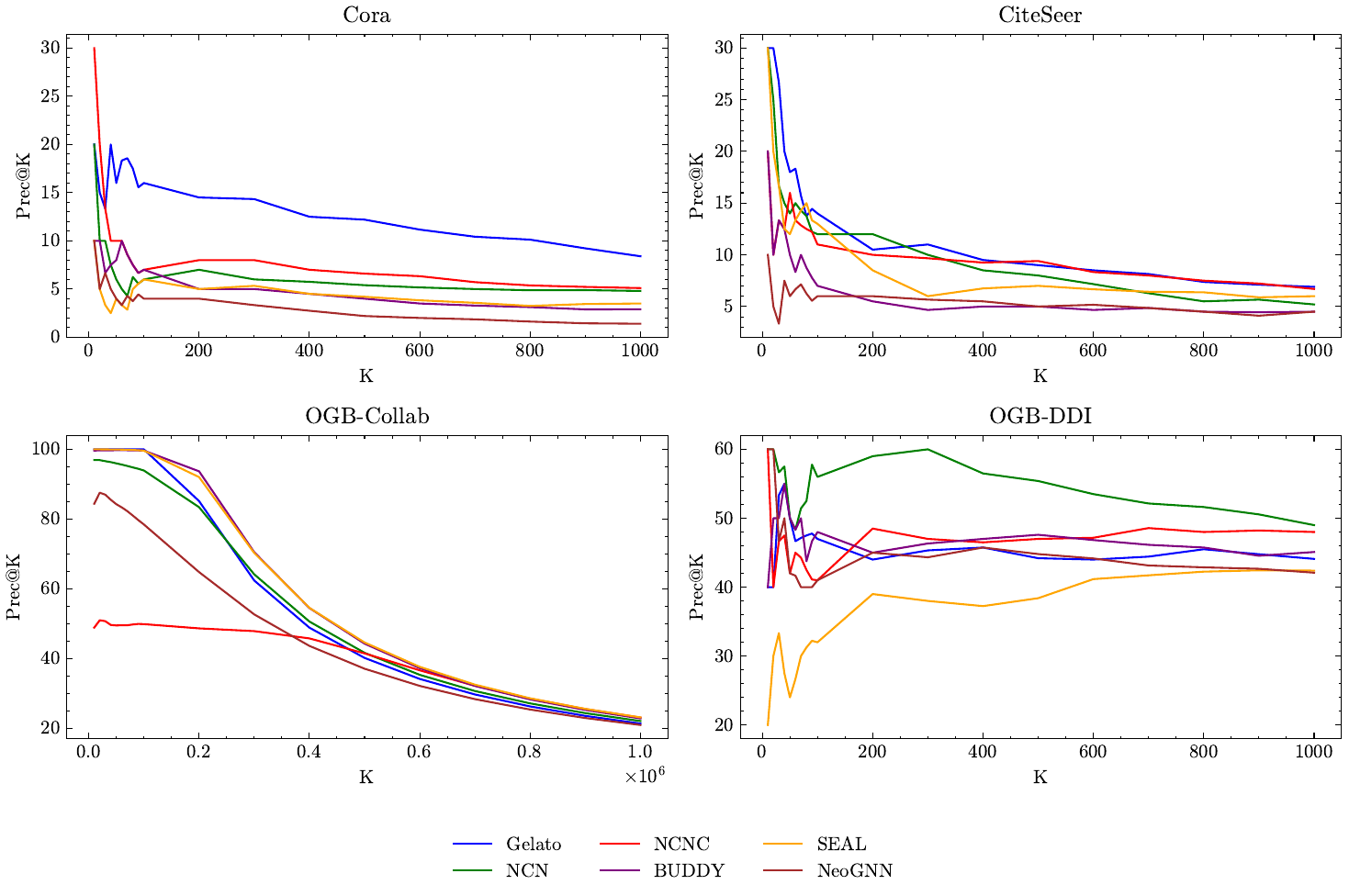}
    \caption{Link prediction comparison in terms of $prec@k$ using Cora, CiteSeer, OGBL-DDI and OGBL-Collab. All datasets were split using \textit{unbiased} sampling, except OGBL-Collab, which was split using \textit{partitioned} sampling. Gelato obtains the best performance on Cora and OGBL-Collab by a large margin and remains competitive on CiteSeer and OGBL-DDI, a dataset in which all methods struggle.}
    \label{fig::ap::prec@k}
\end{figure*}

Despite its simplicity, Gelato is consistently among the best link prediction models considering $prec@k$. We demonstrate the competitive results of Gelato against the GNN-based models by varying $k$ in Figure \ref{fig::ap::prec@k}. We also include additional AP and MRR results in Tables \ref{tab::performance_ap} and \ref{tab::performance_mrr}.

\begin{table*}[htbp]
    \setlength\tabcolsep{4.65pt}
    \small
  \centering
  \caption{Link prediction performance comparison (mean ± std AP) for all datasets considered. Gelato consistently outperforms GNN-based methods, topological heuristics, and two-stage approaches combining attributes/topology, being at least in the top-3 best-performing models in all datasets. For \textsc{Cora}, \textsc{CiteSeer}, \textsc{ogbl-ddi} and \textsc{PubMed} results we used \emph{unbiased} training, while for \textsc{ogbl-collab} \emph{partitioned} sampling is used, for scalability reasons. The top three models are colored by \textcolor{blue}{\textbf{First}}, \textcolor{red}{\textbf{Second}} and \textcolor{brown}{\textbf{Third}}.}
  \label{tab::performance_ap}%
  \begin{threeparttable}
  
    \begin{tabular}{ccccccc}
    \toprule
          &       & \textsc{Cora}  & \textsc{CiteSeer} & \textsc{PubMed} & \textsc{ogbl-ddi} & \textsc{ogbl-collab} \\
    \midrule
    \multirow{5}[2]{*}{GNN} 

        & SEAL & 2.21\tnote{*} & 2.43\tnote{*} & *** & 35.2\tnote{*} &  \textcolor{blue}{\textbf{47.43}}\tnote{*}\\
        & Neo-GNN & 2.15 ± 1.51 & 1.71 ± 0.06 & 1.21 ± 0.14 & 24.42\tnote{*} & 31.86\tnote{*} \\ 
        & BUDDY & 1.20 ± 0.25 & 1.72 ± 0.08 & OOM & 21.59 ± 1.02 & \textcolor{red}{\textbf{47.13 ± 0.22}} \\
        & NCN & 1.82 ± 0.49 & \textcolor{brown}{\textbf{2.79 ± 0.21}} & 0.06 ± 0.07 & \textcolor{red}{\textbf{44.75 ± 0.07}} & 41.38 ± 0.44 \\
        & NCNC & \textcolor{red}{\textbf{2.88 ± 0.16}} & \textcolor{red}{\textbf{3.23 ± 0.44}} & \textcolor{brown}{\textbf{1.54 ± 0.01}} & \textcolor{blue}{\textbf{44.9 ± 0.05}} & 27.67 ± 3.3 \\
    \midrule
    \multicolumn{1}{c}{\multirow{2}[2]{*}{\parbox{1.8cm}{\centering Topological Heuristics}}} 
        & CN & 1.10 ± 0.00 & 0.74 ± 0.00 & 0.36 ± 0.00 & 24.76 ± 0.00 & 24.18 ± 0.00 \\
        & AA & 2.07 ± 0.00 & 1.24 ± 0.00 & \textcolor{red}{\textbf{2.50 ± 0.00}} & 25.25 ± 0.00 & 34.28 ± 0.00 \\
        & AC & \textcolor{brown}{\textbf{2.43 ± 0.00}} & 2.65 ± 0.00 & \textcolor{red}{\textbf{2.50 ± 0.00}} & \textcolor{brown}{\textbf{29.42 ± 0.00}} & 37.92 ± 0.00\\
    \midrule
    \multicolumn{2}{c}{Gelato} & \textcolor{blue}{\textbf{3.90 ± 0.03}} & \textcolor{blue}{\textbf{4.55 ± 0.02}} & \textcolor{blue}{\textbf{2.88 ± 0.00}} & \textcolor{brown}{\textbf{29.42 ± 0.00}}      & \textcolor{brown}{\textbf{42.53}}\tnote{*} \\
    \bottomrule
    \end{tabular}%
	\begin{tablenotes}[para]
	\item[*] Run only once as each run takes 
    $>$24 hrs.
	\hspace{5pt} *** Each run takes $>$1000 hrs; \hspace{5pt} OOM: Out Of Memory.
    \end{tablenotes}
    
  \end{threeparttable}
\end{table*}

\begin{table*}
    \setlength\tabcolsep{4.65pt}
    \small
  \centering
  \caption{Link prediction performance comparison (mean ± std MRR). Gelato shows competitive performance, despite its simplicity, being in the top-3 best-performing models in almost all datasets. We highlight that Gelato is the best-performing method in \textsc{PubMed} and \textsc{ogbl-collab}, the hardest evaluation regimes since we consider the \textit{unbiased testing} scenario for both datasets. The top three models are colored by \textcolor{blue}{\textbf{First}}, \textcolor{red}{\textbf{Second}} and \textcolor{brown}{\textbf{Third}}.}
  \label{tab::performance_mrr}%
  \begin{threeparttable}

    \begin{tabular}{ccccccc}
    \toprule
          &       & \textsc{Cora}  & \textsc{CiteSeer} & \textsc{PubMed} & \textsc{ogbl-ddi} & \textsc{ogbl-collab} \\
    \midrule
    \multirow{5}[2]{*}{GNN} 
        & SEAL & 0.0204\tnote{*} & 0.235\tnote{*} & *** & 0.0071\tnote{*} &  \textcolor{red}{\textbf{4.9441}}\tnote{*}\\
        & Neo-GNN & 0.2216 ± 0.101 & 0.0969 ± 0.0285 & 0.0001 ± 0.0001 & \textcolor{brown}{\textbf{0.0098}}\tnote{*} & 0.3435\tnote{*} \\ 
        & BUDDY & 0.136 ± 0.0607 & 0.121 ± 0.0026 & OOM & 0.0094 ± 0.0003 & \textcolor{brown}{\textbf{1.2285 ± 0.0576}} \\
        & NCN & 0.1216 ± 0.0551 & \textcolor{red}{\textbf{0.1989 ± 0.0515}} & \textcolor{brown}{\textbf{0.0005 ± 0.0007}} & \textcolor{red}{\textbf{0.0117 ± 0.002}} & 0.1343 ± 0.0588 \\
        & NCNC & \textcolor{blue}{\textbf{0.4606 ± 0.1867}} & \textcolor{blue}{\textbf{0.2934 ± 0.1746}} & 0.0002 ± 0.00004 & \textcolor{blue}{\textbf{0.0171 ± 0.0133}} & 0.011 ± 0.0042 \\
    \midrule
    \multicolumn{1}{c}{\multirow{1}[2]{*}{\parbox{1.8cm}{\centering Topological Heuristics}}} 
        & CN & 0.1816 ± 0.00 & 0.0933 ± 0.00 &  0.0001 ± 0.0000 & 0.0103 ± 0.00 & 0.4767 ± 0.00 \\
        & AA & 0.1764 ± 0.00 & 0.1154 ± 0.00 &  0.0001 ± 0.0000 & 0.0104 ± 0.00 & 0.0333 ± 0.00 \\
        & AC & \textcolor{red}{\textbf{0.3069 ± 0.00}} & 0.1245 ± 0.00 & \textcolor{red}{\textbf{0.0006 ± 0.00}} & 0.0084 ± 0.00 & 0.7692 ± 0.00 \\
    \midrule
    \multicolumn{2}{c}{Gelato} & \textcolor{brown}{\textbf{0.2558 ± 0.0001}} & \textcolor{brown}{\textbf{0.1424 ± 0.0028}} & \textcolor{blue}{\textbf{0.0009 ± 0.0003}} & 0.0084 ± 0.001      & \textcolor{blue}{\textbf{6.1422}}\tnote{*} \\
    \bottomrule
    \end{tabular}%
	\begin{tablenotes}[para]
	\item[*] Run only once as each run takes 
    $>$24 hrs.
	\hspace{5pt} *** Each run takes $>$1000 hrs; \hspace{5pt} OOM: Out Of Memory.
    \end{tablenotes}
    
  \end{threeparttable}
\end{table*}

\section{Non-normalized Partitioned Sampling Results}
\label{ap::non_normalized_plots}

We recreate Figure \ref{fig::three_regimes} with non-normalized densities to show the extreme difference in the number of negative and positive pairs. 

\begin{figure*}

    \centering
    \includegraphics[width=1\textwidth]{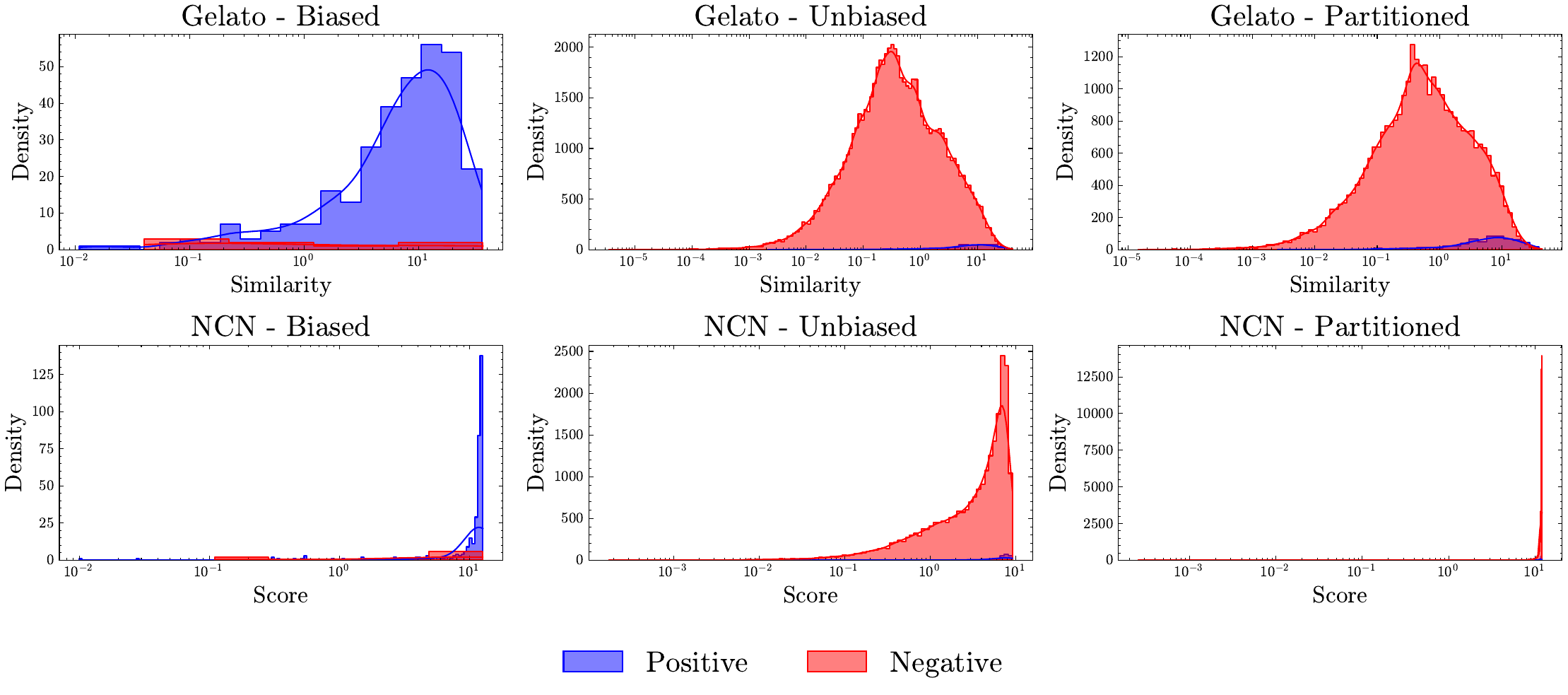}
    \caption{The non-normalized version of the Figure \ref{fig::three_regimes}. Negative pairs are represented in \textcolor{red}{red}, and positive pairs are represented in \textcolor{blue}{blue}. For unbiased and partitioned testing, negative pairs are significantly more likely than positive ones---due to graph sparsity---even for the largest values of similarity or scores. For this reason, for any decision boundary chosen, distinguishing positive pairs from negative ones is like finding ``needles in a haystack''.}
    \label{fig::ap::mesh1}
\end{figure*}

\section{Time comparison}
\label{ap::time_comparison}

In \autoref{tab::ap::time_comparison}, we compare the total training time between Gelato and our two main competitors: BUDDY and NCN (the faster version of NCNC). It is possible to notice a few patterns: Gelato suffers with graphs with a large number of nodes (mainly due to the sparse-tensor operations used in sparse autocovariance), whereas NCN gets worse results in denser networks (due to the Common Neighbors dependency), despite being the fastest. BUDDY relies on storing hashes, which results in an OOM error when running PubMed on the \textit{unbiased training} scenario and also suffers in datasets with many node features, such as CiteSeer.

\begin{table}[h]
  \caption{Estimated total training time (in hours).}
\label{tab::ap::time_comparison}
\centering
\begin{tabular}{llll}
\hline
            & BUDDY & NCN  & Gelato \\ \hline
Cora        & 0.02  & 0.14 & 0.08     \\
CiteSeer    & 38.59  & 0.19 & 0.11   \\
PubMed      & OOM   & 0.21 & 2.00    \\
OGBL-DDI    & 30.00  & 1.67  & 0.02   \\
OGBL-Collab & 5.29  & 0.87  & 30.00*  \\ \hline
\multicolumn{4}{l}{\small *Uses sparse autocovariance implementation.} \\
\end{tabular}

\end{table}



\section{Clustering times}
\label{ap:clustering}

We chose METIS\cite{karypis1998fast} as our graph partitioning method due to its scalability and the fact it produces partitions with a similar number of nodes. METIS runs as a pre-processing step in our pipeline to enable \textit{partitioned} sampling, in which we consider only negative pairs within each partition.  We display in Table \ref{tab::ap::clustering_times} the clustering time for each dataset and the number of partitions considered using the METIS implementation available in the \texttt{}{torch-sparse} (\url{https://github.com/rusty1s/pytorch_sparse}) Python package.

\begin{table}[htbp]
\caption{METIS clustering time for each dataset in seconds. METIS executes scalable and fast graph partitioning, adding negligible running time to the pre-processing step.}
\label{tab::ap::clustering_times}
\centering

\begin{tabular}{@{}lll@{}}

\toprule
            & \# Partitions & Time (s) \\ \midrule
Cora        & 10            & 0.07     \\
CiteSeer    & 10            & 0.03     \\
PubMed      & 100           & 0.16     \\
OGBL-DDI    & 20            & 0.42     \\
OGBL-Collab & 1300          & 1.91     \\ \bottomrule
\end{tabular}
\end{table}

\section{Biased training results}
\label{ap::biased_training}

\newcommand{\first}[1]{\textcolor{blue}{\textbf{#1}}}
\newcommand{\second}[1]{\textcolor{red}{\textbf{#1}}}
\newcommand{\third}[1]{\textcolor{brown}{\textbf{#1}}}

\begin{table}
    \setlength\tabcolsep{4.65pt}
    \small
  \caption{We present results (mean $\pm$ std hits@100) for GNN methods versus Gelato trained in \textit{biased training} splits and evaluated on \textit{unbiased} (Cora and CiteSeer) and \textit{partitioned} (OGBL-DDI) splits.}
  \centering
  \begin{threeparttable}

    \begin{tabular}{ccccc}
    \toprule
          &       & \textsc{Cora}  & \textsc{CiteSeer} & \textsc{ogbl-ddi}\\
    \midrule
    & SEAL & \third{1.14} \tnote{*} & \second{2.2} \tnote{*} & \third{0.05} \tnote{*} \\
    & NeoGNN & 0.82 ± 0.11 & \third{1.83 ± 0.34} & \second{0.066} \tnote{*} \\
    & BUDDY & \second{1.52 ± 0.0} & 1.32 ± 0.0 & \second{0.066 ± 0.0} \\
    & NCN & 0.0 ± 0.0 & 0.11 ± 0.16 & 0.0 ± 0.0 \\
    \midrule    
    & Gelato & \first{3.42 ± 0.33} & \first{3.44 ± 0.13} & \first{0.071 ± 0.0} \\
    \bottomrule
    \end{tabular}%
	\begin{tablenotes}[para]
	\item[*] Run only once as each run takes $>$24 hrs.
    \end{tablenotes}
    
  \end{threeparttable}
  \label{tab::ap::hits_bt}%
\end{table}

\begin{table}
    \setlength\tabcolsep{4.65pt}
    \small
  
  \caption{We present results (mean $\pm$ std hits@100) for GNN methods versus Gelato, all trained and evaluated in \textit{biased}  splits. The top three models are colored by \textcolor{blue}{\textbf{First}}, \textcolor{red}{\textbf{Second}} and \textcolor{brown}{\textbf{Third}}.}
  \centering
  
  \begin{threeparttable}

    \begin{tabular}{ccccc}
    \toprule
          &       & \textsc{Cora}  & \textsc{CiteSeer} & \textsc{ogbl-ddi}\\
    \midrule
    
    & SEAL & \second{89.18}\tnote{*}& \third{90.99}\tnote{*}& \second{4.66}\tnote{*}\\
    & NeoGNN & 62.49 ± 9.87 & 89.16 ± 0.83 & \first{5.22}\tnote{*} \\
    & BUDDY & \first{93.36 ± 0.0} & \first{98.68 ± 0.0} & 0.33 ± 0.0\\
    & NCN & \first{93.36 ± 0.54} & \second{95.93 ± 0.47} & 0.76 ± 0.6 \\
    
    \midrule    
    & Gelato & \third{78.49 ± 0.22} & 84.4 ± 0.0 & \third{4.18 ± 0.0} \\
    \bottomrule
    \end{tabular}%
	\begin{tablenotes}[para]
	\item[*] Run only once as each run takes $>$24 hrs.
    \end{tablenotes}
    
  \end{threeparttable}
  \label{tab::ap::hits_bt_bt}%
\end{table}

We present results for Gelato trained in the \textit{biased} setting and evaluated in the \textit{unbiased} / \textit{partitioned} setting in Table \ref{tab::ap::hits_bt} for the small datasets. The results show a performance degradation for most models in almost all datasets, especially for BUDDY and NCN. SEAL, NeoGNN, and Gelato have better robustness, obtaining even better results comparatively in some scenarios.

We also present results for Gelato trained and evaluated in the \textit{biased} setting in Table \ref{tab::ap::hits_bt_bt}. The results are overly optimistic, not reflecting the performance in the sparse link prediction scenarios.

\section{GNN results}
\label{ap::gnn}

\begin{figure*}
    \centering
    \includegraphics[width=1\textwidth]{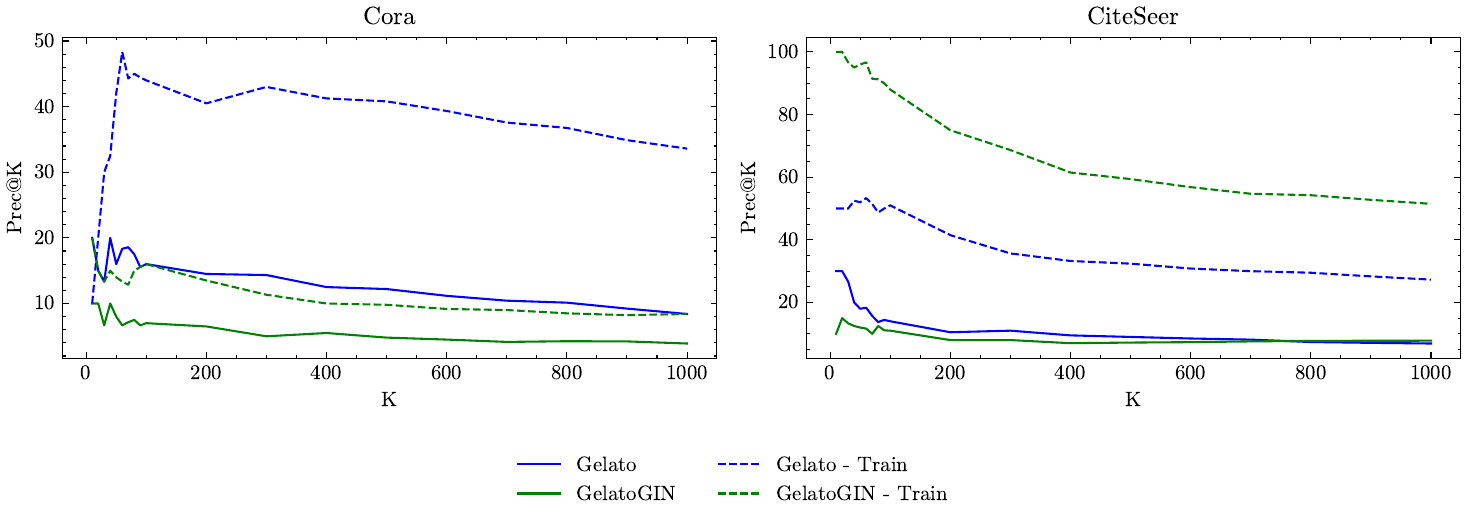}
    \caption{Performance comparison ($prec@k$) between Gelato (in \textcolor{blue}{\textbf{blue}}) against GelatoGIN (in \textcolor{purple}{\textbf{green}}), which replaces the MLP module by GIN. The dashed line represents the performance on training, while the full line represents the performance on test. We can see that despite eventually obtaining better results on training (CiteSeer), this performance is not matched by the test results, demonstrating overfitting.}
    \label{fig::ap::gnn}
\end{figure*}

We substitute the MLP module of Gelato with a GNN module using GIN \cite{xu2018powerful} (GelatoGIN). The results are displayed in Figure \ref{fig::ap::gnn}, depicting an overfitting scenario that is more pronounced in GelatoGIN considering $prec@k$ results.

\section{Sensitivity Analysis and Learning Hyperparameters}
\label{ap::sensitivity_analysis}

\begin{figure*}
    
    \centering
    \includegraphics[width=0.9\textwidth]{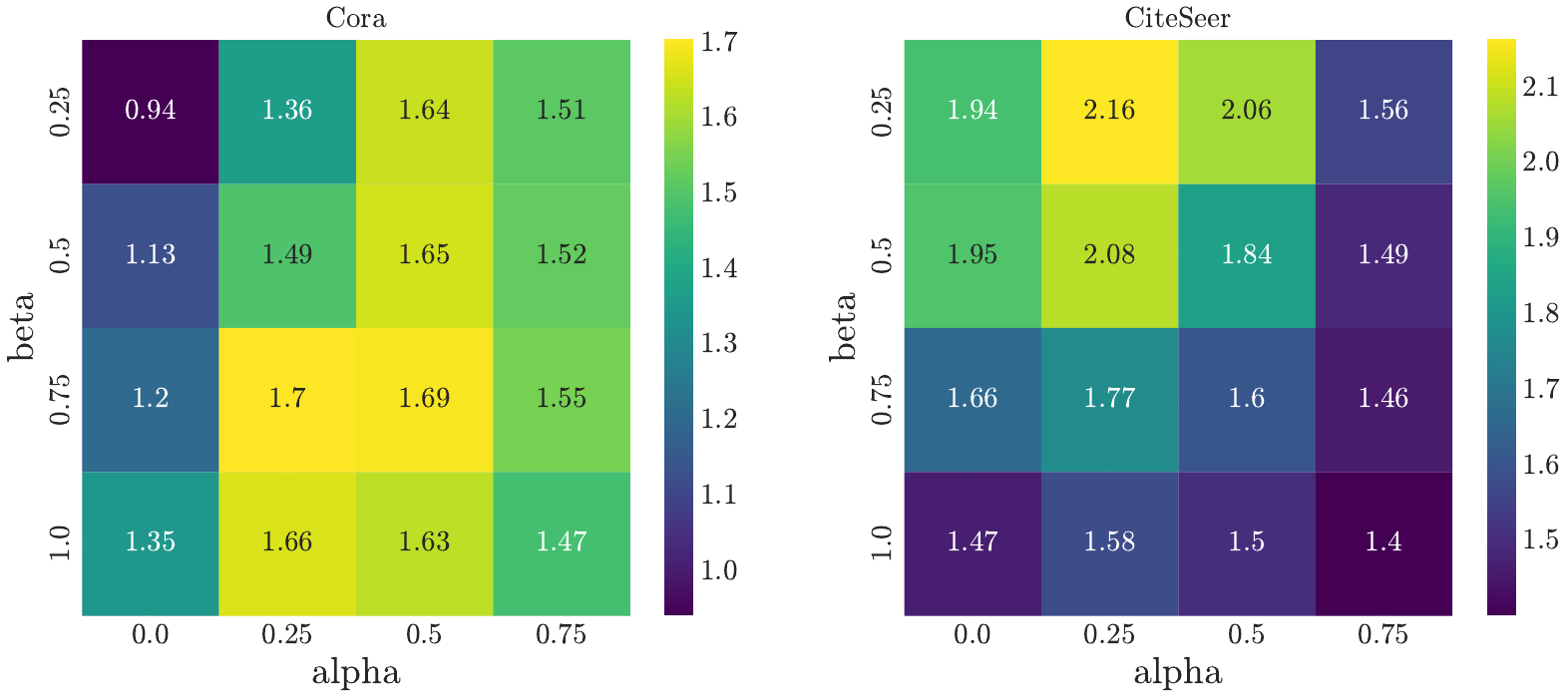}
    \caption{Sensitivity analysis of $\alpha$ and $\beta$ considering $AP$ metric.}
    \label{fig::ap::sensitivity_analysis}
\end{figure*}

\begin{figure*}
    \centering
    \includegraphics[width=0.8\textwidth]{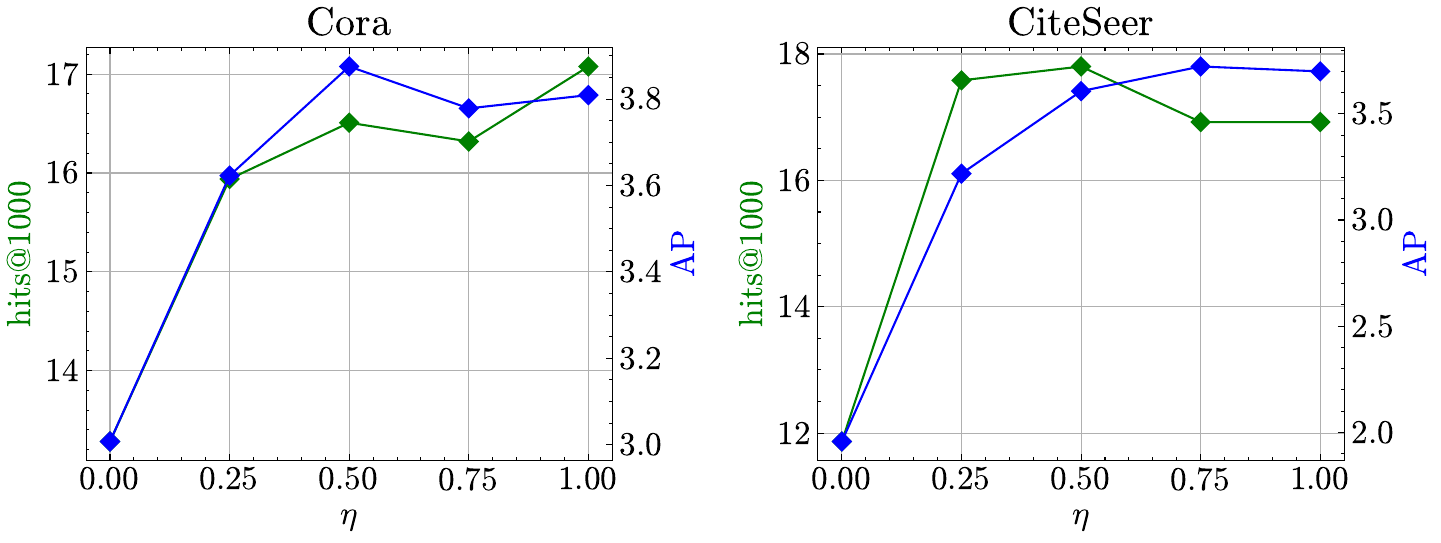}
    \caption{Performance of Gelato with different values of $\eta$. We represent hits@1000 in \textcolor{darkgreen}{\textbf{green}} and AP in \textcolor{blue}{\textbf{blue}}.}
    \label{fig::ap::eta_experiments}
\end{figure*}

We conduct a sensitivity analysis of the $\alpha$ and $\beta$ hyperparameters considering $AP$ on validation as the accuracy metric in Figure \ref{fig::ap::sensitivity_analysis}. The other two hyperparameters are set to $\eta=0$ and $T=3$ in both scenarios. We show that there is a smooth transition between the values of AP obtained through different hyperparameters, facilitating hyperparameter search. Similarly, we conduct a sensitivity analysis of $\eta$, considering both AP and hits@1000 as accuracy metrics in Figure \ref{fig::ap::eta_experiments}. The transition between values of hits@1000 and AP is smooth, showing that the addition of edges is, in general, beneficial to model performance. For the highest values of $\eta$, it is possible to see a small performance drop, which can be attributed to noisy edges added by the procedure.

\begin{figure*}

    \centering
    \includegraphics[width=0.8\textwidth]{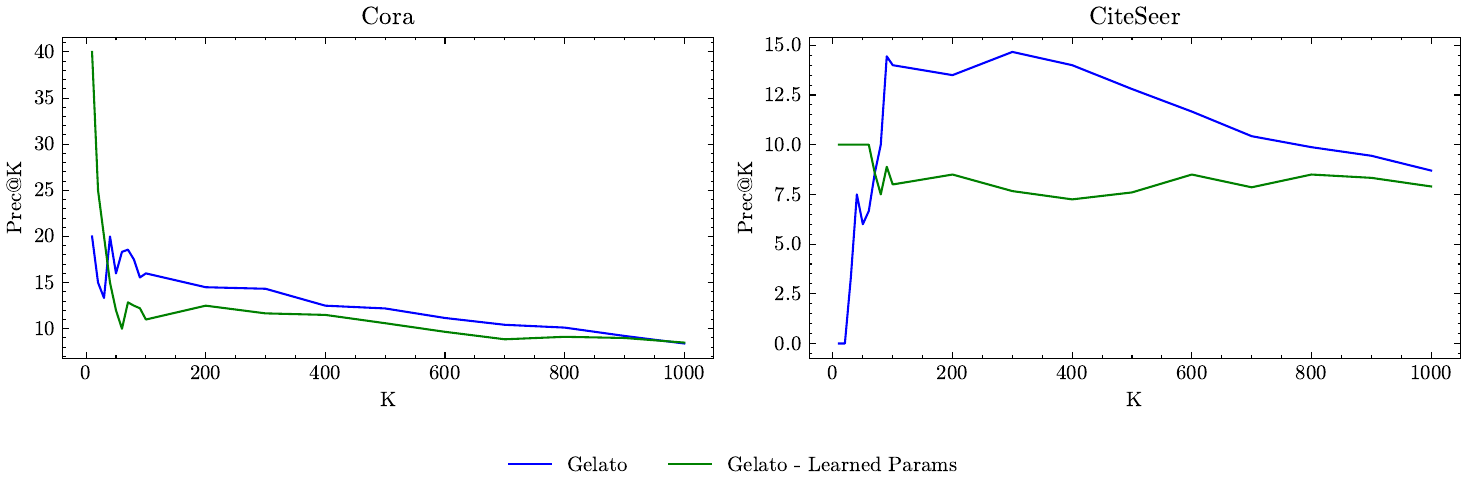}
    \vspace{1em}
    \includegraphics[width=0.8\textwidth]{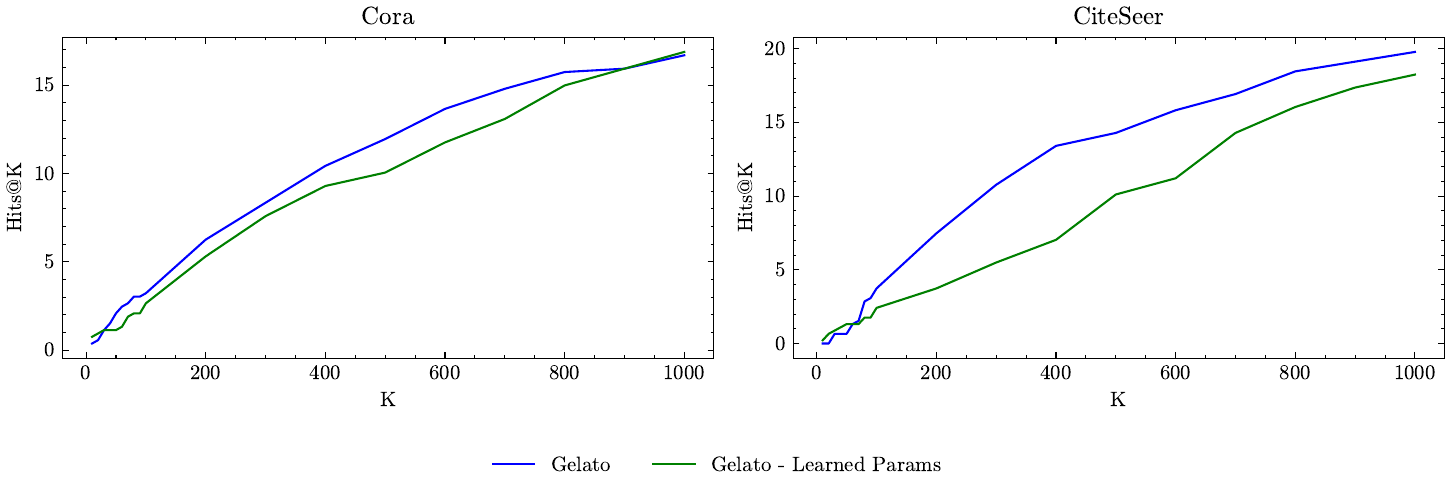}
    \caption{Results of $prec@k$ (top) and $hits@k$ (bottom) of Gelato (in \textcolor{blue}{\textbf{blue}}) against Gelato with $\alpha$ and $\beta$ as learning parameters (in \textcolor{purple}
    {\textbf{green}}). In both datasets and metrics considered, the learned $\alpha$ and $\beta$ obtained worse values than the values found by the grid search hyperparameter tuning strategy.}
    \label{fig::ap::learning_params}
\end{figure*}

We also present in \autoref{fig::ap::learning_params} results treating both $\alpha$ and $\beta$ as learnable parameters, showing that this procedure does not improve the $prec@k$ or $hits@k$ results. The values found for the hyperparameters were $\alpha=0.5670$ and $\beta=0.4694$ on Cora and $\alpha=0.5507$ and $\beta=0.4555$ on CiteSeer.

\end{document}


\maketitle

\appendix

\section{Analysis of link prediction evaluation metrics with different test settings}
\label{ap::example}
In Section 2, we present an example scenario where a bad link prediction model that ranks 1M false positives higher than the 100k true edges achieves good AUC/AP with \emph{biased testing}. Here, we provide the detailed computation steps and compare the results with those based on \emph{unbiased testing}. \looseness=-1

\begin{figure*}[htbp]
  \centering
  \subfloat[ROC]{\label{subfig::roc}\includegraphics[width=0.333\textwidth]{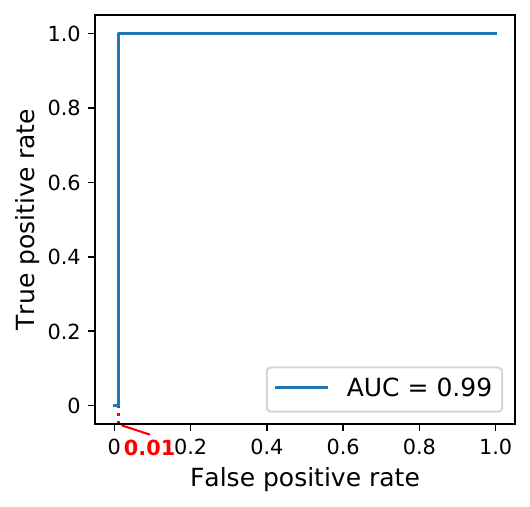}}
  \subfloat[PR under \textit{biased testing}]{\label{subfig::pr}\includegraphics[width=0.333\textwidth]{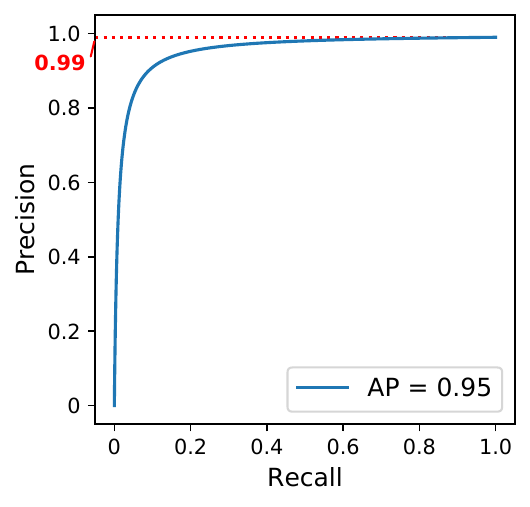}}
  \subfloat[PR under \textit{unbiased testing}]{\label{subfig::pr_unbiased}\includegraphics[width=0.333\textwidth]{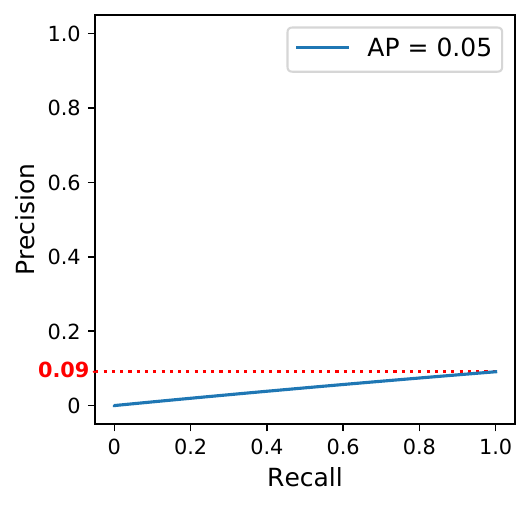}}
  \caption{Receiver operating characteristic and precision-recall curves for the bad link prediction model that ranks 1M false positives higher than the 100k true edges. 
  The model achieves 0.99 in AUC and 0.95 AP under \emph{biased testing}, while the more informative performance evaluation metric, Average Precision (AP) under \emph{unbiased testing}, is only 0.05. }
  \label{fig::roc_pr}
\end{figure*}

\autoref{subfig::roc} and \autoref{subfig::pr} show the receiver operating characteristic (ROC) and precision-recall (PR) curves for the model under \emph{biased testing} with equal negative samples. Due to the downsampling, only 100k (out of 99.9M) negative pairs are included in the test set, among which only 
${100\text{k}}/{99.9\text{M}} \times 1\text{M} \approx 1\text{k}$ pairs are ranked higher than the positive edges. In the ROC curve, this means that once the false positive rate reaches ${1\text{k}}/{100\text{k}}=0.01$, the true positive rate would reach 1.0, leading to an AUC score of 0.99. Similarly, in the PR curve, when the recall reaches 1.0, the precision is ${100\text{k}}/({1\text{k}+100\text{k}})\approx 0.99$, leading to an overall AP score of \raise.17ex\hbox{$\scriptstyle\sim$}0.95. 

By comparison, as shown in \autoref{subfig::pr_unbiased}, when the recall reaches 1.0, the precision under \emph{unbiased testing} is only
${100\text{k}}/({1\text{M}+100\text{k}})\approx 0.09$, leading to an AP score of \raise.17ex\hbox{$\scriptstyle\sim$}0.05. 
This demonstrates that evaluation metrics based on \emph{biased testing} provide an overly optimistic measurement of link prediction model performance compared to the more realistic \emph{unbiased testing} setting.

\section{Can message-passing neural networks learn autocovariance?}

\begin{figure*}[htbp]
\centering
\includegraphics[height=.2\textheight]{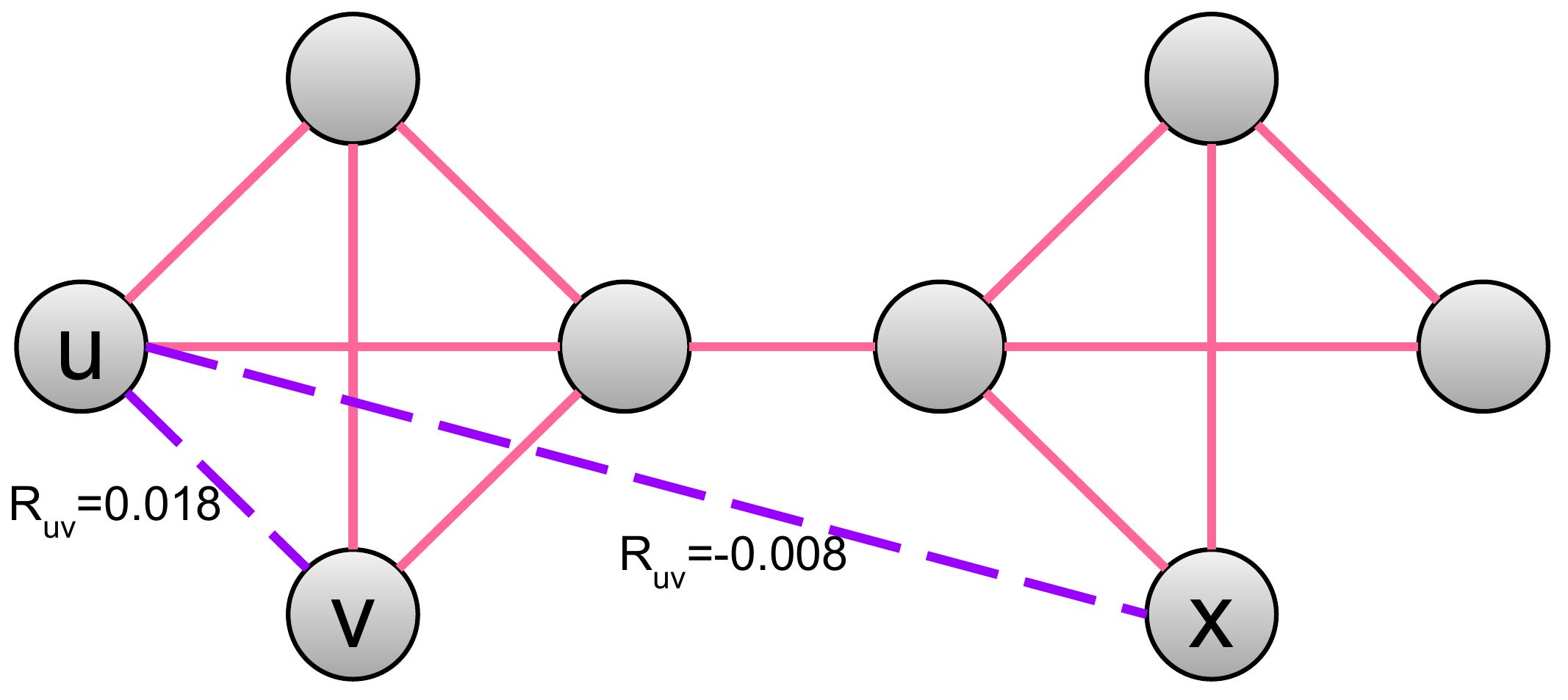}
\caption{Message-passing neural networks for link prediction cannot distinguish pairs $(u,v)$ and $(u,x)$ but they have different autocovariances, $R_{uv}=0.018$ and $R_{ux}=0.008$, for $t=2$. \label{fig::mpnn_autocov}}
\end{figure*}

Classical message-passing neural networks (MPNNs) are known to be as powerful as the 1-WL isomorphism test. Recent papers have shown how this limitation affects link prediction performance \citep{chamberlain2022graph,you2021identity,zhang2017weisfeiler,srinivasanequivalence}. Node pairs $(u,v)$ and $(u,x)$ are indistinguishable by MPNNs if $v$ and $x$ have the same receptive field (or k-hop neighborhood). Figure \ref{fig::mpnn_autocov} shows that MPNNs are also unable distinguish node pairs with different values of autocovariance within a graph $G$.

Recently, more powerful GNNs for link prediction have also been proposed \citep{hu2022two}. These GNNs have been shown to be as powerful as the 2-WL and 2-FWL isomorphism tests, which are two versions of the 2-dimensional WL test, and are more discriminative than 1-WL for link prediction. While 2-WL powerful GNNs are still not able to distinguish pairs $(u,v)$ and $(u,x)$ in Figure \ref{fig::mpnn_autocov}, 2-FWL powerful GNNs are able to. This is due to the ability of 2-FWL powerful GNNs to count open and closed triads involving pairs of nodes---$(u,v)$ is part of two triangles while $(u,x)$ is part of none. However, we notice that counting triads is not sufficient to compute probabilities of paths longer than 2 hops connecting a pair of nodes. Moreover, training a GNN based on a 2-dimensional WL test takes $O(n^3)$ time, which prevents their application to large graphs \citep{hu2022two}.

\section{Description of datasets}
\label{ap::dataset}
We use the following datasets in our experiments (with their statistics in Table  1):
\begin{itemize}
    \item \textsc{Cora} \citep{mccallum2000automating} and \textsc{CiteSeer} \citep{giles1998citeseer} are citation networks where nodes represent scientific publications (classified into seven and six classes, respectively) and edges represent the citations between them. Attributes of each node is a binary word vector indicating the absence/presence of the corresponding word from a dictionary. 
    \item \textsc{PubMed} \citep{sen2008collective} is a citation network where nodes represent scientific publications (classified into three classes) and edges represent the citations between them. Attributes of each node is a TF/IDF weighted word vector.
    \item \textsc{Photo} and \textsc{Computers} are subgraphs of the Amazon co-purchase graph \citep{mcauley2015image} where nodes represent products (classified into eight and ten classes, respectively) and edges imply that two products are frequently bought together. Attributes of each node is a bag-of-word vector encoding the product review.
\end{itemize}

We use the publicly available version of the datasets from the \texttt{pytorch-geometric} library \citep{pyg} (under the \href{https://github.com/pyg-team/pytorch_geometric/blob/master/LICENSE}{MIT licence}) curated by \citet{yang2016revisiting} and \citet{shchur2018pitfalls}.



\section{Detailed experiment settings}
\label{ap::setting}


\noindent\textbf{Positive masking.} For \emph{unbiased training}, a trick similar to \emph{negative injection} \citep{zhang2018link} in \emph{biased training} is needed to guarantee model generalizability. Specifically, we divide the training positive edges into batches and during the training with each batch $E_b$, we feed in only the residual edges $E-E_b$ as the structural information to the model. This setting simulates the testing phase, where the model is expected to predict edges without using their own connectivity information. We term this trick \textit{positive masking}.


\noindent\textbf{Other implementation details.} We add self-loops to the enhanced adjacency matrix to ensure that each node has a valid transition probability distribution that is used in computing Autocovariance. The self-loops are added to all isolated nodes in the training graph for \textsc{PubMed}, \textsc{Photo}, and \textsc{Computers}, and to all nodes for \textsc{Cora} and \textsc{CiteSeer}. Following the postprocessing of the Autocovariance matrix for embedding in \citet{random-walk-embedding}, we standardize Gelato similarity scores before computing the loss. For training with the cross entropy loss, we further add a linear layer with the sigmoid activation function to map our prediction score to a probability. We optimize our model with gradient descent via \texttt{autograd} in \texttt{pytorch} \citep{pytorch}. We find that the gradients are sometimes invalid when training our model (especially with the cross entropy loss), and we address this by skipping the parameter updates for batches leading to invalid gradients. Finally, we use $prec@100\%$ on the (unbiased) validation set as the criteria for selecting the best model from all training epochs. The maximum number of epochs for \textsc{Cora}/\textsc{CiteSeer}/\textsc{PubMed} and \textsc{Photo}/\textsc{Computers} are set to be 100 and 250, respectively. For ClusterGelato, we apply METIS \citep{karypis1998fast} as our graph partitioning algorithm, due to its scalability and a balanced number of nodes per community.

\noindent\textbf{Experiment environment.} We run our experiments in an \emph{a2-highgpu-1g} node of the Google Cloud Compute Engine. It has one NVIDIA A100 GPU with 40GB HBM2 GPU memory and 12 Intel Xeon Scalable Processor (Cascade Lake) 2nd Generation vCPUs with 85GB memory. 

\noindent\textbf{Reference of baselines.} We list link prediction baselines and their reference repositories we use in our experiments in \autoref{tab::baseline}. Note that we had to implement the batched training and testing for several baselines as their original implementations do not scale to \emph{unbiased training} and \emph{unbiased testing} without downsampling. 


\begin{table}[htbp]
  \small
  \centering
  \setlength{\tabcolsep}{3pt}
  \caption{Reference of baseline code repositories. }
    \begin{tabular}{cc}
    \toprule
     Baseline  & Repository \\
    \midrule
    GAE \citep{kipf2016semi} & \url{https://github.com/zfjsail/gae-pytorch} 
    \\
    SEAL \citep{zhang2018link}  & \url{https://github.com/facebookresearch/SEAL_OGB} 
    \\
    HGCN  \citep{chami2019hyperbolic}  & \url{https://github.com/HazyResearch/hgcn}  
    \\
    LGCN  \citep{zhang2021lorentzian}  & \url{https://github.com/ydzhang-stormstout/LGCN/}  
    \\
    TLC-GNN  \citep{yan2021link} & \url{https://github.com/pkuyzy/TLC-GNN/} 
    \\
    Neo-GNN  \citep{yun2021neo}  & \url{https://github.com/seongjunyun/Neo-GNNs} 
    \\
    NBFNet  \citep{zhu2021neural} & \url{https://github.com/DeepGraphLearning/NBFNet} 
    \\
    BScNets  \citep{chen2022bscnets}  & \url{https://github.com/BScNets/BScNets} 
    \\
    WalkPool \citep{pan2021neural}  & \url{https://github.com/DaDaCheng/WalkPooling} 
    \\
    AC  \citep{random-walk-embedding}  & \url{https://github.com/zexihuang/random-walk-embedding}  
    \\
    ELPH \citep{chamberlain2022graph} & \url{https://github.com/melifluos/subgraph-sketching}\\
    NCNC \citep{wang2023neural} & \url{https://github.com/GraphPKU/NeuralCommonNeighbor}\\
    \bottomrule
    \end{tabular}%
  \label{tab::baseline}%
\end{table}

\noindent\textbf{Number of trainable parameters.} The only trainable component in Gelato is the graph learning MLP, which for \texttt{Photo} has 208,130 parameters. By comparison, the best performing GNN-based method, Neo-GNN, has more than twice the number of parameters (455,200). 

\subsection{Sensitivity analysis}
\label{subsec::hyperparameter}
\begin{figure}[htbp]
  \centering
  \subfloat[\textsc{Photo} performance]{\includegraphics[width=0.49\columnwidth]{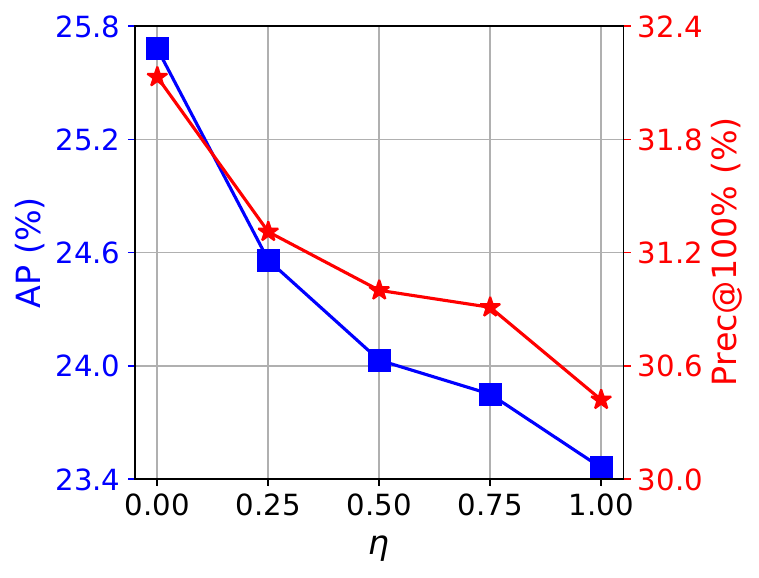}}
  \subfloat[\textsc{Cora} performance]{\includegraphics[width=0.49\columnwidth]{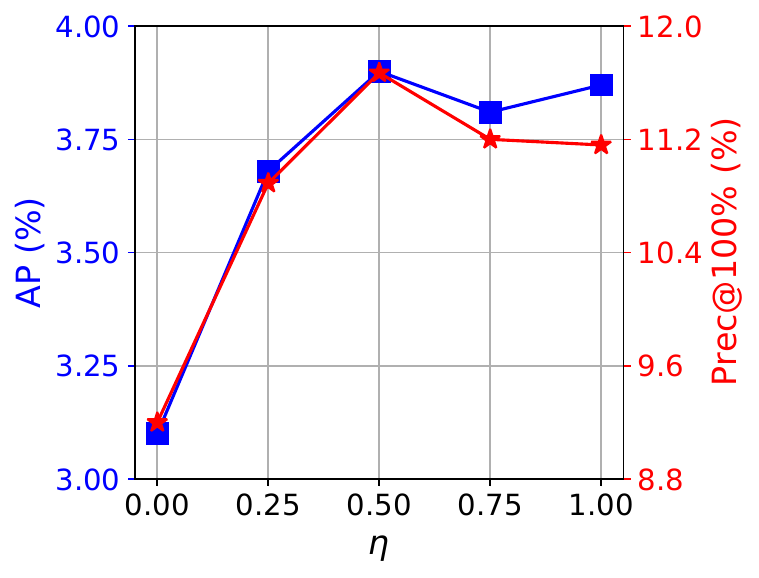}}
  \caption{Performance of Gelato with different values of $\eta$. }
  \label{fig::sensitivity_eta}
\end{figure}

\begin{figure}[htbp]
  \centering
  \subfloat[\textsc{Photo} AP scores]{\includegraphics[width=0.49\columnwidth]{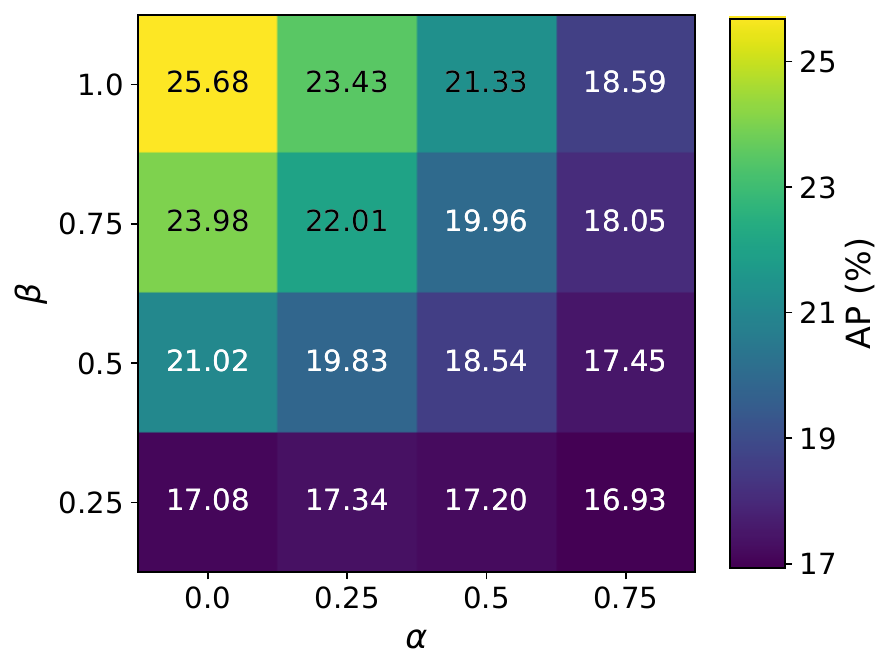}}
  \subfloat[\textsc{Photo} $prec@100\%$ scores]{\includegraphics[width=0.49\columnwidth]{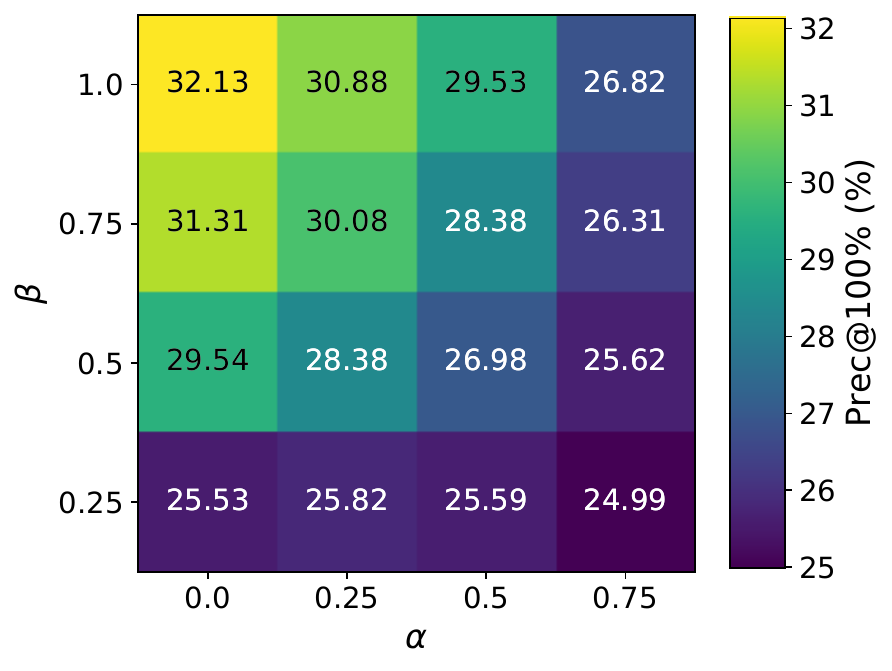}}\\
  \subfloat[\textsc{Cora} AP scores]{\includegraphics[width=0.49\columnwidth]{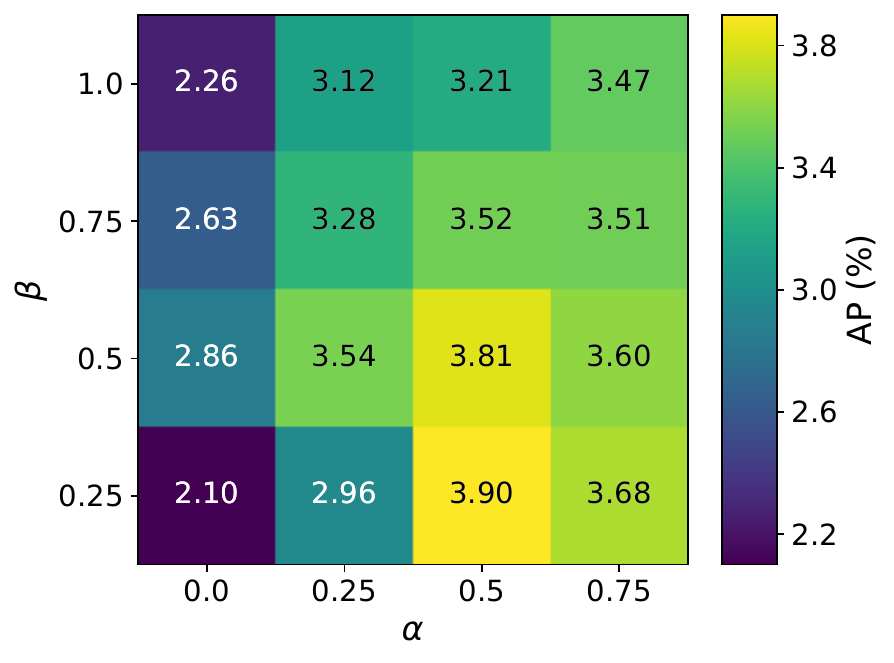}}
  \subfloat[\textsc{Cora} $prec@100\%$ scores]{\includegraphics[width=0.49\columnwidth]{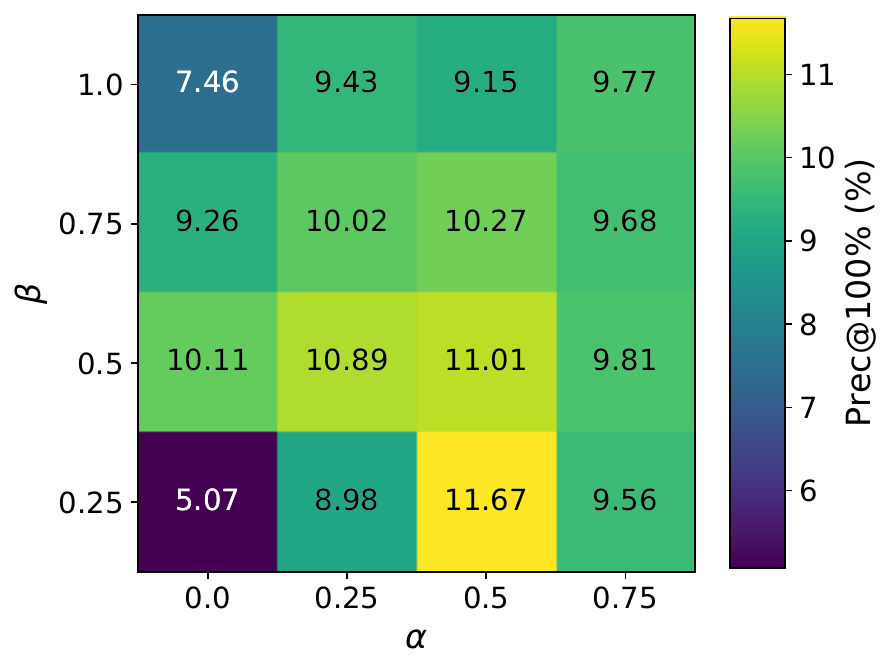}}
   \caption{Performance of Gelato with different $\alpha$ and $\beta$. \looseness=-1}
  \label{fig::sensitivity_ab}
\end{figure}

The selected hyperparameters of Gelato for each dataset are recorded in \autoref{tab::hyperparameters}, and a sensitivity analysis of $\eta$, $\alpha$, and $\beta$ are shown in \autoref{fig::sensitivity_eta} and \autoref{fig::sensitivity_ab} respectively for \textsc{Photo} and \textsc{Cora}.

\begin{table}[htbp]
  \centering
  \caption{Selected hyperparameters of Gelato. }
    \begin{tabular}{cccccc}
    \toprule
                 & \textsc{Cora}  & \textsc{CiteSeer} & \textsc{PubMed} & \textsc{Photo} & \textsc{Computers} \\
    \midrule
     $\eta$  & 0.5 & 0.75 & 0.0 & 0.0 & 0.0\\ 
    $\alpha$  & 0.5 & 0.5 & 0.0 & 0.0 & 0.0\\ 
     $\beta$  & 0.25 & 0.5 & 1.0 & 1.0 & 1.0\\ 
    \bottomrule
    \end{tabular}%
  \label{tab::hyperparameters}%
\end{table}
\begin{table*}[htbp]
    \setlength\tabcolsep{3pt}
  \centering
  \caption{Training and inference time comparison between supervised link prediction methods for \textsc{Photo}. Gelato has competitive training time (even under \emph{unbiased training}) and is significantly faster than most baselines in terms of inference, especially compared to the best GNN model, Neo-GNN.}
  \begin{threeparttable}
    \begin{tabular}{ccccccccccc}
    \toprule
          & GAE & SEAL & HGCN & LGCN & TLC-GNN & Neo-GNN & NBFNet & BScNets & MLP & Gelato \\
    \midrule
        Train & 1,022 & 11,493 & 92 & 56 & 42,440 & 14,807 & 30,896 & 115 & 232 & 1,265\\
        Test & 0.031 & 380 & 0.093 & 0.099 & 5.722 & 346 & 76,737 & 0.394 & 1.801 & 0.057\\        
    \bottomrule
    \end{tabular}%
  \end{threeparttable}
  \label{tab::running_time}%
\end{table*}

For most datasets, we find that simply setting $\beta=1.0$ and $\eta=\alpha=0.0$ leads to the best performance, corresponding to the scenario where no edges based on cosine similarity are added and the edge weights are completely learned by the MLP. For \textsc{Cora} and \textsc{CiteSeer}, however, we first notice that adding edges based on untrained cosine similarity alone leads to improved performance (see Table 2), which motivates us to set $\eta=0.5/0.75$. In addition, we find that a large trainable weight $\beta$ leads to overfitting of the model as the number of node attributes is large while the number of (positive) edges is small for \textsc{Cora} and \textsc{CiteSeer} (see Table 1). We address this by decreasing the relative importance of trained edge weights ($\beta=0.25/0.5$) and increasing that of the topological edge weights ($\alpha=0.5$), which leads to better generalization and improved performance. Based on our experiments, these hyperparameters can be easily tuned using simple hyperparameter search techniques, such as line search, using a small validation set.

\subsection{Ablation study}
\label{subsec::ablation}
We have demonstrated the superiority of Gelato over its individual components and two-stage approaches in Table 2 and analyzed the effect of losses and training settings in Section 4.3. Here, we collect the results with the same hyperparameter setting as Gelato and present a comprehensive ablation study in \autoref{tab::ablation}. Specifically, \emph{Gelato$-$MLP} (\emph{AC}) represents Gelato without the MLP (Autocovariance) component, i.e., only using Autocovariance (MLP) for link prediction. 
\emph{Gelato$-$NP} (\emph{UT}) replaces the proposed N-pair loss (\emph{unbiased training}) with the cross entropy loss (\emph{biased training}) applied by the baselines. Finally, \emph{Gelato$-$NP+UT} replaces both the loss and the training setting. 

\begin{table*}[htbp]
    \setlength\tabcolsep{7pt}
  \centering
  \caption{Results of the ablation study based on AP scores. Each component of Gelato plays an important role in enabling state-of-the-art link prediction performance. }
  \begin{threeparttable}
    \begin{tabular}{lccccc}
    \toprule
          & \textsc{Cora}  & \textsc{CiteSeer} & \textsc{PubMed} & \textsc{Photo} & \textsc{Computers} \\
    \midrule
        \emph{Gelato$-$MLP} & 2.43 ± 0.00 & 2.65 ± 0.00 & 2.50 ± 0.00 & 16.63 ± 0.00 &  11.64 ± 0.00 \\
        \emph{Gelato$-$AC} & 1.94 ± 0.18 & 3.91 ± 0.37 & 0.83 ± 0.05 & 7.45 ± 0.44 & 4.09 ± 0.16  \\
        \emph{Gelato$-$NP+UT} & 2.98 ± 0.20 & 1.96 ± 0.11 & 2.35 ± 0.24 & 14.87 ± 1.41 & 9.77 ± 2.67 \\
        \emph{Gelato$-$NP} & 1.96 ± 0.01 & 1.77 ± 0.20 & 2.32 ± 0.16 & 19.63 ± 0.38 & 9.84 ± 4.42 \\
        \emph{Gelato$-$UT} & 3.07 ± 0.01 & 1.95 ± 0.05 & 2.52 ± 0.09 & 23.66 ± 1.01 & 11.59 ± 0.35  \\
        \emph{Gelato} & \textbf{3.90 ± 0.03} & \textbf{4.55 ± 0.02} & \textbf{2.88 ± 0.09} & \textbf{25.68 ± 0.53} & \textbf{18.77 ± 0.19} \\
    \bottomrule
    \end{tabular}%
  \end{threeparttable}
  \label{tab::ablation}%
\end{table*}

We observe that removing either MLP or Autocovariance leads to inferior performance, as the corresponding attribute or topology information would be missing. Further, to address the class imbalance problem of link prediction, both the N-pair loss and \emph{unbiased training} are crucial for effective training of Gelato.


\subsection{Running time}
\begin{figure}[htb]
  \centering
  \subfloat[Training time until convergence]{%
  \label{subfig::training_time}%
  \includegraphics[width=0.5\textwidth]{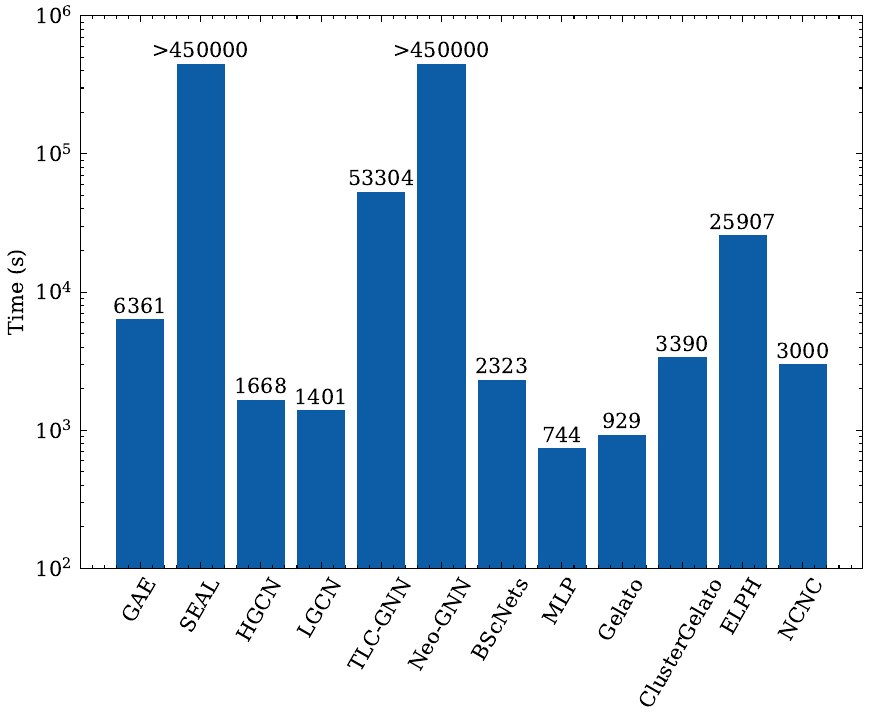}}

    \qquad
  
  \subfloat[Inference time per \emph{unbiased testing}]{\label{subfig::testing_time}\includegraphics[width=0.5\textwidth]{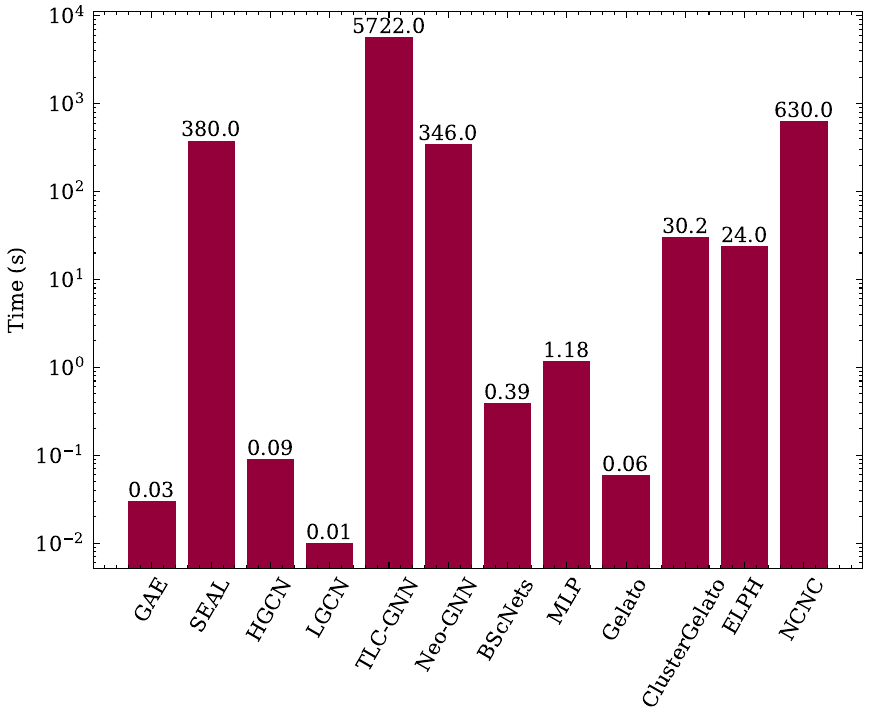}}

  \caption{Training and testing time comparison between supervised link prediction methods for Photo under full training. Gelato, while achieving the best performance, is also the second most efficient method in terms of total training time, slower only than the vanilla MLP. ClusterGelato is still among the fastest methods, despite being slower than Gelato.
  }
  \label{fig:training_time}
\end{figure}

We compare Gelato and ClusterGelato with other supervised link prediction methods in terms of running time in Figure \ref{fig:training_time}. As the only method that applies full training, Gelato shows a competitive training speed that is \textbf{3.2x} faster than the best-performing method, NCNC. In terms of inference time, both Gelato and ClusterGelato are significantly faster than most baselines with 10,500 and 20 times speedup compared to NCNC, respectively. 

\section{ClusterGelato}

\paragraph{VRAM consumption}

To evaluate the effect of the number of communities $p$, we ran ClusterGelato on the Computers dataset varying this parameter and exposed our results in Table \ref{tab:p_param}. We also provide a comparison between the VRAM consumption of Gelato and ClusterGelato in Table \ref{tab:vram_consumption}, showing average savings of \textbf{2.8x}. The parameter $p$ provides control over VRAM consumption to ClusterGelato, but introduces a trade-off between scalability and link prediction performance. Thus, it is advised always to select the lowest possible amount of communities to maximize link prediction metrics. 



\paragraph{Link Prediction performance} 

We evaluate the link prediction performance and the scalability of ClusterGelato compared to two pairwise heuristics, CN and AA, and two GNN-based methods, GAE and NCNC. All models were evaluated using \textit{unbiased testing} within each community, i.e. both negative and positive pairs sets were sampled only from within each community $G_i$ (no intercluster pairs), thus being a more challenging link prediction scenario. The results exposed in Table \ref{tab::performance_ap_clustergelato} show that ClusterGelato obtains competitive link prediction performance against all the methods while being scalable and enabling fast inference.

\begin{table}[h]
\centering
\caption{Impact of the number of communities $p$ in VRAM consumption and inference time, evaluated on Computers. There is a trade-off between scalability and performance that suggests that the best results are obtained using $p$ as small as possible. The first row ($p=1$) refers to Gelato, the base comparison.}
\label{tab:p_param}
\begin{tabular}{@{}ccc@{}}
\toprule
$p$ & \begin{tabular}[c]{@{}c@{}}VRAM \\ (GB)\end{tabular} & \begin{tabular}[c]{@{}c@{}}Inference Time\\ (s)\end{tabular} \\ \midrule
1   & 21.5                                                 & 34                                                           \\
2   & 15.0                                                 & 45                                                           \\
4   & 10.8                                                 & 58                                                           \\
8   & 5.6                                                  & 52                                                           \\
16  & 4.2                                                  & 34                                                           \\
32  & 3.0                                                  & 26                                                           \\
64  & 2.7                                                  & 24                                                           \\ \bottomrule
\end{tabular}
\end{table}

\begin{table}[h]
\centering
\caption{VRAM consumption (GB) comparison between Gelato and ClusterGelato. Numbers in the last row show the proportion of {\textbf{decrease}} of VRAM consumption.}
\label{tab:vram_consumption}
\begin{tabular}{lccccc}
\hline
              & \begin{tabular}[c]{@{}c@{}}Cora\end{tabular} & \begin{tabular}[c]{@{}c@{}}CiteSeer\end{tabular} & \begin{tabular}[c]{@{}c@{}}PubMed\end{tabular} & \begin{tabular}[c]{@{}c@{}}Photo\end{tabular} & \begin{tabular}[c]{@{}c@{}}Computers\end{tabular} \\ \hline
Gelato        & 3.1                                                   & 6.2                                                       & 34.0                                                    & 10.3                                                    & 21.4                                                                                                           \\
ClusterGelato & \textbf{2.1}                                                   & \textbf{2.5}                                                      & \textbf{6.9}                                            & \textbf{7.4}                                           & \textbf{10.3}                                                                                    \\ \hline
VRAM Gain     & { \textbf{(1.5x)}}                & { \textbf{(2.5x)}}                    & { \textbf{(4.9x)}}                 & { \textbf{(1.4x)}}                 & { \textbf{(2.1x)}}                                                                        
\end{tabular}
\end{table}



  


\begin{table}[!h]
\caption{Link prediction performance comparison (mean ± std AP). ClusterGelato shows competitive performance to the baselines in the intracluster link prediction scenario. GAE and NCNC were trained using \textit{biased training}, while ClusterGelato uses \textit{unbiased training}. All methods were evaluated with the same \textit{unbiased testing} splits.}
\label{tab::performance_ap_clustergelato}
\centering
\begin{tabular}{cccccc}
\hline
              & Cora & CiteSeer & PubMed & Photo & Computers \\ \hline
CN            & 1.26 ± 0.00 & 2.34 ± 0.00    & 0.58 ± 0.00  & 7.84 ± 0.00  & 5.56 ± 0.00      \\
AA            & 2.34 ± 0.00 & 0.02 ± 0.00    & 0.74 ± 0.00  & 9.80 ± 0.00  & 7.10 ± 0.00      \\
GAE           & 2.56 ± 0.20 & 0.02 ± 0.02    & 1.58 ± 0.03 & 15.63 ± 0.95 & 10.35 ± 0.97    \\
NCNC          & \textbf{3.15 ± 0.49} & 0.02 ± 0.01    & 1.54 ± 0.01 & 17.00 ± 0.97 & 10.04 ± 1.99    \\
\textbf{ClusterGelato} & 2.88 ± 0.15 & \textbf{2.12 ± 0.16}    & \textbf{3.32 ± 0.20} & \textbf{17.81 ± 0.79} & \textbf{10.86 ± 0.19}    \\ \hline
\end{tabular}
\end{table}

\section{Number of communities used for each dataset}
Table \ref{tab:p_values} shows, for reproducibility purposes, the different values of $p$ used in all experiments.

\begin{table}[h]
\centering
\caption{Number of communities $p$ used in all the link prediction performance and running time experiments, except when the $p$ value is explicitly mentioned.}
\label{tab:p_values}
\begin{tabular}{@{}ccccccc@{}}
\toprule
  & Cora & CiteSeer & PubMed & Photo & Computers  \\ \midrule
$p$ & 2    & 2        & 3     & 2    & 4               \\ \bottomrule
\end{tabular}
\end{table}

%

 {
\small
\bibliographystyle{unsrtnat}
\bibliography{reference}
}